\documentclass{article}

\PassOptionsToPackage{numbers, compress}{natbib}

\usepackage[final]{neurips_2020}

\usepackage[utf8]{inputenc} 
\usepackage[T1]{fontenc}    
\usepackage{hyperref}       
\usepackage{url}   
\usepackage{booktabs}       
\usepackage{amsfonts}       
\usepackage{nicefrac}       
\usepackage{microtype}      

\usepackage{multirow}
\usepackage{wrapfig}
\usepackage{graphicx}
\usepackage{subfigure}
\usepackage{amsmath}
\usepackage{color}
\usepackage{algorithm}
\usepackage{algorithmic}
\usepackage{enumitem}
\usepackage[compact]{titlesec}

\def \xb {\mathbf{x}}
\def \fb {\mathbf{f}}
\def \wb {\mathbf{w}}
\def \db {\mathbf{d}}
\def \vb {\mathbf{v}}

\title{Adversarial Weight Perturbation Helps \\ Robust Generalization}

\author{%
Dongxian Wu\textsuperscript{1,3}  \qquad    Shu-Tao Xia\textsuperscript{1,3}  \qquad  Yisen Wang\textsuperscript{2}\footnotemark[2]
\\
\textsuperscript{1}Tsinghua University\\
\textsuperscript{2}Key Lab. of Machine Perception (MoE), School of EECS, Peking University\\
\textsuperscript{3}PCL Research Center of Networks and Communications, Peng Cheng Laboratory\\
}

\begin{document}

\maketitle

\renewcommand{\thefootnote}{\fnsymbol{footnote}}
\footnotetext[2]{Corresponding author: Yisen Wang (yisen.wang@pku.edu.cn)}

\begin{abstract}
The study on improving the robustness of deep neural networks against adversarial examples grows rapidly in recent years. Among them, adversarial training is the most promising one, which flattens the \textit{input loss landscape} (loss change with respect to input) via training on adversarially perturbed examples. However, how the widely used \textit{weight loss landscape} (loss change with respect to weight) performs in adversarial training is rarely explored. In this paper, we investigate the weight loss landscape from a new perspective, and identify a clear correlation between the flatness of weight loss landscape and robust generalization gap. Several well-recognized adversarial training improvements, such as early stopping, designing new objective functions, or leveraging unlabeled data, all implicitly flatten the weight loss landscape. Based on these observations, we propose a simple yet effective \textit{Adversarial Weight Perturbation (AWP)} to explicitly regularize the flatness of weight loss landscape, forming a \textit{double-perturbation} mechanism in the adversarial training framework that adversarially perturbs both inputs and weights. Extensive experiments demonstrate that AWP indeed brings flatter weight loss landscape and can be easily incorporated into various existing adversarial training methods to further boost their adversarial robustness. 
\end{abstract}

\section{Introduction}
Although deep neural networks (DNNs) have been widely deployed in a number of fields such as computer vision \citep{he2016deep}, speech recognition \citep{wang2017residual}, and natural language processing \citep{devlin2019bert}, they could be easily fooled to confidently make incorrect predictions by adversarial examples that are crafted by adding intentionally small and human-imperceptible perturbations to normal examples \citep{szegedy2013intriguing,goodfellow2015explaining,wu2020skip,bai2020improving}. As DNNs penetrate almost every corner in our daily life, ensuring their security, \textit{e.g.,} improving their robustness against adversarial examples, becomes more and more important. 

There have emerged a number of defense techniques to improve adversarial robustness of DNNs
\citep{papernot2016distillation,madry2018towards,wang2019convergence}. Across these defenses, Adversarial Training (AT) \citep{goodfellow2015explaining, madry2018towards} is the most effective and promising approach, which not only demonstrates moderate robustness, but also has thus far not been comprehensively attacked \citep{athalye2018obfuscated}. AT directly incorporates adversarial examples into the training process to solve the following optimization problem: 
\begin{equation}
    \min_{\wb} \rho(\wb), \quad \text{where} \quad \rho(\wb)=\frac{1}{n} \sum_{i=1}^n \max_{\Vert \xb_i^\prime - \xb_i \Vert_p \leq \epsilon} \ell(f_{\wb}(\xb_i^\prime), y_i),
\label{eqn:robust_loss}
\end{equation}
where $n$ is the number of training examples, $\xb_i^\prime$ is the adversarial example within the $\epsilon$-ball (bounded by an $L_{p}$-norm) centered at natural example $\xb_i$, $f_\wb$ is the DNN with weight $\wb$, $\ell(\cdot)$ is the standard classification loss (\textit{e.g.}, the cross-entropy (CE) loss), and $\rho(\wb)$ is called the ``adversarial loss'' following \citet{madry2018towards}. Eq. \eqref{eqn:robust_loss} indicates that AT restricts the change of loss when its input is perturbed (\textit{i.e.}, flattening the \textbf{input loss landscape}) to obtain a certain of robustness, but its robustness is still far from satisfactory because of the huge robust generalization gap \cite{schmidt2018adversarially,rice2020overfitting}, for example, an adversarially trained PreAct ResNet-18 \cite{he2016identity}  on CIFAR-10 \cite{krizhevsky2009learning} only has 43\% test robustness, even it has already achieved $84\%$ training robustness after 200 epochs (see Figure \ref{fig:motivation_piecewise}). Its robust generalization gap reaches $41\%$, which is very different from the standard training (on natural examples) whose standard generalization gap is always lower than $10\%$. Thus, how to mitigate the robust generalization gap becomes essential for the robustness improvement of adversarial training methods. 

Recalling that \textbf{weight loss landscape} is a widely used indicator to characterize the standard generalization gap in standard training scenario \cite{neyshabur2017exploring,li2018visualizing}, however, there are few explorations under adversarial training, among which, \citet{prabhu2019understanding} and \citet{yu2018interpreting} tried to use the pre-generated adversarial examples to explore but failed to draw the expected conclusions. In this paper, we explore the weight loss landscape under adversarial training using on-the-fly generated adversarial examples, and identify a strong connection between the flatness of weight loss landscape and robust generalization gap. Several well-recognized adversarial training improvements, \textit{i.e.}, AT with early stopping \citep{rice2020overfitting}, TRADES \citep{Zhang2019theoretically}, MART \citep{wang2020improving} and RST \citep{carmon2019unlabeled}, all implicitly flatten the weight loss landscape to narrow the robust generalization gap. Motivated by this, we propose an explicit weight loss landscape regularization, named \textit{Adversarial Weight Perturbation (AWP)}, to directly restrict the flatness of weight loss landscape. Different from random perturbations \citep{he2019parametric}, AWP injects the strongest worst-case weight perturbations, forming a \textit{double-perturbation} mechanism (\textit{i.e.}, inputs and weights are both adversarially perturbed) in the adversarial training framework. AWP is generic and can be easily incorporated into existing adversarial training approaches with little overhead. Our main contributions are summarized as follows:

\begin{itemize}
    \item We identify the fact that flatter weight loss landscape often leads to smaller robust generalization gap in adversarial training via characterizing the weight loss landscape using adversarial examples generated on-the-fly. 
    
    \item We propose \textit{Adversarial Weight Perturbation (AWP)} to explicitly regularize the weight loss landscape of adversarial training, forming a double-perturbation mechanism that injects the worst-case input and weight perturbations. 
    
    \item Through extensive experiments, we demonstrate that AWP consistently improves the adversarial robustness of state-of-the-art methods by a notable margin. 
    
\end{itemize}

\section{Related Work}

\subsection{Adversarial Defense}
Since the discovery of adversarial examples, many defensive approaches have been developed to reduce this type of security risk such as defensive distillation \citep{papernot2016distillation}, feature squeezing \citep{xu2017feature}, input denoising \citep{bai2019hilbert}, adversarial detection \citep{ma2018characterizing}, gradient regularization \citep{papernot2017practical,tramer2018ensemble}, and adversarial training \citep{goodfellow2015explaining,madry2018towards,wang2019convergence}. Among them, adversarial training has been demonstrated to be the most effective method \citep{athalye2018obfuscated}. Based on adversarial training, a number of new techniques are introduced to enhance its performance further. 

\textbf{TRADES \citep{Zhang2019theoretically}.} TRADES optimizes an upper bound of adversarial risk that is a trade-off between accuracy and robustness:
\begin{equation}
\begin{aligned}
\rho^{\text{TRADES}}(\wb) &=
\frac{1}{n} \sum_{i=1}^n \bigg\{ \text{CE} \big(f_\wb(\xb_i), y_i \big) + \beta \cdot \max \text{KL} \big(f_\wb(\xb_i) \Vert f_\wb(\xb^\prime_i) \big) \bigg\},
\label{eqn:trades_loss}
\end{aligned}
\end{equation}
where $\text{KL}$ is the Kullback--Leibler divergence, $\text{CE}$ is the cross-entropy loss, and $\beta$ is the hyperparameter to control the trade-off between natural accuracy and robust accuracy. 

\textbf{MART \citep{wang2020improving}.} MART incorporates an explicit differentiation of misclassified examples as a regularizer of adversarial risk:
\begin{equation}
\begin{aligned}
\rho^{\text{MART}}(\wb) = 
  \frac{1}{n} \sum_{i=1}^n \bigg\{ \text{BCE} \big(\fb_\wb(\xb^\prime_i), y_i\big) + \lambda \cdot \text{KL}\big(\fb_\wb(\xb_i) \Vert \fb_\wb(\xb^\prime_i)\big) \cdot \big(1-[\fb_\wb(\xb_i)]_{y_i}\big) \bigg\}, 
\end{aligned}
\label{eqn:mart_loss}
\end{equation}
where $[\fb_\wb(\xb_i)]_{y_i}$ denotes the $y_i$-th element of output vector $\fb_\wb(\xb_i)$ and $\text{BCE}\big(\fb_\wb(\xb_i), y_i \big) = - \log\big( [\fb_\wb(\xb_i^\prime)]_{y_i} \big) - \log \big( 1 - \max_{k \neq y_i} [\fb_\wb(\xb^\prime_i)]_k \big)$.

\textbf{Semi-Supervised Learning (SSL) \citep{carmon2019unlabeled,stanforth2019labels,najafi2019robustness,zhai2019adversarially}.} SSL-based methods utilize additional unlabeled data. They first generate pseudo labels for unlabeled data by training a natural model on the labeled data. Then, adversarial loss $\rho(\wb)$ is applied to train a robust model based on both labeled and unlabeled data:
\begin{equation}
    \rho^{\text{SSL}}(\wb) = \rho^{\text{labeled}}(\wb) + \lambda \cdot \rho^{\text{unlabeled}}(\wb),
\label{eqn:rst_loss}
\end{equation}
where $\lambda$ is the weight on unlabeled data. $\rho^{\text{labeled}}(\wb)$ and  $\rho^{\text{unlabeled}}(\wb)$ are usually the same adversarial loss. For example, RST in \citet{carmon2019unlabeled} uses TRADES loss and semi-supervised MART in \citet{wang2020improving} uses MART loss.

\subsection{Robust Generalization}

Compared with standard generalization (on natural examples), training DNNs with robust generalization (on adversarial examples) is particularly difficult \citep{madry2018towards}, and often possesses significantly higher sample complexity \citep{khim2018adversarial,yin2019rademacher,montasser2019vc} and needs more data \citep{schmidt2018adversarially}. \citet{nakkiran2019adversarial} showed that a model requires more capacity to be robust. \citet{tsipras2018robustness} and \citet{Zhang2019theoretically} demonstrated that adversarial robustness may be inherently at odds with natural accuracy. Moreover, there are a series of works studying the robust generalization from the view of loss landscape. In adversarial training, there are two types of loss landscape: 1) \textit{input loss landscape} which is the loss change with respect to 
the input. It depicts the change of loss in the vicinity of training examples. AT explicitly flattens the input loss landscape by training on adversarially perturbed examples, while there are other methods doing this by gradient regularization \cite{lyu2015unified,ross2018improving}, curvature regularization \cite{moosavi2019robustness}, and local linearity regularization \cite{qin2019adversarial}. These methods are fast on training but only achieve comparable robustness with AT. 2) \textit{weight loss landscape} which is the loss change with respect to the weight. It reveals the geometry of loss landscape around model weights. Different from the standard training scenario where numerous studies have revealed the connection between the weight loss landscape and their standard generalization gap \cite{keskar2017large,neyshabur2017exploring,chaudharientropy}, whether the connection exists in adversarial training is still under exploration. \citet{prabhu2019understanding} and \citet{yu2018interpreting} tried to establish this connection in adversarial training but failed due to the inaccurate weight loss landscape characterization.

Different from these studies, we characterize the weight loss landscape from a new perspective, and identify a clear relationship between weight loss landscape and robust generalization gap.

\section{Connection of Weight Loss Landscape and Robust Generalization Gap}\label{sec:geometry}
In this section, we first propose a new method to  characterize the weight loss landscape, and then investigate it from two perspectives: 1) in the training process of adversarial training, and 2) across different adversarial training methods, which leads to a clear correlation between weight loss landscape and robust generalization gap. To this end, some discussions about the weight loss landscapes are provided. 

\textbf{Visualization.} 
We visualize the weight loss landscape by plotting the adversarial loss change when moving the weight $\wb$ along a random direction $\db$ with magnitude $\alpha$: 
\begin{equation}
    g(\alpha) = \rho(\wb + \alpha \db) = \frac{1}{n} \sum_{i=1}^n \max_{\Vert \xb_i^\prime - \xb_i \Vert_p \leq \epsilon} \ell(f_{\wb + \alpha \db}(\xb_i^\prime), y_i), 
\end{equation}
where $\db$ is sampled from a Gaussian distribution and filter normalized by $\db_{l,j} \leftarrow \frac{\db_{l,j}}{\Vert \db_{l,j} \Vert_F} \Vert \wb_{l,j} \Vert_F$ ($\db_{l,j}$ is the $j$-th filter at the $l$-th layer of $\db$ and $\Vert \cdot \Vert_F$ denotes the Frobenius norm) to eliminate the scaling invariance of DNNs following \citet{li2018visualizing}. The adversarial loss $\rho$ is usually approximated by the cross-entropy loss on adversarial examples following \citet{madry2018towards}. Here, we generate adversarial examples on-the-fly by PGD (attacks are reviewed in Appendix \ref{sec:attack}) for the current perturbed model $f_{\wb + \alpha \db}$ and then compute their cross-entropy loss (refer to Appendix \ref{sec:visualization_code} for details). The key difference to previous works lies on the adversarial examples used for visualization. \citet{prabhu2019understanding} and \citet{yu2018interpreting} used a fixed set of pre-generated adversarial examples on the original model $f_{\wb}$ in the visualization process, which will severely underestimate the adversarial loss due to the inconsistency between the source model (original model $f_{\wb}$) and the target model (current perturbed model $f_{\wb + \alpha \db}$). Considering $\db$ is randomly selected, we repeat the visualization 10 times with different $\db$ in Appendix \ref{sec:repeatability} and their shapes are similar and stable. Thus, the visualization method is valid to characterize the weight loss landscape, based on which, we can carefully investigate the connection between weight loss landscape and robust generalization gap.

\begin{figure}[!t]
\centering
    \subfigure[In the training process of vanilla AT (\emph{Left}: Learning curve; \emph{Mid}: Landscape before ``best''; \emph{Right}: Landscape after ``best'')]{
        \includegraphics[width=0.185\columnwidth]{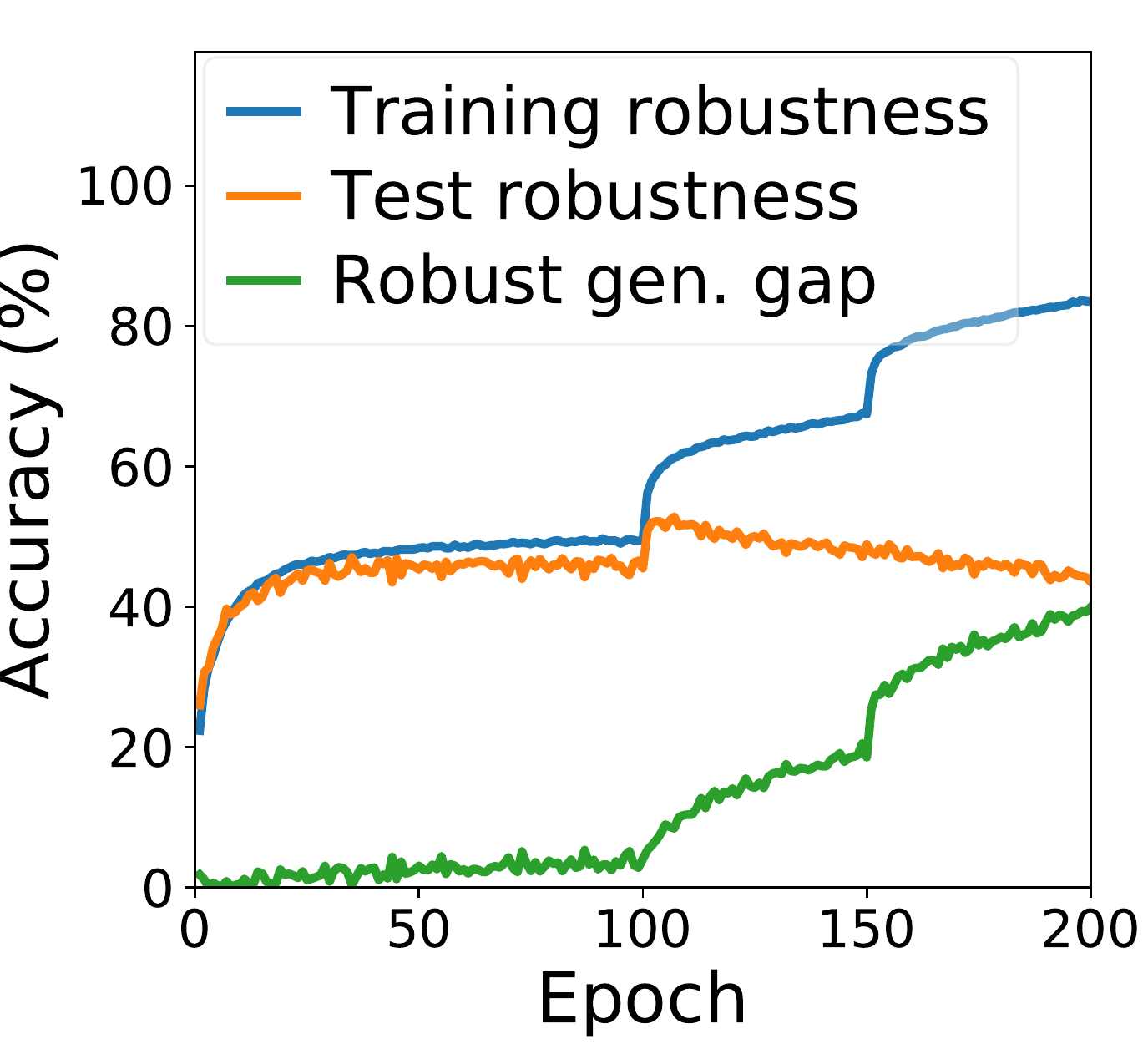}
        \includegraphics[width=0.185\columnwidth]{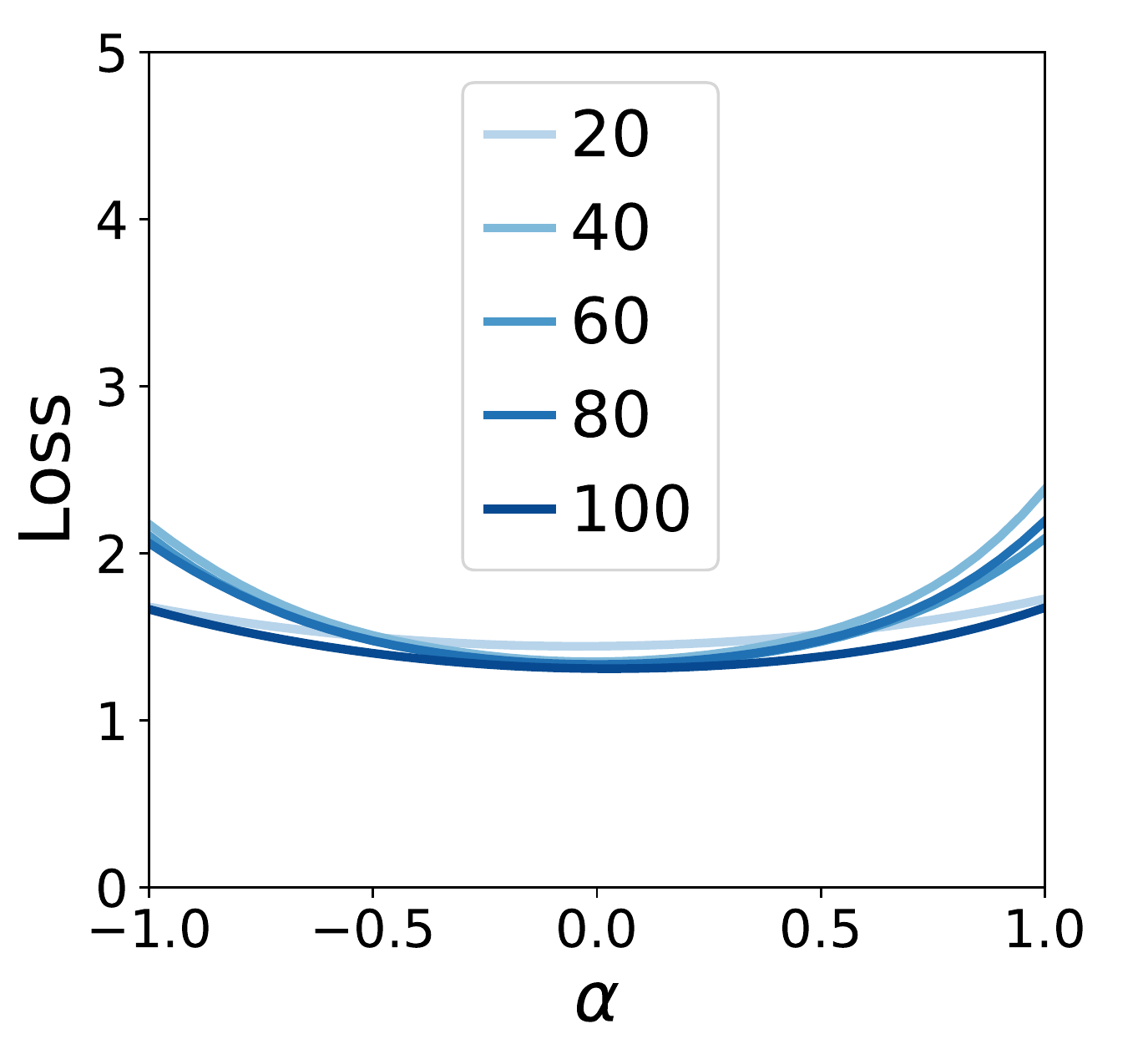}
        \includegraphics[width=0.185\columnwidth]{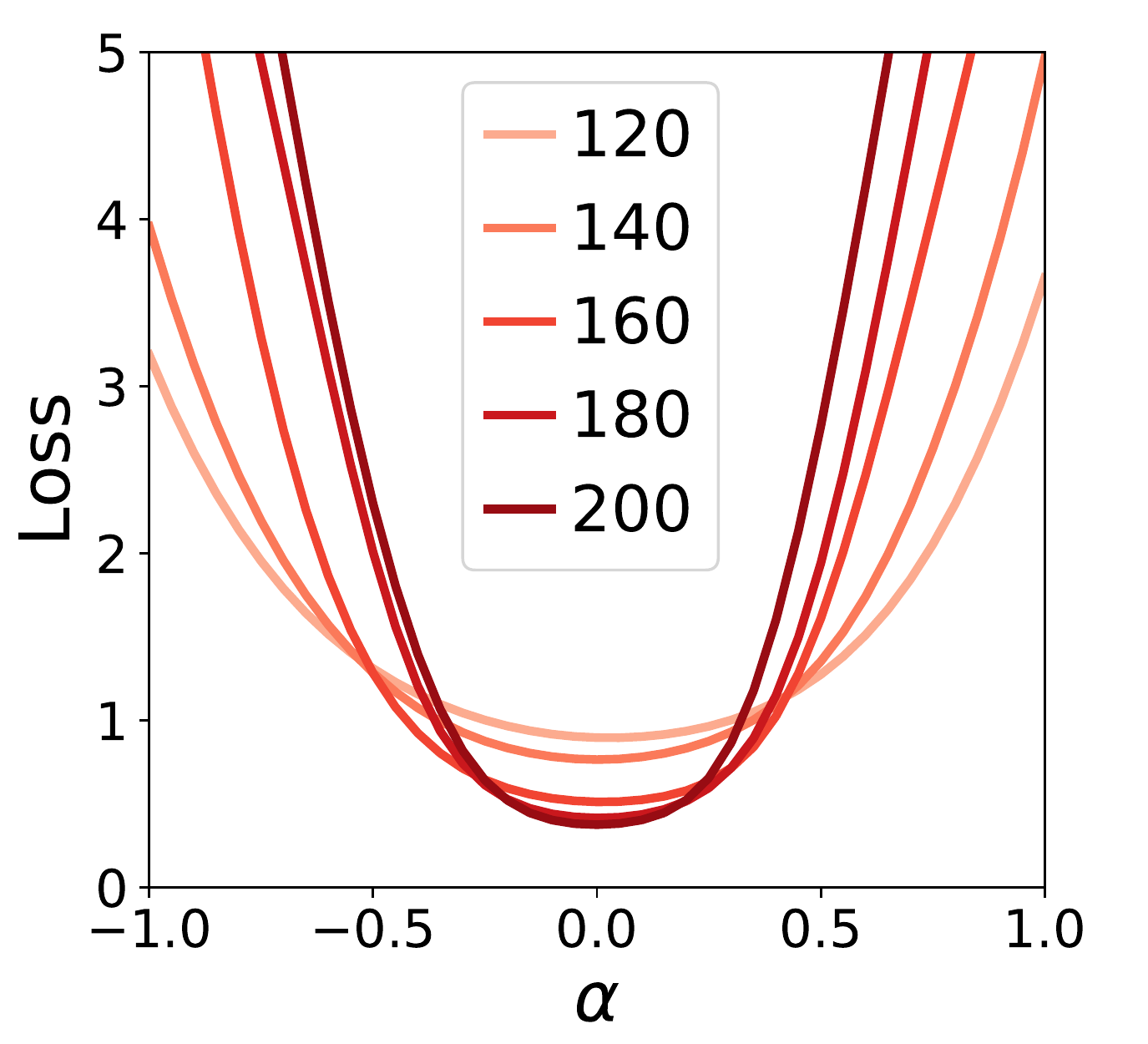}
    } \hfill
    \subfigure[Across different methods (\emph{Left}: Generalization gap; \emph{Right}: Landscape)]{
        \includegraphics[width=0.185\columnwidth]{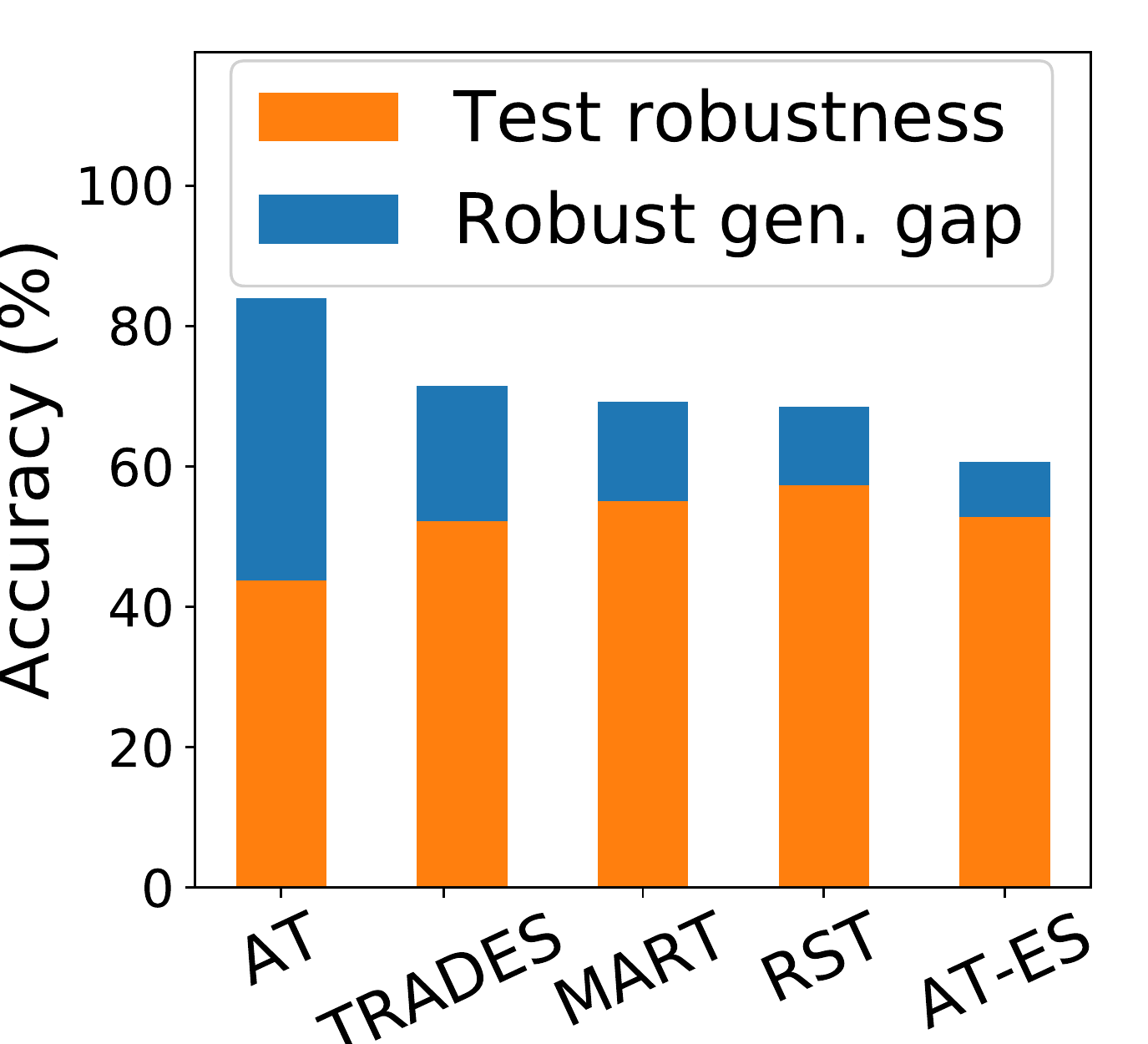}
        \includegraphics[width=0.185\columnwidth]{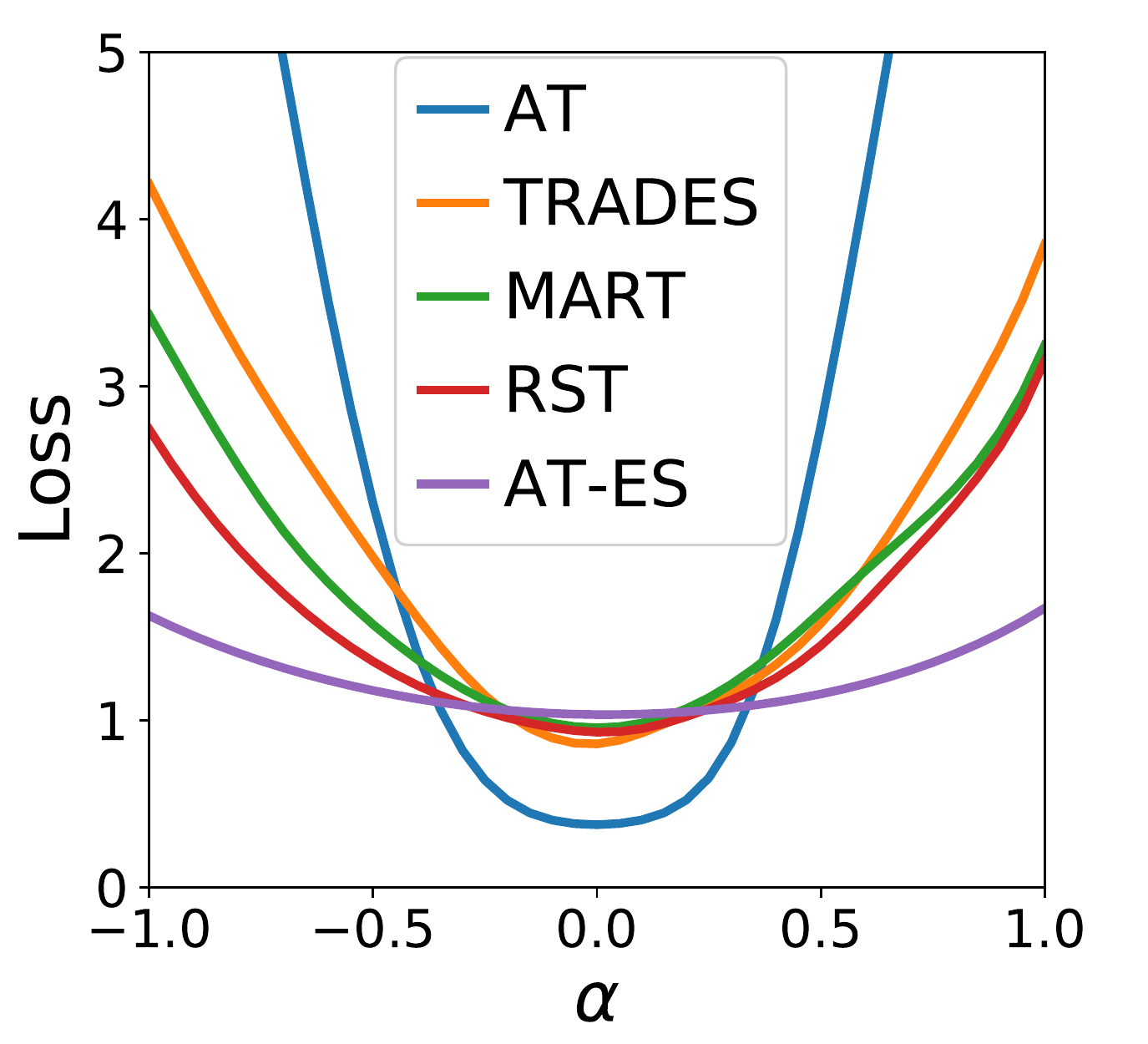}
    }
\vskip -0.1in
\caption{The relationship between weight loss landscape and robust generalization gap is investigated a) in the training process of vanilla AT; and b) across different adversarial training methods on CIFAR-10 using PreAct ResNet-18 and $L_\infty$ threat model. (``Landscape'' is a abbr. of weight loss landscape)}
\vskip -0.1in
\label{fig:motivation_piecewise}
\end{figure}

\textbf{The Connection in the Learning Process of Adversarial Training.} We firstly show how the weight loss landscape changes along with the robust generalization gap in the learning process of adversarial training. We train a PreAct ResNet-18 \cite{he2016identity} on CIFAR-10 for 200 epochs using vanilla AT with a piece-wise learning rate schedule (initial learning rate is 0.1, and divided by 10 at the 100-th and 150-th epoch). The training and test attacks are both 10-step PGD (PGD-10) with step size $2/255$ and maximum $L_\infty$ perturbation $\epsilon = 8/255$. The learning curve and weight loss landscape are shown in Figure \ref{fig:motivation_piecewise}(a) where the ``best'' (highest test robustness) is at the 103-th epoch. Before the ``best'', the test robustness is close to the training robustness, thus the robust generalization gap (green line) is small. Meanwhile, the weight loss landscape (plotted every 20 epochs) before the ``best'' is also very flat. After the ``best'', the robust generalization gap (green line) becomes larger as the training continues, while the weight loss landscape becomes sharper simultaneously. The trends also exist on other model architectures (VGG-19 \cite{simonyan2015very} and WideResNet-34-10 \cite{zagoruyko2016wide}), datasets (SVHN \cite{netzer2011reading} and CIFAR-100 \cite{krizhevsky2009learning}), threat model ($L_2$), and learning rate schedules (cyclic \cite{smith2017cyclical}, cosine \cite{loshchilov2016sgdr}), as shown in Appendix \ref{sec:more_visualization}. Thus, the flatness of weight loss landscape is well-correlated with the robust generalization gap during the training process.

\textbf{The Connection across Different Adversarial Training Methods.} Furthermore, we explore whether the relationship between weight loss landscape and robust generalization gap still exists across different adversarial training methods. Under the same settings as above, we train PreAct ResNet-18 using several state-of-the-art adversarial training methods like TRADES \cite{Zhang2019theoretically}, MART \cite{wang2020improving}, RST \cite{carmon2019unlabeled}, and AT with Early Stopping (AT-ES) \citep{rice2020overfitting}. Figure \ref{fig:motivation_piecewise}(b) demonstrates their training/test robustness and weight loss landscape. Compared with vanilla AT, all methods have a smaller robust generalization gap and a flatter weight loss landscape. Although these state-of-the-art methods improve adversarial robustness using various techniques, they all implicitly flatten the weight loss landscape. It can be also observed that the smaller generalization gap one method achieves, the flatter weight loss landscape it has. This observation is consistent with that in the training process, which verifies that weight loss landscape has a strong correlation with robust generalization gap.

\textbf{Does Flatter Weight Loss Landscape Certainly Lead to Higher Test Robustness?} Revisiting Figure \ref{fig:motivation_piecewise}(b), AT-ES has the flattest weight loss landscape (also the smallest robust generalization gap), but does not obtain the highest test robustness. Since the robust generalization gap is defined as the difference between training and test robustness, the low test robustness of AT-ES is caused by the low training robustness. It indicates that early stopping technique does not make full use of the whole training process, \textit{e.g.}, it stops training around 100-th epoch only with 60\% training robustness which is 20\% lower than that of 200-th epoch. Therefore, a flatter weight loss landscape does directly lead to a smaller robust generalization gap but is only beneficial to the final test robustness on condition that the training process is sufficient (\textit{i.e.}, training robustness is high). 

\textbf{Why Do We Need Weight Loss Landscape?} As aforementioned, adversarial training has already optimized the input loss landscape via training on adversarial examples. However, the adversarial example is generated by injecting input perturbation on each individual example to obtain the highest adversarial loss, which is an example-wise ``local'' worst-case that does not consider the overall effect on 
multiple examples. The weight of DNNs can influence the losses of all examples such that it could be perturbed to obtain a model-wise ``global'' worst-case (highest adversarial loss over multiple examples). Weight perturbations can serve as a good complement for input perturbations. Also, optimizing on perturbed weights (\textit{i.e.}, making the loss remains small even if perturbations are added on the weights) could lead to a flat weight loss landscape, which further will narrow the robust generalization gap. In the next section, we will propose such a weight perturbation for adversarial training.

\section{Proposed Adversarial Weight Perturbation}
In this section, we propose \textit{Adversarial Weight Perturbation (AWP)} to explicitly flatten the weight loss landscape via injecting the worst-case weight perturbation into DNNs. As discussed above, in order to improve the test robustness, we need to focus on both the training robustness and the robust generalization gap (delivered by the flatness of weight loss landscape). 
Thus, we have the objective: 
\begin{equation}
    \min_{\wb} \big\{\rho(\wb) + \big(\rho(\wb + \vb) - \rho(\wb) \big)\big\} \to \min_{\wb} \rho(\wb+\vb),
\end{equation}
where $\rho(\wb)$ is the original adversarial loss in Eq. \eqref{eqn:robust_loss}, $\rho(\wb+\vb) - \rho(\wb)$ is a term to characterize the flatness of weight loss landscape, and $\vb$ is weight perturbation that needs to be carefully selected.

\subsection{Weight Perturbation}\label{sec:weight_direction}
\textbf{Perturbation Direction.} Different from the commonly used random weight perturbation (sampling a random direction) \cite{wen2018flipout,khan2018fast,he2019parametric}, we propose the \textit{Adversarial Weight Perturbation} (AWP), along which the adversarial loss increases dramatically. That is,
\begin{equation}
    \min_{\wb} \max_{\vb \in \mathcal{V}} \rho(\wb+\vb) \rightarrow \min_{\wb} \max_{\vb \in \mathcal{V}} \frac{1}{n} \sum_{i=1}^n \max_{\Vert \xb_i^\prime - \xb_i \Vert_p \leq \epsilon} \ell(\fb_{\wb+\vb}(\xb_i^\prime), y_i),
    \label{eqn:obj}
\end{equation}
where $\mathcal{V}$ is a feasible region for the perturbation $\vb$. Similar to the adversarial input perturbation, AWP also injects the worst-case on weights in a small region around $\fb_\wb$. Note that the maximization on $\vb$ depends on the entire examples (at least the batch examples) to make the whole loss (not the loss on each example) maximal, thus these two maximizations are not exchangeable.

\textbf{Perturbation Size.}\label{sec:weight_size}
Following the weight perturbation direction, we need to determine how much perturbation should be injected. Different from the fixed value constraint $\epsilon$ on adversarial inputs, we restrict the weight perturbation $\vb_l$ using its relative size to the weights of $l$-th layer $\wb_l$:
\begin{equation}
    \Vert \vb_l \Vert \leq \gamma \Vert \wb_l \Vert,
\label{eqn:ineq}
\end{equation}
where $\gamma$ is the constraint on weight perturbation size. The reasons for using relative size to constrain weight perturbation lie on two aspects: 1) the numeric distribution of weights is different from layer to layer, so it is impossible to constrain weights of different layers using a fixed value; and 2) there is scale invariance on weights, \textit{e.g.}, when nonlinear ReLU is used, the network remains unchanged if we multiply the weights in one layer by $10$, and divide by $10$ at the next layer. 

\subsection{Optimization}\label{sec:AWP_optimization}
Once the direction and size of weight perturbation are determined, we propose an algorithm to optimize the double-perturbation adversarial training problem in Eq. \eqref{eqn:obj}. For the two maximization problems,
we circularly generate adversarial example $\xb'_i$ and then update weight perturbation $\vb$ both empirically using PGD\footnote{We find it works well in our experiments. With regard to the results on theoretical measurements or guarantees for the maximization problem like \citet{wang2019convergence}, we leave it for further work.}. The procedure of AWP-based vanilla AT, named AT-AWP, is as follows. 

\textbf{Input Perturbation.} We craft adversarial examples $\xb^\prime$ using PGD attack on $\fb_{\wb + \vb}$:
\begin{equation}
\xb_i^\prime \leftarrow \Pi_\epsilon \big(\xb_i^\prime + \eta_1 \text{sign}(\nabla_{\xb_i^\prime} \ell(\fb_{\wb+\vb}(\xb_i^\prime), y_i)) \big)
\label{eqn:craft_adv_example}
\end{equation}
where $\Pi(\cdot)$ is the projection function and $\vb$ is 0 for the first iteration. 

\textbf{Weight Perturbation.} We calculate the adversarial weight perturbation based on the generated adversarial examples $\xb'$:
\begin{equation}
    \vb \leftarrow \Pi_\gamma \big(\vb + \eta_2  \frac{\nabla_\vb \frac{1}{m} \sum_{i=1}^m \ell(\fb_{\wb + \vb}(\xb_i^\prime), y_i)}{\Vert \nabla_\vb \frac{1}{m} \sum_{i=1}^m \ell(\fb_{\wb+\vb}(\xb_i^\prime), y_i) \Vert} \Vert \wb \Vert \big),
\label{eqn:nu_direction}
\end{equation}
where $m$ is the batch size, and $\vb$ is layer-wise updated (refer to Appendix \ref{sec:awp_code} for details). 
Similar to generating adversarial examples $\xb'$ via FGSM (one-step) or PGD (multi-step), $\vb$ can also be solved by one-step or multi-step methods. 
Then, we can alternately generate $\xb^\prime$ and calculate $\vb$ for a number of iterations $A$. As shortly will be shown in Section \ref{sec:exploring}, one iteration for $A$ and one-step for $\vb$ (default settings) are enough to get good robustness improvements. 

\textbf{Model Training.} Finally, we update the parameters of the perturbed model $\fb_{\wb+\vb}$ using SGD. Note that after optimizing the loss of a perturbed point on the landscape, we should come back to the center point again for the next start. Thus, the actual parameter update follows:
\begin{equation}\label{eqn:w_update}
    \wb \leftarrow (\wb+\vb) - \eta_3 \nabla_{\wb+\vb} \frac{1}{m}\sum_{i=1}^m \ell(\fb_{\wb+\vb}(\xb_i^\prime, y_i)) - \vb.
\end{equation}
The complete pseudo-code of AT-AWP and extensions of AWP to other adversarial training approaches like TRADES, MART and RST are shown in Appendix \ref{sec:awp_code}. 

\subsection{Theoretical Analysis}

We also provide a theoretical view on why AWP works. Based on previous work on PAC-Bayes bound \cite{neyshabur2017exploring}, in adversarial training, let $\ell(\cdot, \cdot)$ be 0-1 loss, then $\rho(\mathbf{w}) = \frac{1}{n} \sum_{i=1}^n \max_{\Vert \mathbf{x}_i^\prime - \mathbf{x}_i \Vert_p \leq \epsilon} \ell(f_{\mathbf{w}}(\mathbf{x}_i^\prime), y_i) \in [0, 1]$. Given a ``prior'' distribution $P$ (a common assumption is zero mean, $\sigma^2$ variance Gaussian distribution) over the weights, the expected error of the classifier can be bounded with probability at least $1 - \delta$ over the draw of $n$ training data: 
\begin{equation}
\label{robust_gap}
    \mathbb{E}_{\{\mathbf{x}_i, y_i\}_{i=1}^n, \mathbf{u}}[\rho(\mathbf{w+u}) ] \leq
    \rho(\mathbf{w}) + \big\{ \mathbb{E}_{\mathbf{u}}[ \rho(\mathbf{w+u})] - \rho(\mathbf{w})\big\} + 4 \sqrt{\frac{1}{n} KL (\mathbf{w+u} \Vert P) + \ln\frac{2n}{\delta}}.
\end{equation}
Following \citet{neyshabur2017exploring}, we choose $\mathbf{u}$ as a zero mean spherical Gaussian perturbation with variance $\sigma^2$ in every direction, 
and set the variance of the perturbation to the weight with respect to its magnitude $\sigma = \alpha \Vert \mathbf{w} \Vert$, which makes the third term of Eq. \eqref{robust_gap} become a constant $4\sqrt{\frac{1}{n} (\frac{1}{2\alpha} + \ln{\frac{2n}{\delta}})}$. Thus, the robust generalization gap is bounded by the second term that is the expectation of the flatness of weight loss landscape. Considering the optimization efficiency and effectiveness on expectation, $\mathbb{E}_{\mathbf{u}}[ \rho(\mathbf{w+u})] \leq \max_{\mathbf{u}}[ \rho(\mathbf{w+u})]$. AWP exactly optimizes the worst-case of the flatness of weight loss landscape $\{ \max_{\mathbf{u}}[
\rho(\mathbf{w+u})] - \rho(\mathbf{w})\}$ to control the above PAC-Bayes bound, which theoretically justifies why AWP works. 

\subsection{A Case Study on Vanilla AT and AT-AWP}
\label{sec:a_case_study}
In this part, we conduct a case study on vanilla AT and AT-AWP across three benchmark datasets (SVHN \cite{netzer2011reading}, CIFAR-10 \cite{krizhevsky2009learning}, CIFAR-100 \cite{krizhevsky2009learning}) and two threat models ($L_\infty$ and $L_2$) using PreAct ResNet-18 for 200 epochs. We follow the same settings in \citet{rice2020overfitting}: for $L_\infty$ threat model, $\epsilon = 8/255$, step size is $1/255$ for SVHN, and $2/255$ for CIFAR-10 and CIFAR-100; for $L_2$ threat model, $\epsilon = 128/255$, step size is $15/255$ for all datasets. The training/test attacks are PGD-10/PGD-20 respectively. For AT-AWP, $\gamma = 1 \times 10^{-2}$.
The test robustness is reported in Table \ref{table:various_dataset_norm} (natural accuracy is in Appendix \ref{sec:case_study}) where ``best'' means the highest robustness that ever achieved at different checkpoints for each dataset and threat model while ``last'' means the robustness at the last epoch checkpoint. We can see that AT-AWP consistently improves the test robustness for all cases. It indicates that AWP is generic and can be applied on various threat models and datasets. 

\begin{table}[!htbp]
\caption{Test robustness (\%) of AT and AT-AWP across different datasets and threat models. We omit the standard deviations of 5 runs as they are very small ($< 0.40 \%$), which hardly effect the results.}
\label{table:various_dataset_norm}
\centering
\small
\vspace{-0.1 in}
\begin{tabular}{clcccccc}
\toprule
\multirow{2}{*}{Threat Model} & \multirow{2}{*}{Method} & \multicolumn{2}{c}{SVHN} & \multicolumn{2}{c}{CIFAR-10} & \multicolumn{2}{c}{CIFAR-100} \\
\cmidrule(lr){3-4} \cmidrule(lr){5-6} \cmidrule(lr){7-8} & & Best & Last & Best & Last & Best & Last \\ \midrule
\multirow{2}{*}{$L_\infty$} & AT & 53.36 & 44.49 & 52.79 & 44.44 & 27.22 & 20.82 \\
 & AT-AWP & \textbf{59.12} & \textbf{55.87} & \textbf{55.39} & \textbf{54.73} & \textbf{30.71} & \textbf{30.28} \\ \midrule
\multirow{2}{*}{$L_2$}      & AT & 66.87 & 65.03 & 69.15 & 65.93  & 41.33 & 35.27         \\
 & AT-AWP & \textbf{72.57} & \textbf{67.73} & \textbf{72.69} & \textbf{72.08} & \textbf{45.60} & \textbf{44.66}
\\ \bottomrule
\end{tabular}
\vspace{-0.1 in}
\end{table}

\section{Experiments}
In this section, we conduct comprehensive experiments to evaluate the effectiveness of AWP including its benchmarking robustness, ablation studies and comparisons to other regularization techniques.

\subsection{Benchmarking the State-of-the-art Robustness}\label{benchmark}
In this part, we evaluate the robustness of our proposed AWP on CIFAR-10 to benchmark the state-of-the-art robustness against white-box and black-box attacks. Two types of adversarial training methods are considered here: One is only based on original data: 1) AT \cite{madry2018towards}; 2) TRADES \cite{Zhang2019theoretically}; and 3) MART \cite{wang2020improving}. The other uses additional data: 1) Pre-training \cite{hendrycks2019using}; and 2) RST \cite{carmon2019unlabeled}.

\begin{table*}[!t]
\caption{Test robustness (\%) on CIFAR-10 using WideResNet under $L_\infty$ threat model. We omit the standard deviations of 5 runs as they are very small ($< 0.40 \%$), which hardly effect the results.}
\label{table:white_box}
\centering
\footnotesize
\begin{tabular}{lccccc|c|c}
\toprule
Defense & Natural & FGSM & PGD-20 & PGD-100 & CW$_\infty$ & SPSA & AA\\ \midrule
AT & \textbf{86.07} & 61.76 & 56.10 & 55.79 & 54.19 & 61.40 & 52.60 \footnotemark[5] \\
AT-AWP & 85.57 & \textbf{62.90} & \textbf{58.14}  & \textbf{57.94} & \textbf{55.96}& \textbf{62.65} & \textbf{54.04}  \\ \midrule

TRADES & 84.65 & 61.32 & 56.33 & 56.07 & 54.20 & 61.10 & 53.08 \\
TRADES-AWP & \textbf{85.36} & \textbf{63.49} & \textbf{59.27} & \textbf{59.12} & \textbf{57.07} & \textbf{63.85} & \textbf{56.17} \\ \midrule

MART & 84.17 & 61.61 & 58.56 & 57.88 & 54.58 & 58.90 & 51.10 \\ 
MART-AWP & \textbf{84.43} & \textbf{63.98} & \textbf{60.68} & \textbf{59.32} & \textbf{56.37} & \textbf{62.75} & \textbf{54.23} \\ 
\midrule
Pre-training  & 87.89 & 63.27 & 57.37 & 56.80 & 55.95 & 62.55 & 54.92 \\
Pre-training-AWP & \textbf{88.33} & \textbf{66.34} & \textbf{61.40} & \textbf{61.21} & \textbf{59.28} & \textbf{65.55} & \textbf{57.39} \\ \midrule
RST  & \textbf{89.69} & \textbf{69.60} & 62.60 & 62.22 & 60.47 & 67.60 & 59.53  \\
RST-AWP & 88.25 & 67.94 & \textbf{63.73} & \textbf{63.58} & \textbf{61.62} & \textbf{68.72} & \textbf{60.05} \\  
\bottomrule
\end{tabular}
\vspace{-0.15 in}
\end{table*}
\footnotetext[5]{Here is the result on WideResNet-34-10 while the leaderborder one is on WideResNet-34-20.}

\textbf{Experimental Settings.} 
For CIFAR-10 under $L_{\infty}$ attack with $\epsilon=8/255$, we train WideResNet-34-10 for AT, TRADES, and MART, while WideResNet-28-10 for Pre-training and RST, following their original papers. For pre-training, we fine-tune 50 epochs using a learning rate of 0.001 as \cite{hendrycks2019using}. Other defenses are trained for 200 epochs using SGD with momentum 0.9, weight decay $5 \times 10^{-4}$, and an initial learning rate of 0.1 that is divided by 10 at the 100-th and 150-th epoch. Simple data augmentations such as $32 \times 32$ random crop with 4-pixel padding and random horizontal flip are applied. The training attack is PGD-10 with step size $2/255$. For AWP, we set $\gamma = 5\times10^{-3}$. Other hyper-parameters of the baselines are configured as per their original papers.

\textbf{White-box/Black-box Robustness.}
Table \ref{table:white_box} reports the ``best'' test robustness (the highest robustness ever achieved at different checkpoints for each defense against each attack) against white-box and black-box attacks. ``Natural'' denotes the accuracy on natural test examples. First, for white-box attack, we test FGSM, PGD-20/100, and CW$_{\infty}$ ($L_{\infty}$ version of CW loss optimized by PGD-100). AWP
almost improves the robustness of state-of-the-art methods against all types of attacks. This is because AWP aims at achieving a flat weight loss landscape, which is generic across different methods. Second, for black-box attack, we test the query-based 
attack SPSA \cite{uesato2018adversarial} (100 iterations with perturbation size 0.001 (for gradient estimation), learning rate 0.01, and 256 samples for each gradient estimation). Again, the robustness improved by AWP is consistent amongst different methods.
In addition, we test AWP against Auto Attack (AA) \cite{croce2020reliable}, which is a strong and reliable attack to verify the robustness via an ensemble of diverse parameter-free attacks including three white-box attacks (APGD-CE \cite{croce2020reliable}, APGD-DLR \cite{croce2020reliable}, and FAB \cite{croce2019minimally}) and a black-box attack (Square Attack \cite{ACFH2019square}). Compared with their leaderboard results\footnote{\url{https://github.com/fra31/auto-attack}}, AWP can further boost their robustness, ranking the 1st on both with and without additional data. Even some AWP methods without additional data can surpass the results under additional data\footnote{\url{https://github.com/csdongxian/AWP/tree/main/auto_attacks}}. 
This verifies that AWP improves adversarial robustness reliably rather than improper tuning of hyper-parameters of attacks, gradient obfuscation or masking.

\subsection{Ablation Studies on AWP}
\label{sec:exploring}
In this part, we delve into AWP to investigate its each component. We train PreAct ResNet-18 using vanilla AT and AT-AWP with $L_{\infty}$ threat model with $\epsilon=8/255$ for 200 epochs following the same setting in Section \ref{benchmark}. The training/test attacks are PGD-10/PGD-20 (step size $2/255$) respectively.

\begin{figure*}[!htbp]
\centering
    \subfigure[Optimization]{
        \includegraphics[width=0.23\columnwidth]{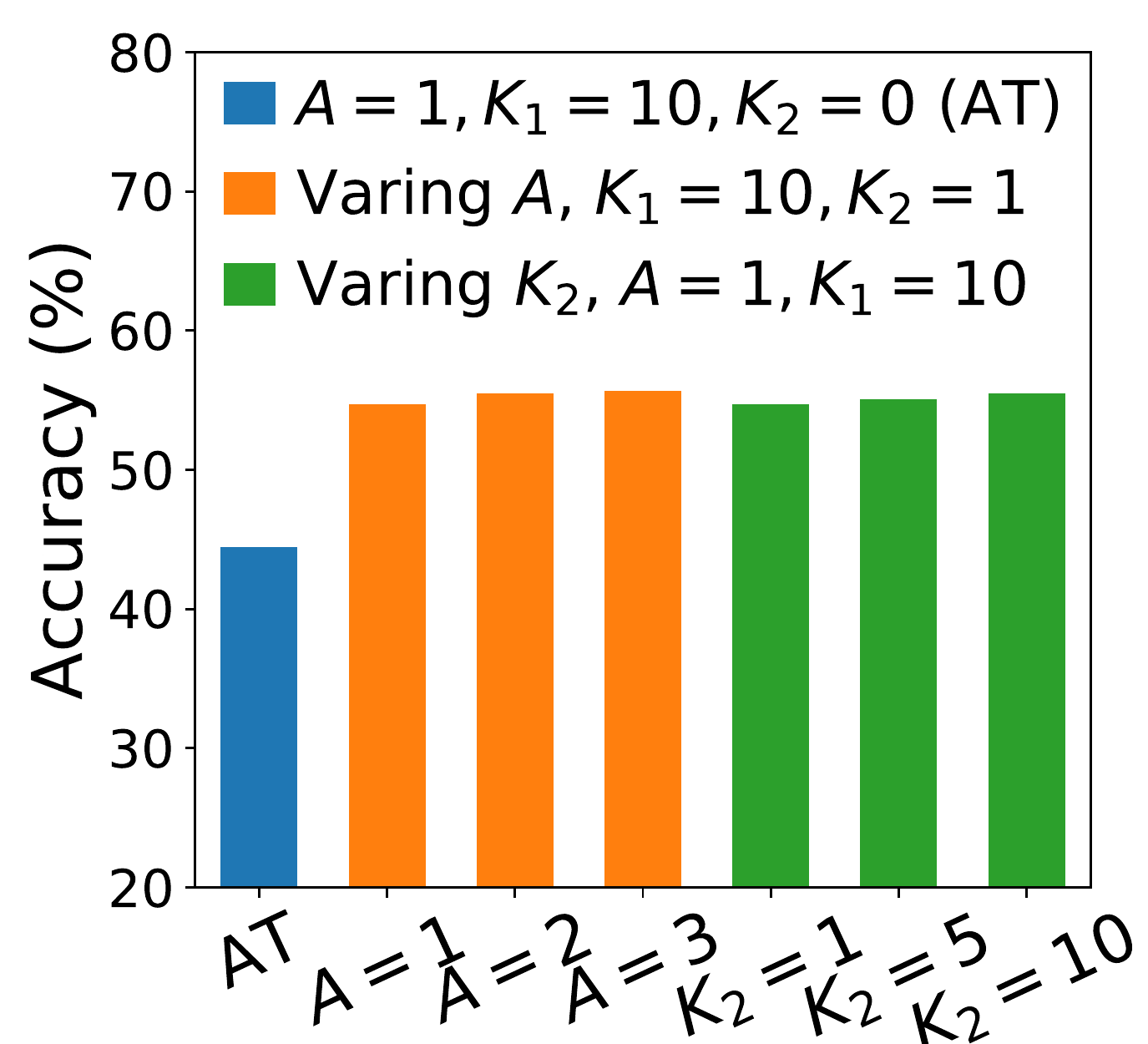}
    }
    \subfigure[Weight perturbation]{
        \includegraphics[width=0.23\columnwidth]{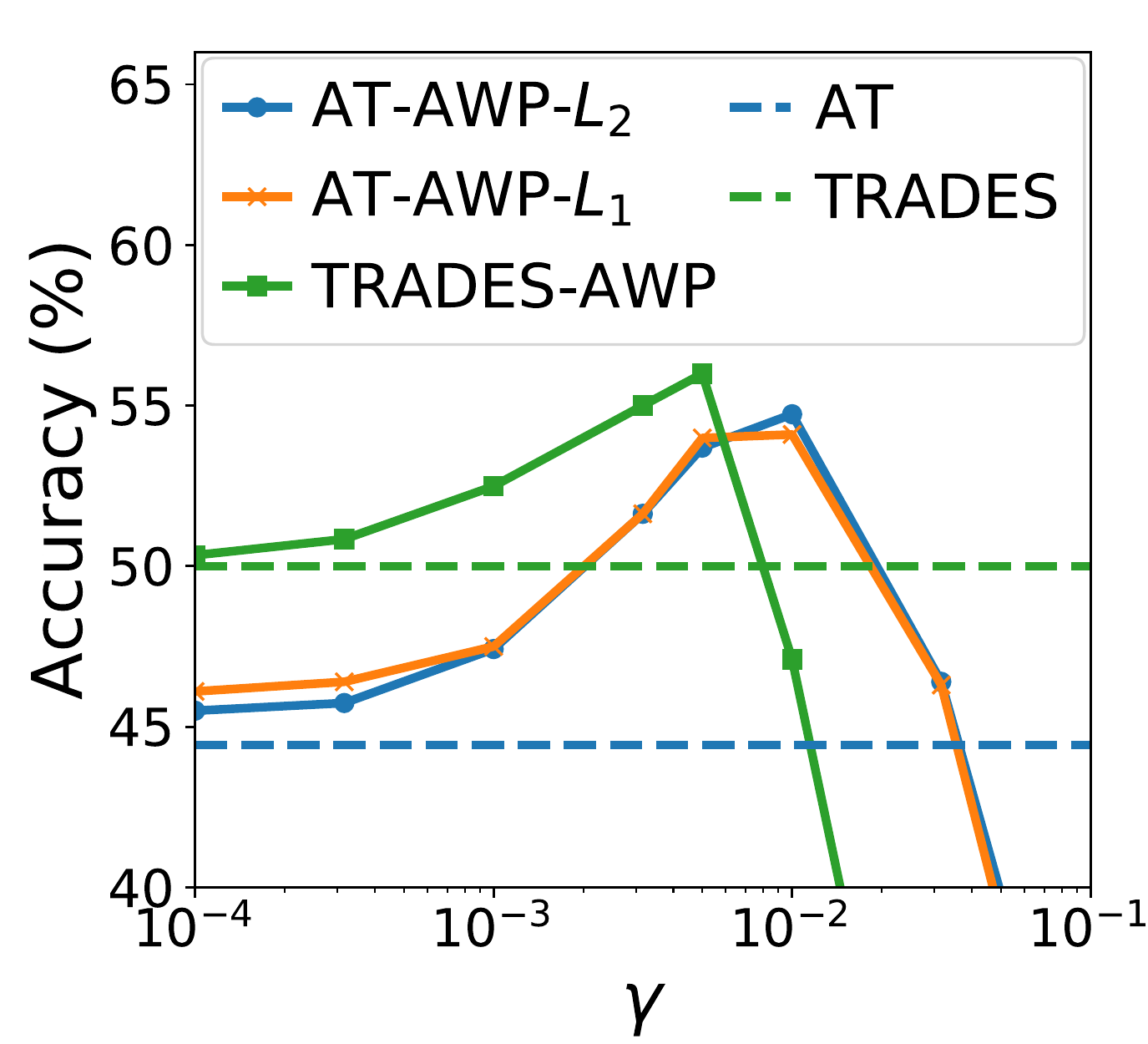}
    }
    \subfigure[Weight loss landscape]{
        \includegraphics[width=0.23\columnwidth]{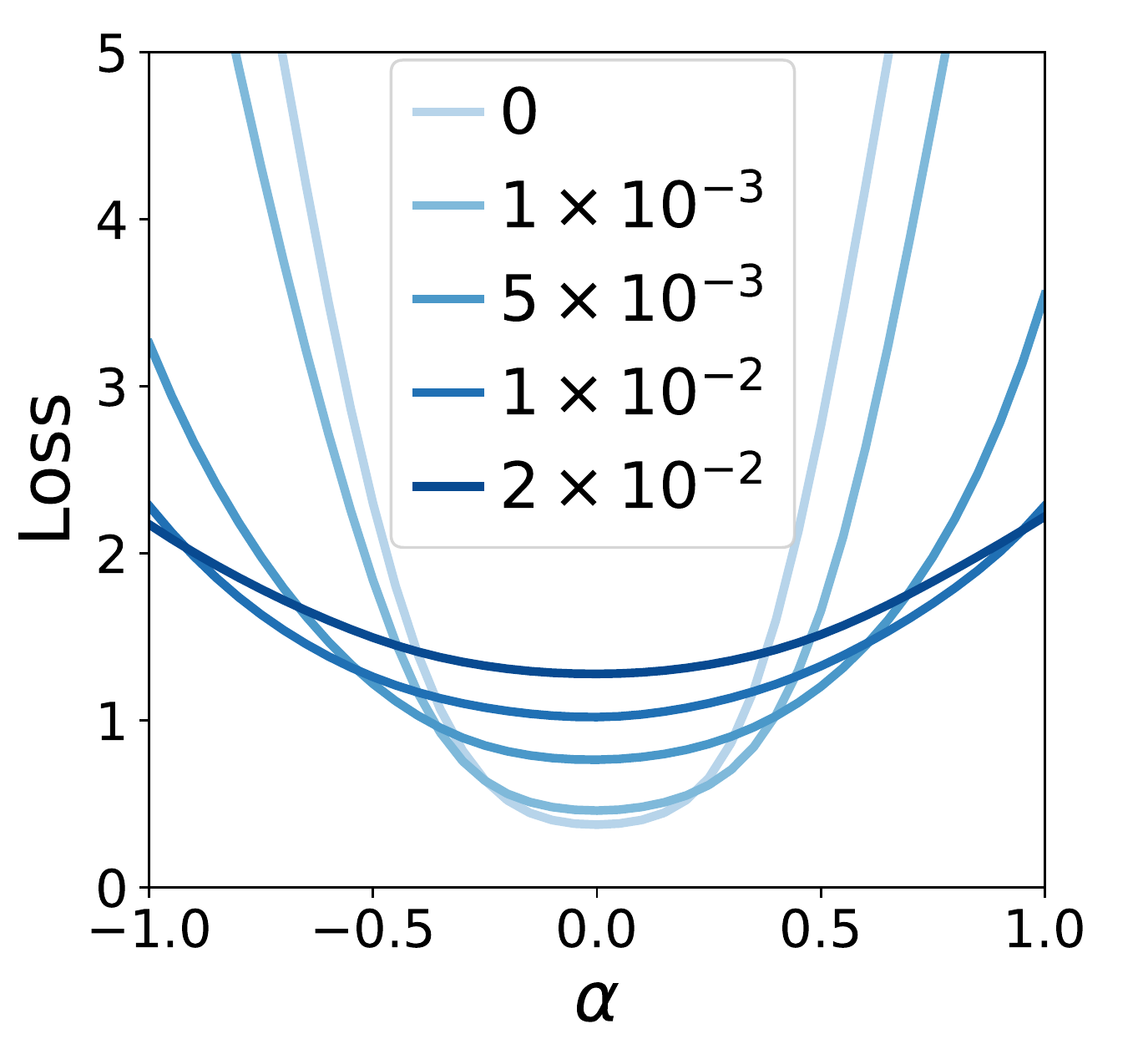}
    }
    \subfigure[Generalization gap]{
	    \includegraphics[width=0.23\columnwidth]{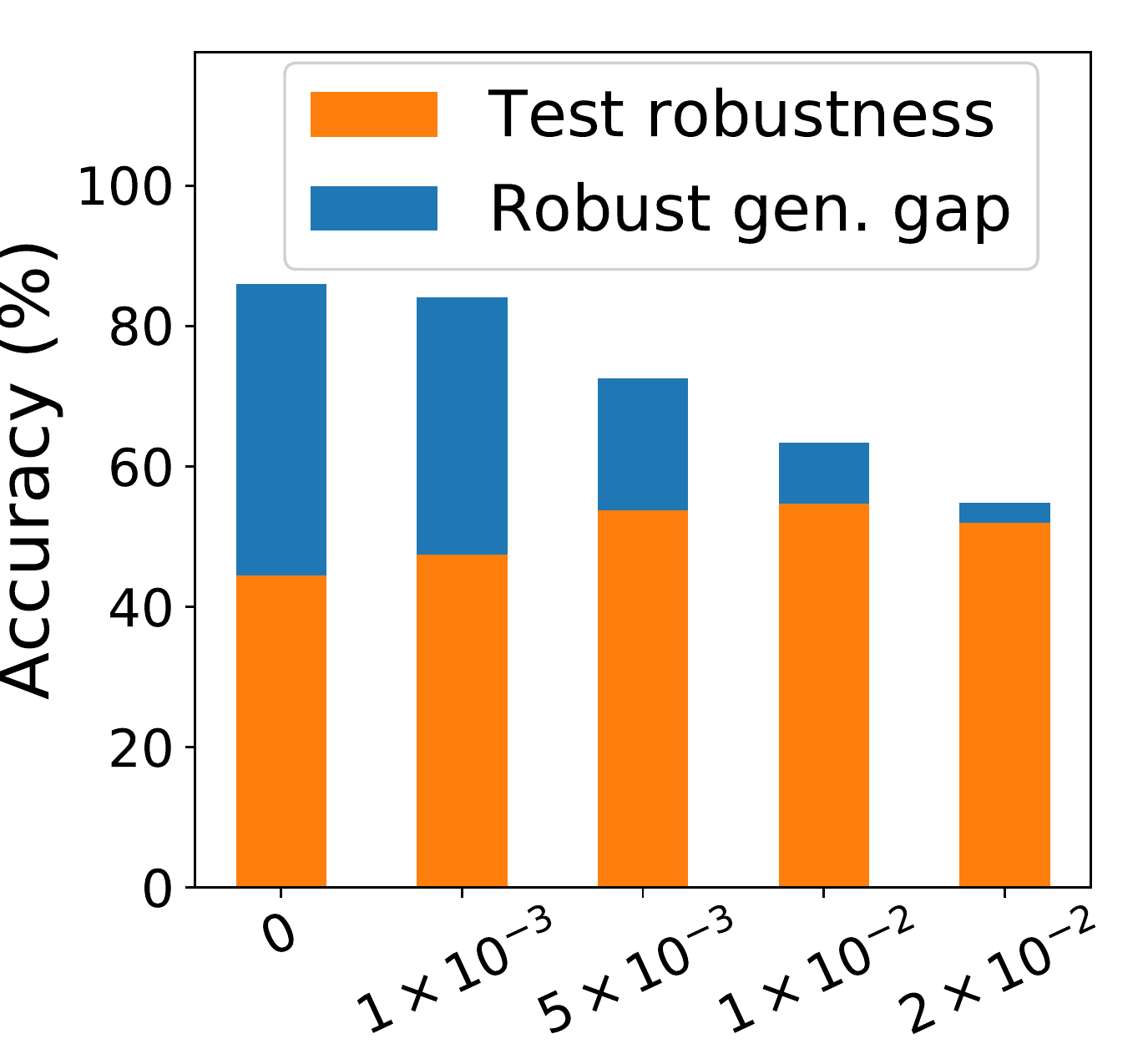}
    }
\vspace{-0.1 in}
\caption{The ablation study experiments on CIFAR-10 using AT-AWP unless otherwise specified.}
\label{fig:understanding}
\vspace{-0.05 in}
\end{figure*}

\textbf{Analysis on Optimization Strategy.} Recalling Section \ref{sec:AWP_optimization}, there are 3 parameters when optimizing AWP, \textit{i.e.}, step number $K_1$ in generating adversarial example $\xb'$, step number $K_2$ in solving adversarial weight perturbation $\vb$, and alternation iteration $A$ between $\xb'$ and $\vb$. For step number $K_1$ in generating $\xb'$, previous work has showed that PGD-10 based AT usually obtains good robustness \cite{wang2019convergence}, so we set $K_1 = 10$ by default. For step number $K_2$ in solving $\vb$, we assess AT-AWP with $K_2 \in \{1,5,10\}$ while keeping $A=1$. The green bars in Figure \ref{fig:understanding}(a) show that varying $K_2$ achieves almost the same test robustness. 
For alternation iteration $A$, we test $A \in \{1,2,3 \}$ while keeping $K_2 = 1$. The orange bars show that one iteration ($A=1$) already has 55.39\% test robustness, and extra iterations only bring few improvements but with much overhead. Based on these results, the default setting for AWP is $A=1, K_1 = 10, K_2=1$ whose training time overhead is $\sim 8\%$.

\textbf{Analysis on Weight Perturbation.} Here, we explore the effect of weight perturbation size (direction will be analyzed in Section \ref{sec:other_reg}) from two aspects: size constraint $\gamma$ and size measurement norm. The test robustness with varying $\gamma$ on AT-AWP and TRADES-AWP are shown in Figure \ref{fig:understanding}(b). We can see that both methods can achieve notable robustness improvements in a certain range $\gamma \in [1 \times 10^{-3}, 5 \times 10^{-3}]$. It implies that the perturbation size cannot be too small to ineffectively regularize the flatness of weight loss landscape and also cannot be too large to make DNNs hard to train. 
Once $\gamma$ is properly selected, it has a relatively good transferability across different methods (improvements of AT-AWP and TRADES-AWP have an overlap on $\gamma$, though their highest points are not the same). As for the size measurement norm, $L_1$ and $L_2$ (also called Frobenius norm $L_F$) almost have no difference on test robustness. 

\textbf{Effect on Weight Loss Landscape and Robust Generalization Gap.} We visualize the weight loss landscape of AT-AWP with different $\gamma$ in Figure \ref{fig:understanding}(c) and present its corresponding training/test robustness in Figure \ref{fig:understanding}(d). The gray line of $\gamma = 0$ is the vanilla AT (without AWP). As $\gamma$ grows, the regularization becomes stronger, thus the weight loss landscape becomes flatter. Accordingly, the robust generalization gap becomes smaller. This verifies that AWP indeed brings flatter weight loss landscape and smaller robust generalization gap. In addition, 
the flattest weight loss landscape (smallest robust generalization gap) is obtained at a large $\gamma = 2\times10^{-2}$ but its training/test robustness decreases, which implies that $\gamma$ should be properly selected by balancing the training robustness and the flatness of weight loss landscape to obtain the test robustness improvement. 

\subsection{Comparisons to Other Regularization Techniques}\label{sec:other_reg}

In this part, we compare AWP with other regularizations using the same setting as Section \ref{sec:exploring}.

\begin{figure*}[!t]
\vspace{-0.1 in}
\centering
    \subfigure[Loss curve]{
        \includegraphics[width=0.23\columnwidth]{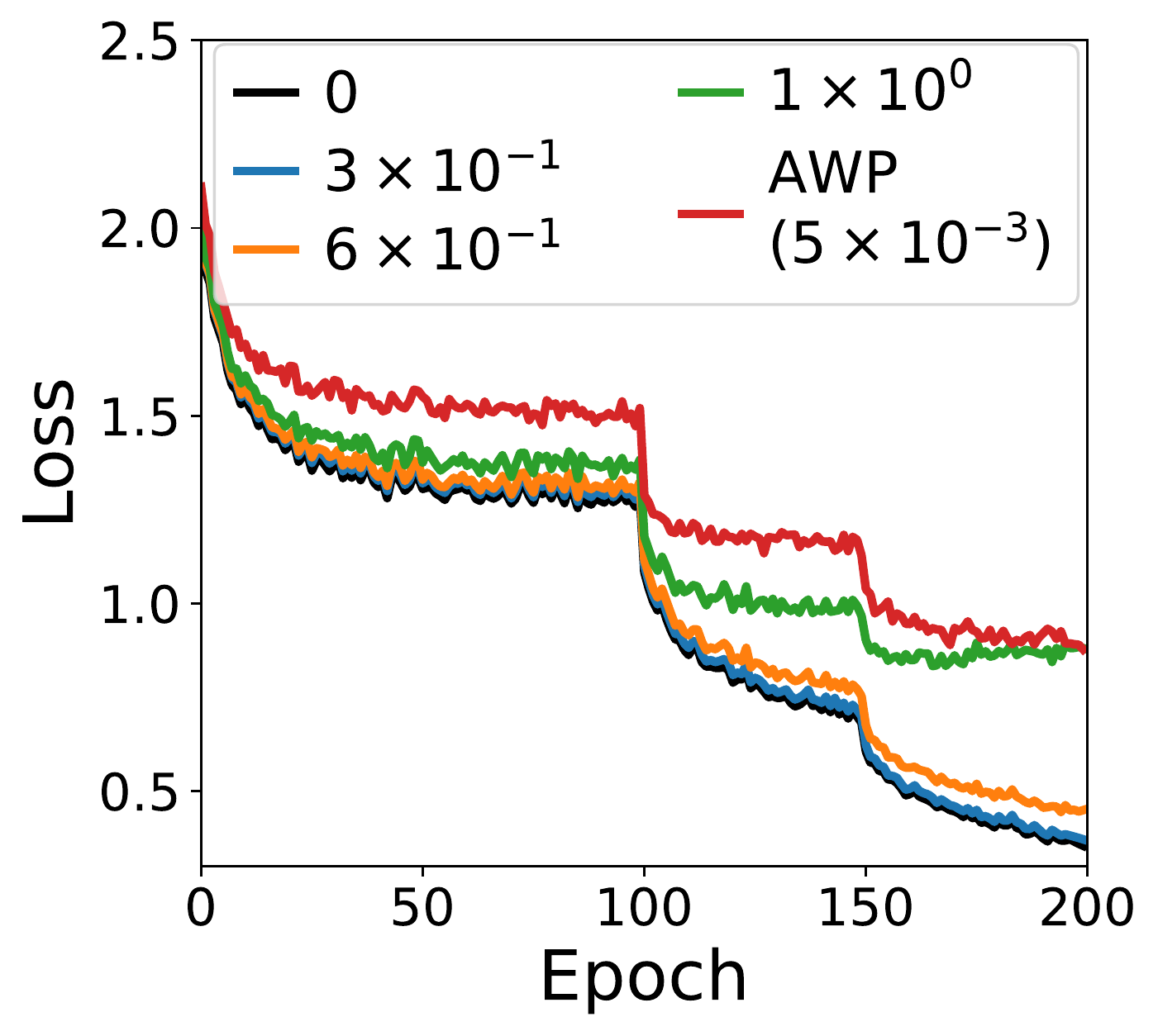}}
    \subfigure[Weight loss landscapes]{
        \includegraphics[width=0.23\columnwidth]{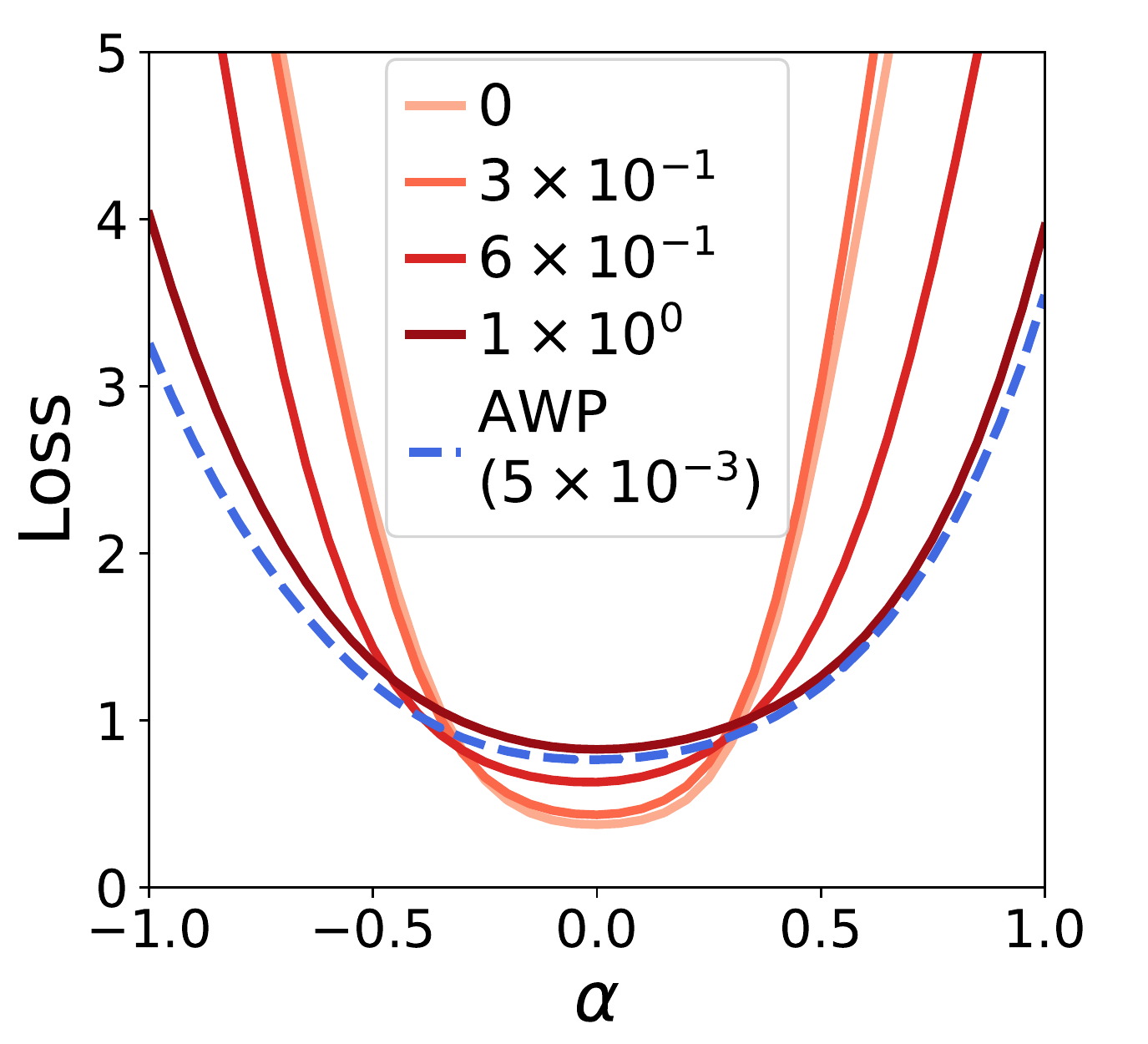}}
    \subfigure[Robustness]{
        \includegraphics[width=0.23\columnwidth]{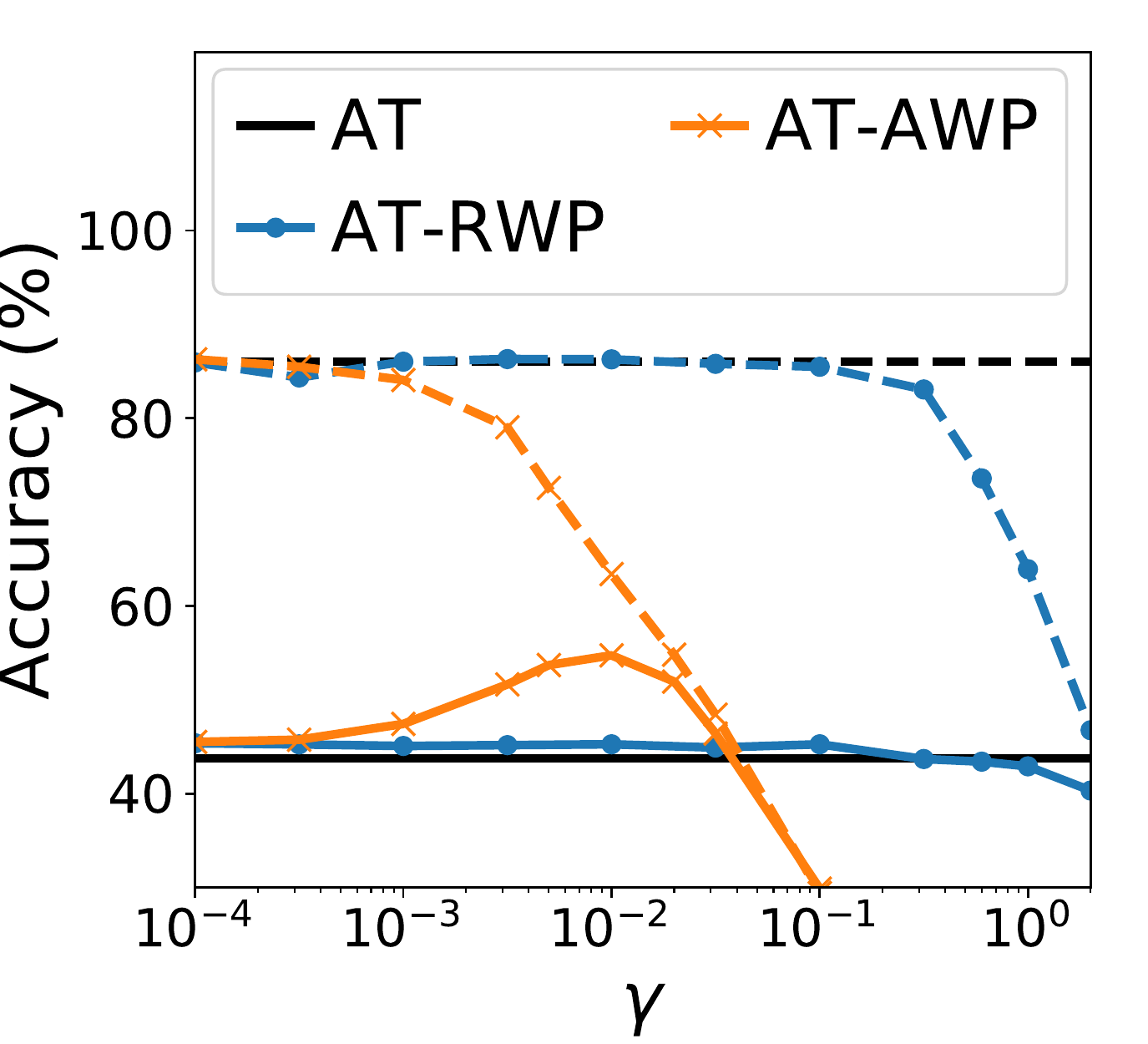}}
    \subfigure[Learning curve]{
        \includegraphics[width=0.23\columnwidth]{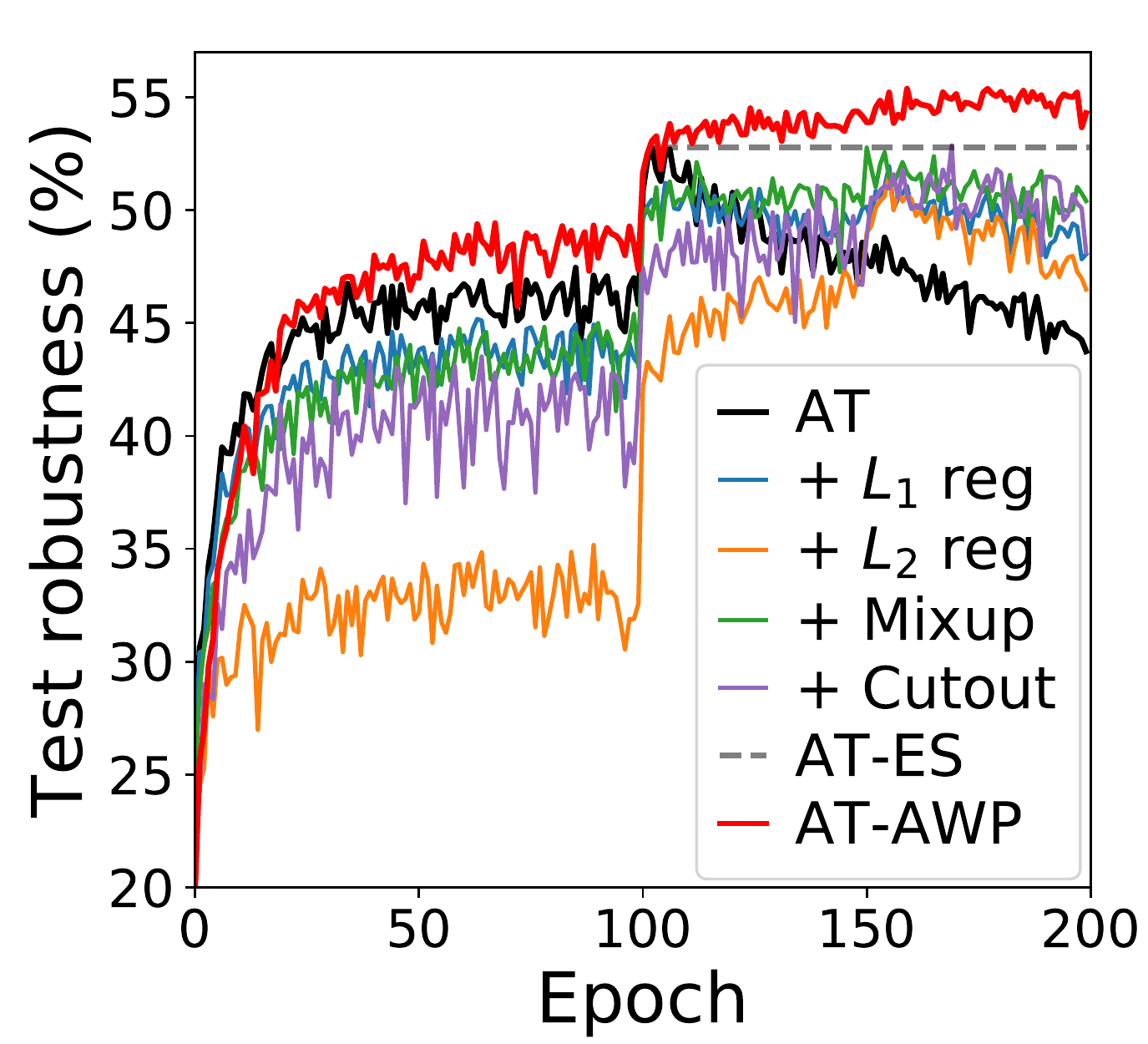}
    }
\vspace{-0.1 in}
\caption{Comparisons of AWP and other regularization techniques (the values in (a)/(c) legend are $\gamma$ in RWP unless otherwise specified) on CIFAR-10 using PreAct ResNet-18 and $L_\infty$ threat model.}
\label{fig:compariosn_other_reg}
\vspace{-0.18 in}
\end{figure*}

\textbf{Comparison to Random Weight Perturbation (RWP).} \label{sec:random_perturbation}
We evaluate the difference of AWP and RWP from the following 3 views: 1) Adversarial loss of AT pre-trained model perturbed by RWP and AWP. As shown in Figure \ref{fig:compariosn_other_reg}(a), RWP only has an obvious increase of adversarial loss at a extremely large $\gamma = 1$ (others are similar to pre-trained AT ($\gamma = 0$)), while AWP (red line) has much higher adversarial loss than others just using a very small perturbation ($\gamma=5\times 10^{-3}$). Therefore, AWP can find the worst-case perturbation in a small region while RWP needs a relatively large perturbation. 
2) Weight loss landscape of models trained by AT-RWP and AT-AWP. As shown in Figure \ref{fig:compariosn_other_reg}(b), RWP only flattens the weight loss landscape at a large $\gamma \ge 0.6$. Even, RWP under $\gamma = 1$ can only obtain a similar flatter weight loss landscape as AWP under $\gamma=5\times 10^{-3}$. 3) Robustness. We test AT-AWP and AT-RWP with a large range $\gamma \in [1 \times 10^{-4}, 2.0]$. Figure \ref{fig:compariosn_other_reg}(c) (solid/dashed lines are test/training robustness respectively) shows that AWP can significantly improve the test robustness at a small $\gamma \in [1\times 10^{-3}, 1\times 10^{-2}]$. For RWP, the test robustness almost does not improve at $\gamma \le 0.3$ because of the unchanged weight loss landscape, even begins to decrease when $\gamma \ge 0.6$. This is because such a large weight perturbation makes DNNs hard to train and severely reduces the training robustness (dashed blue line), which in turns reduces the test robustness though the weight loss landscape is flattened. In summary, AWP is much better than RWP for weight perturbation.

\textbf{Comparison to Weight Regularization and Data Augmentation.} Here, we compare AWP ($\gamma=5\times 10^{-3}$) with $L_1$/$L_2$ weight regularization and data augmentation of mixup \cite{zhang2018mixup}/cutout \cite{devries2017improved}. We follow the \textit{best} hyper-parameters tuned in \citet{rice2020overfitting}: $\lambda = 5 \times 10^{-6}/5 \times 10^{-3}$ for $L_1/L_2$ regularization respectively, patch length $14$ for cutout, and $\alpha = 1.4$ for mixup. We show the test robustness (natural accuracy is in Appendix \ref{sec:cifar10}) of all checkpoints for different methods in Figure \ref{fig:compariosn_other_reg}(d). The vanilla AT achieves the best robustness after the first learning rate decay and starts overfitting. Other techniques, except of AWP, do not obtain a better robustness than early stopped AT (AT-ES), which is consistent with the observations in \citet{rice2020overfitting}. 
However, AWP (red line) behaves very differently from the others: it does improve the best robustness (52.79\% of vanilla AT $\rightarrow$ 55.39\% of AT-AWP). 
AWP shows its superiority over other weight regularization and data augmentation, and improves the best robustness further compared with early stopping. More experiments under $L_2$ threat model could be found in Appendix \ref{sec:cifar10}, which also demonstrates the effectiveness of AWP.

\subsection{A Closer Look at the Weights Learned by AWP}

\begin{wrapfigure}[7]{r}{0.30\textwidth}
\vspace{-0.15 in}
\centering
\includegraphics[width=0.23\textwidth]{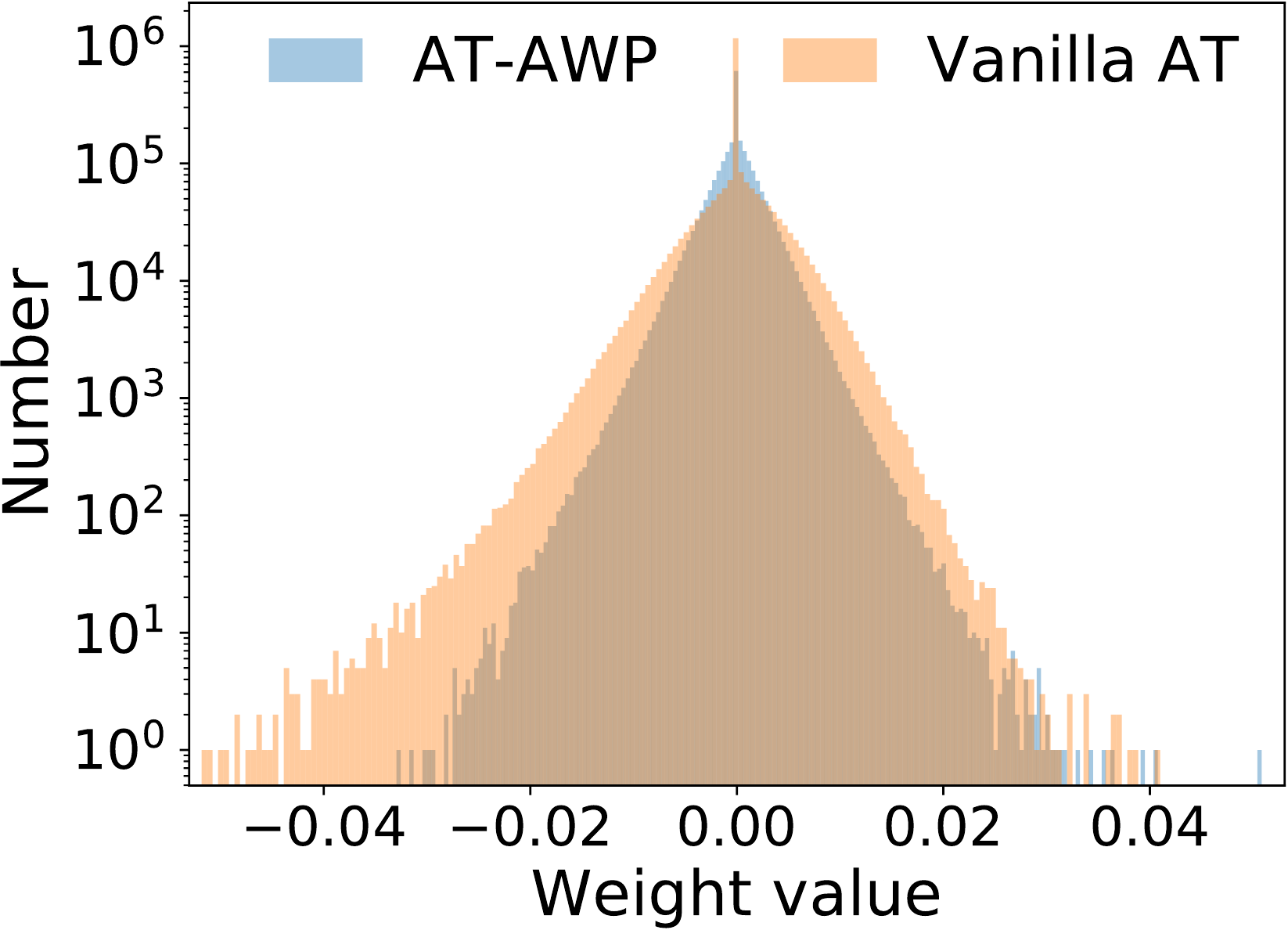}
\vspace{-0.12 in}
\caption{Weight distribution}
\label{weight_distribution}
\end{wrapfigure}
In this part, we explore how the distribution of weights changes when we apply AWP on it. 
We plot the histogram of weight values in different layers, and find that AT-AWP and vanilla AT are similar in shallower layers, while AT-AWP has smaller magnitudes and a more symmetric distribution in deeper layers. Figure \ref{weight_distribution} demonstrates the distribution of weight values in the last convolutional layer of PreAct ResNet-18 on CIFAR-10 dataset.

\section{Conclusion}
In this paper, we characterized the weight loss landscape using the on-the-fly generated adversarial examples, and identified that the weight loss landscape is closely related to the robust generalization gap. Several well-recognized adversarial training variants all introduce a flatter weight loss landscape though they use different techniques to improve adversarial robustness. Based on these findings, we proposed \textit{Adversarial Weight Perturbation (AWP)} to directly make the weight loss landscape flat, and developed a double-perturbation (adversarially perturbing both inputs and weights) mechanism in the adversarial training framework. Comprehensive experiments show that AWP is generic and can improve the state-of-the-art adversarial robustness across different adversarial training approaches, network architectures, threat models, and benchmark datasets.

\newpage
\section*{Broader Impact}
Adversarial training is the currently most effective and promising defense against adversarial examples. In this work, we propose AWP to improve the robustness of adversarial training, which may help to build a more secure and robust deep learning system in real world. At the same time, AWP introduces extra computation, which probably has negative impacts on the environmental protection (\textit{e.g.}, low-carbon).
Further, the authors do not want this paper to bring overoptimism about AI safety to the society. The majority of adversarial examples are based on known threat models (\textit{e.g.} $L_p$ in this paper), and the robustness is also achieved on them. Meanwhile, the deployed machine learning system faces attacks from all sides, and we are still far from complete model robustness.

\begin{ack}
Yisen Wang is partially supported by the National Natural Science Foundation of China under Grant 62006153, and CCF-Baidu Open Fund. Shu-Tao Xia is partially supported by the National Key Research and Development Program of China under Grant 2018YFB1800204, the National Natural Science Foundation of China under Grant 61771273, the R\&D Program of Shenzhen under Grant JCYJ20180508152204044, and the project PCL Future Greater-Bay Area Network Facilities for Large-scale Experiments and Applications (LZC0019).
\end{ack}

\small
\bibliography{ref}
\bibliographystyle{plainnat}

\clearpage
\normalsize
\appendix

\section{Adversarial Attack}\label{sec:attack}
Given a natural example $\xb_i$ with class label $y_i$ and a DNN model $\fb_\wb$, the goal of an adversary is to find an adversarial example $\xb^\prime_i$ that fools the network to make incorrect predictions while still remains in the $\epsilon$-ball centered at $\xb_i$ ($\|\xb_i^\prime - \xb_i\|_{p} \leq \epsilon$). A lot of attacking methods have been proposed for the crafting of adversarial examples. Here, we only name a few.

\vspace{-0.05 in}
\textbf{Fast Gradient Sign Method (FGSM) \citep{goodfellow2015explaining}.} FGSM perturbs the natural example $\xb_i$ for one step by the amount of $\epsilon$ along the gradient direction:
\begin{equation}
    \xb_i' = \xb_i + \epsilon \cdot \text{sign}(\nabla_{\xb_i} \ell(\fb_\wb(\xb_i), y_i)). 
\end{equation}

\vspace{-0.05 in}
\textbf{Projected Gradient Descent (PGD)  \citep{madry2018towards}.} PGD perturbs the natural example $\xb_i$ for $K_1$ steps with small step size $\eta_1$. After each step of perturbation, PGD projects the adversarial example back onto the $\epsilon$-ball of $\xb_i$, if it goes beyond the $\epsilon$-ball:
\begin{equation}
    \xb_i'^{(k+1)} = \Pi_{\epsilon} \big(\xb_i'^{(k)} + \eta_1 \cdot \text{sign}(\nabla_{\xb_i'} \ell(\fb_\wb(\xb_i'^{(k)}), y_i)) \big),
\end{equation}
where  $\Pi(\cdot)$ is the projection operation, and $\xb_i'^{(k)}$ is the adversarial example at the $k$-th step. There are also other types of attacks including Jacobian-based Saliency Map Attack (JSMA) \citep{papernot2016limitations}, Carlini and Wagner (CW) \citep{carlini2017towards}, and so on.

\section{Details for the Weight Loss Landscape Visualization Method}

In this section, we first provide the pseudo-code of our proposed visualization method for the weight loss landscape in the adversarial training, and then verify its reliability.

\subsection{Pseudo-code of the Visualization Method}
\label{sec:visualization_code}
\vspace{-0.05 in}

As shown in Algorithm \ref{alg:visualization} for the visualization of weight loss landscape, we firstly sample a random direction $\db$ from a Gaussian distribution. Then, we apply the ``filter normalization'' technique (Line 3-7) from \citet{li2018visualizing} to avoid the scaling effect\footnote{In DNNs with ReLU activation, the network remains unchanged if we multiply the weights in one layer by 10, and divide by 10 at the next layer.} of DNNs. 
Next, we calculate the adversarial loss for a series of perturbed weights independently, \textit{i.e.}, $\rho(\wb+\alpha \db), \alpha \in \{\alpha_{min}, \cdots,\alpha_{max}\}$. For a given perturbed weights $\wb + \alpha \db$, we generate its own adversarial examples using PGD following \citet{madry2018towards} (Line 9-14). Then, we approximate the adversarial loss of the current perturbed model $f_{\wb + \alpha \db}$ by the cross-entropy loss on these on-the-fly generated adversarial examples (Line 15). Finally, we plot the weight loss landscape (Line 17). 

\begin{algorithm}[!htbp]
   \caption{Visualization of Weight Loss Landscape}
   \label{alg:visualization}
\begin{algorithmic}[1]
   \STATE {\bfseries Input:} Network $\fb_\wb$ with $L$-layer ($F_l$ filters in the $l$-th layer), training data $ \{(\xb_i, y_i)\}_{i=1}^n$, PGD step size $\eta_1$, PGD step number $K_1$, the scalar parameter $\alpha \in [\alpha_{min}, \alpha_{max}]$.
   \STATE Sample a random direction $\db \sim \mathcal{N}(0, 1)$
   \FOR{$l = 1, \dots, L$}
      \FOR{$j = 1, \dots, F_l$}
   	 \STATE $\db_{l, j} \leftarrow \frac{\db_{l,j}}{\Vert \db_{l,j} \Vert_F} \Vert \wb_{l,j} \Vert_F$
      \ENDFOR
   \ENDFOR
   \FOR{$\alpha = \alpha_{min}, \cdots, \alpha_{max}$}
      \FOR{$i = 1, \dots, n$ (in parallel) }
      \STATE $\xb_i^\prime \leftarrow \xb_i + \epsilon \delta$, where $\delta \sim \text{Uniform}(-1, 1)$
      \FOR{$k = 1, \cdots, K_1$}
      \STATE $\xb_i^\prime \leftarrow \Pi_{\epsilon} \big(\xb_i^\prime + \eta_1  \text{sign}( \nabla_{\xb_i^\prime} \ell(\fb_{\wb + \alpha \db}(\xb_i^\prime), y_i)) \big)$
      \ENDFOR
      \ENDFOR
   \STATE $\rho(\wb + \alpha \db) \leftarrow \frac{1}{n} \sum_{i=1}^{n} \ell(f_{\wb + \alpha \db}(\xb_i^\prime), y_i)$
   \ENDFOR
   \STATE Plot $(\alpha, \rho(\wb + \alpha \db)), \forall \alpha \in [\alpha_{min}, \alpha_{max}]$
\end{algorithmic}
\end{algorithm}

\subsection{Reliability of the 1-D Visualization}
\label{sec:repeatability}

To improve the trustworthiness of our results, we check the reliability of our visualization method from two perspectives: 1) repeatability; 2) comparisons to 2-D visualization. 

\begin{figure*}[t]
\vspace{-0.1 in}
\centering
    \subfigure[Epoch 20]{
        \includegraphics[width=0.18\columnwidth]{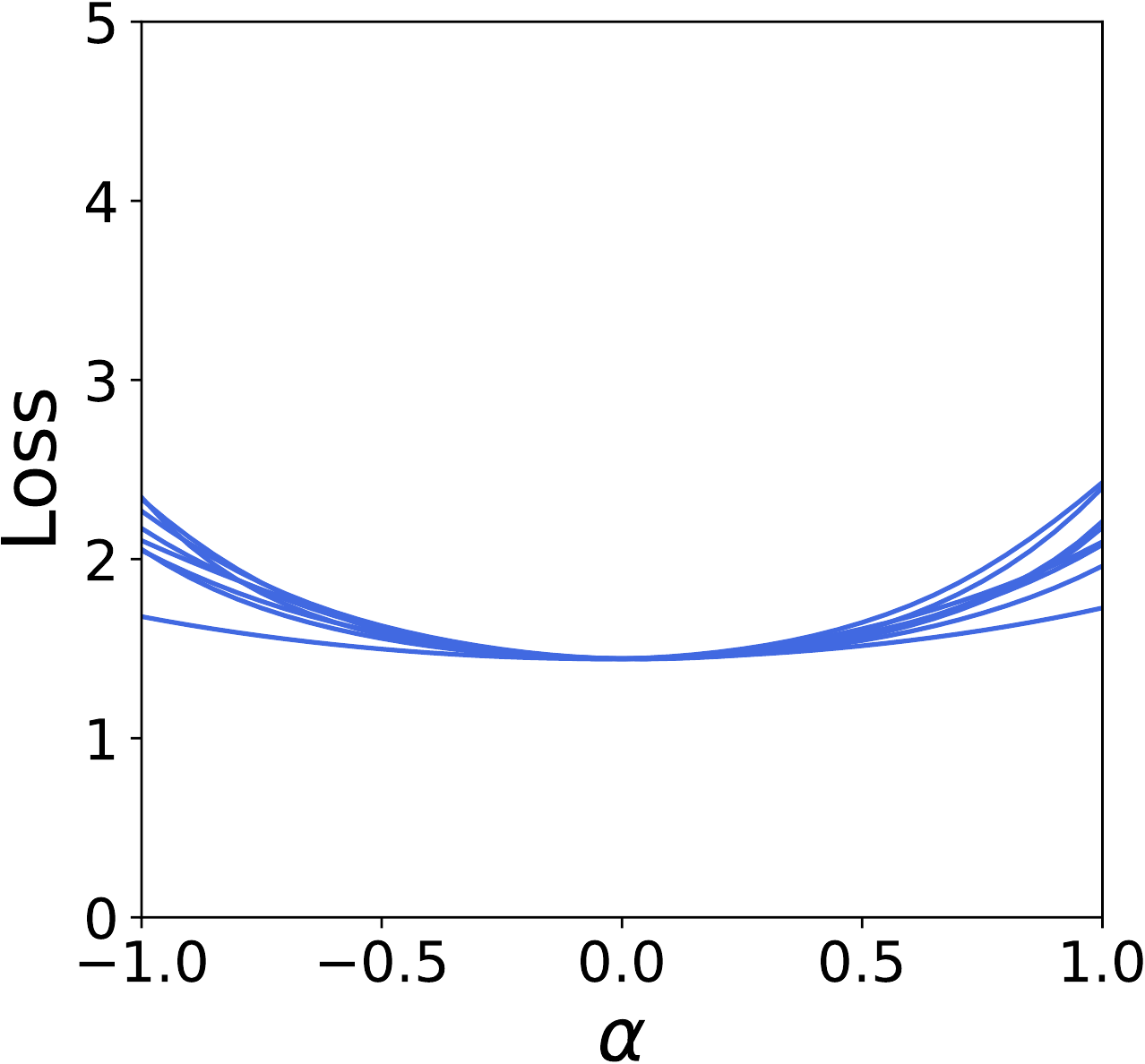}
    }
    \subfigure[Epoch 40]{
        \includegraphics[width=0.18\columnwidth]{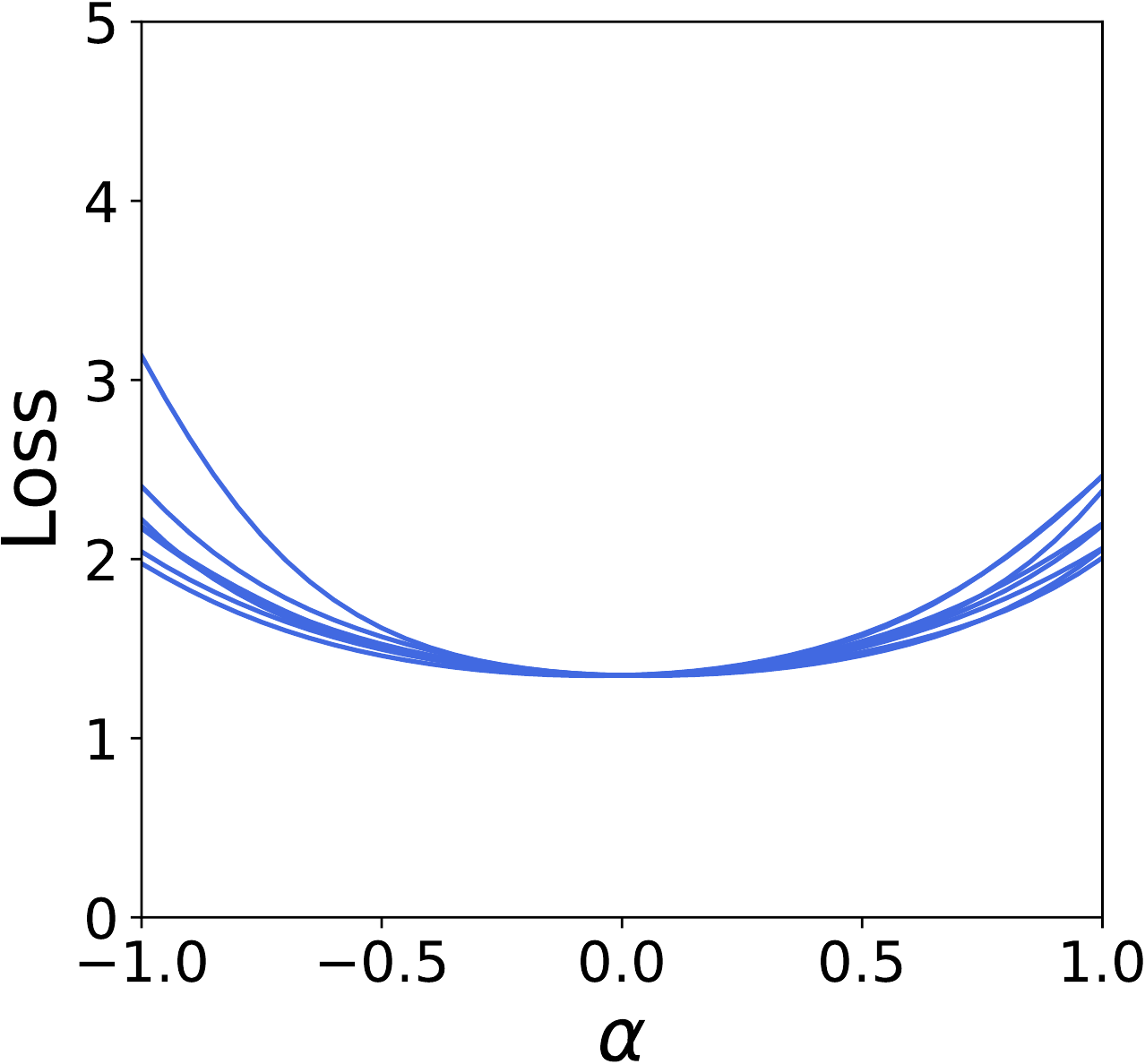}
    }
    \subfigure[Epoch 60]{
        \includegraphics[width=0.18\columnwidth]{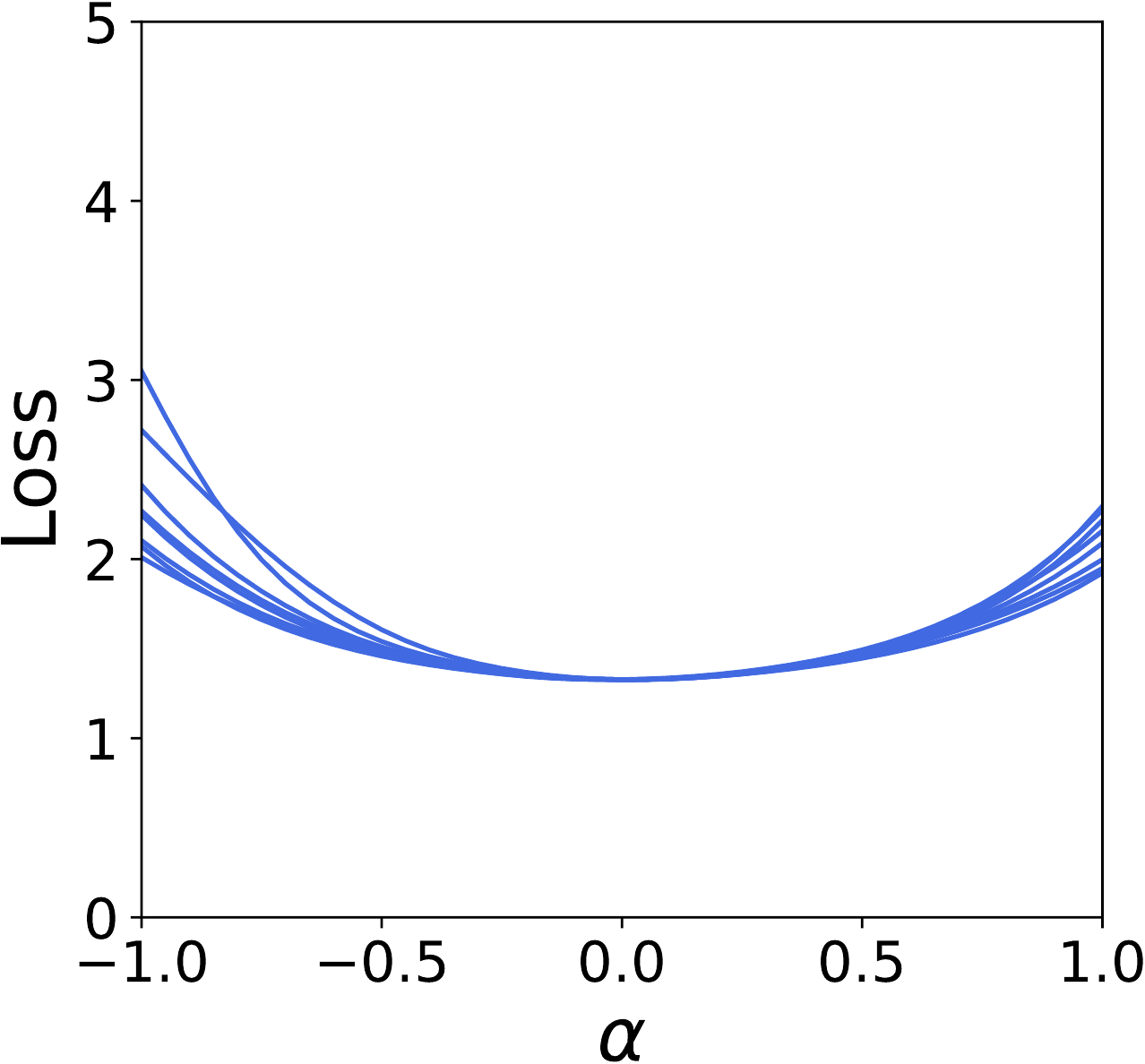}
    }
    \subfigure[Epoch 80]{
	    \includegraphics[width=0.18\columnwidth]{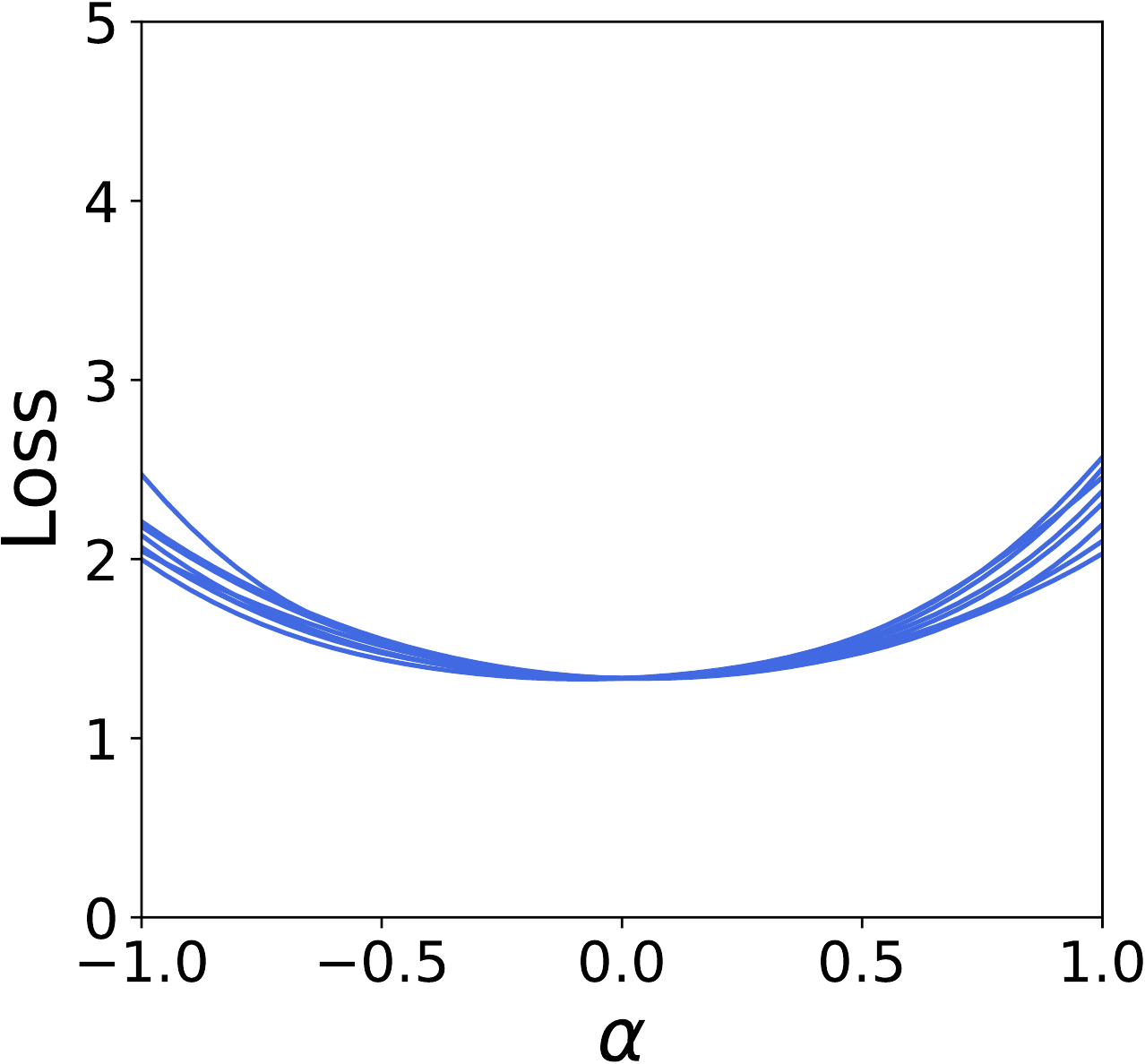}
    }
    \subfigure[Epoch 100]{
	    \includegraphics[width=0.18\columnwidth]{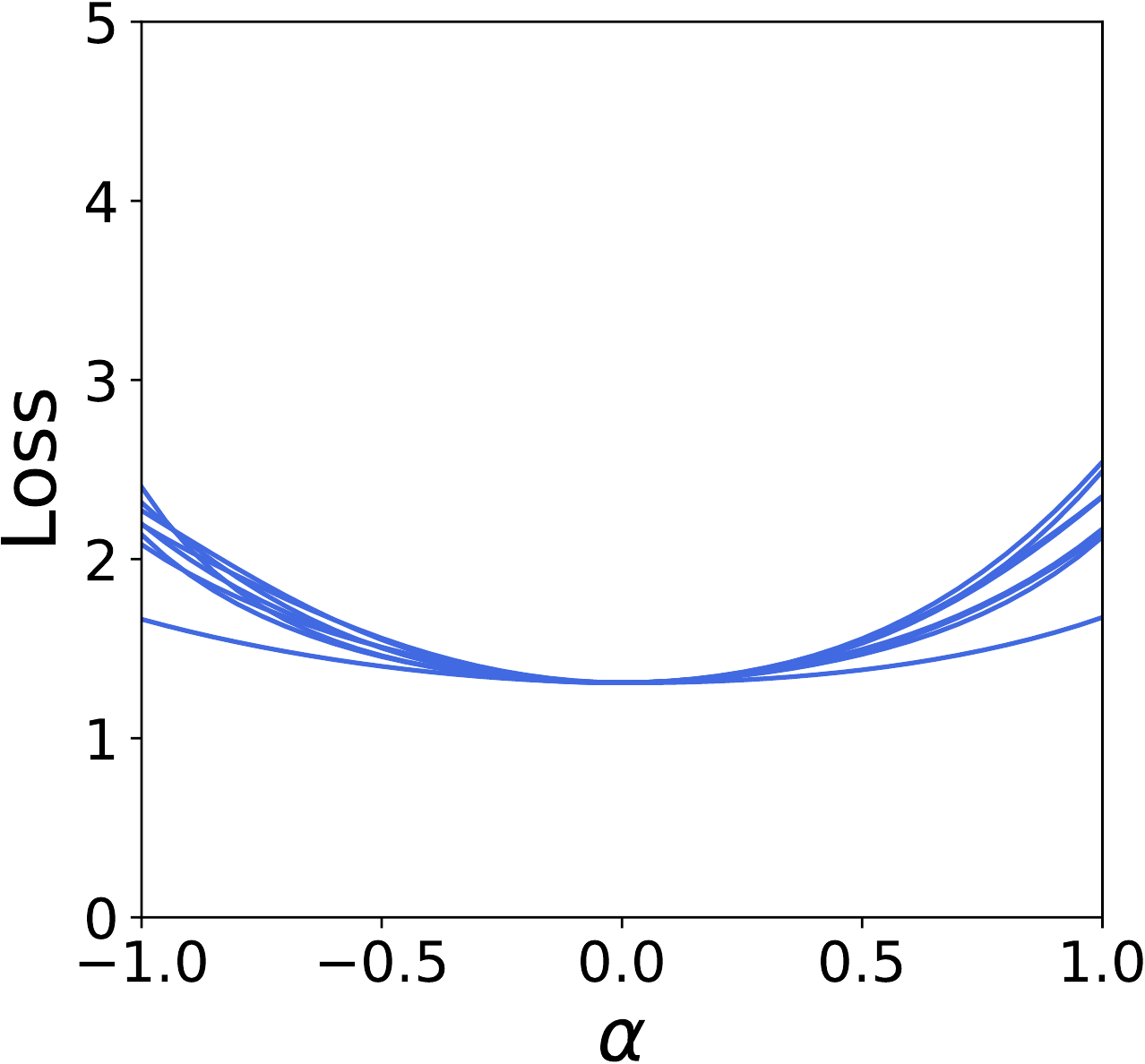}
    }\\
    \subfigure[Epoch 120]{
        \includegraphics[width=0.18\columnwidth]{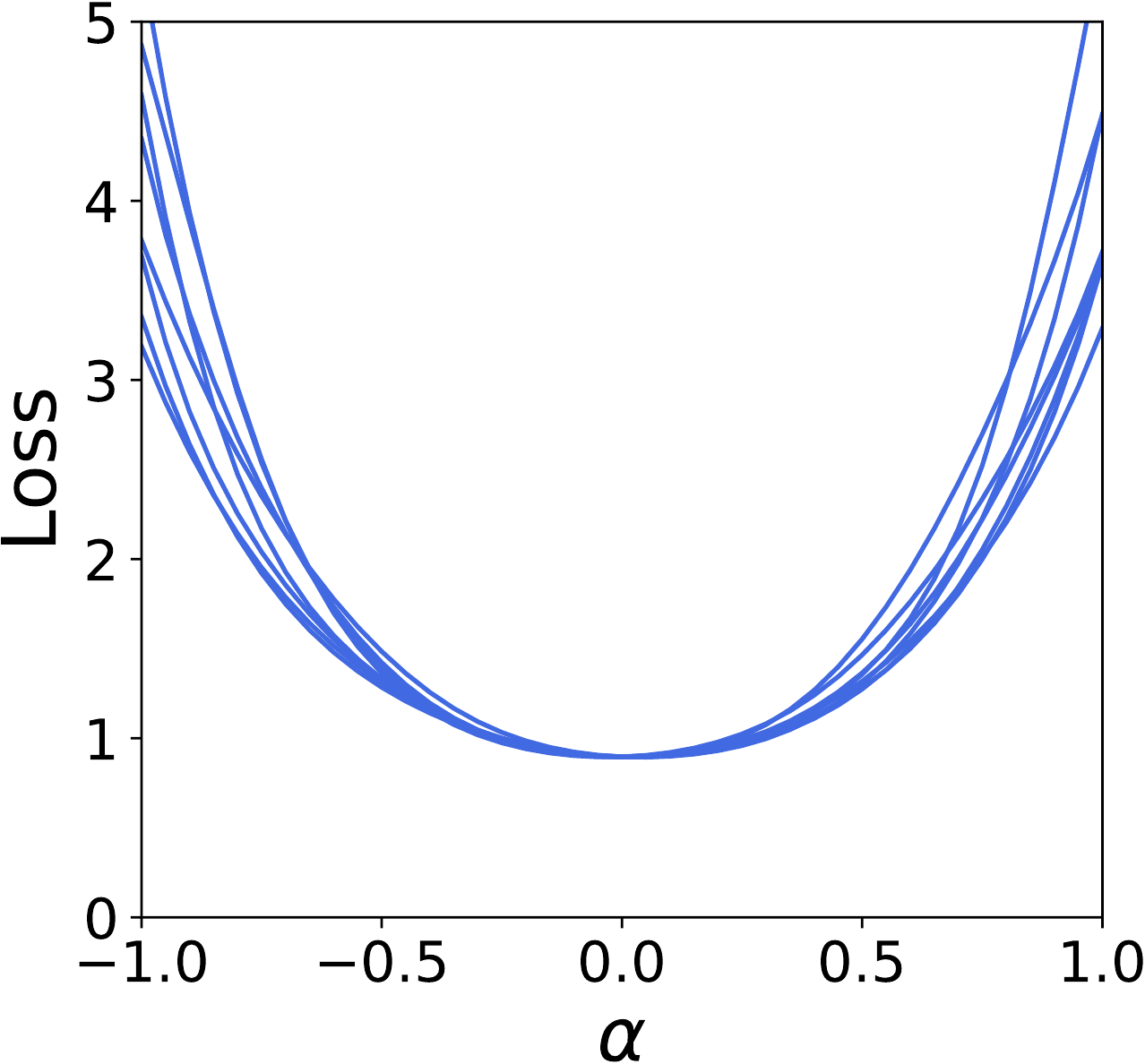}
    }
    \subfigure[Epoch 140]{
        \includegraphics[width=0.18\columnwidth]{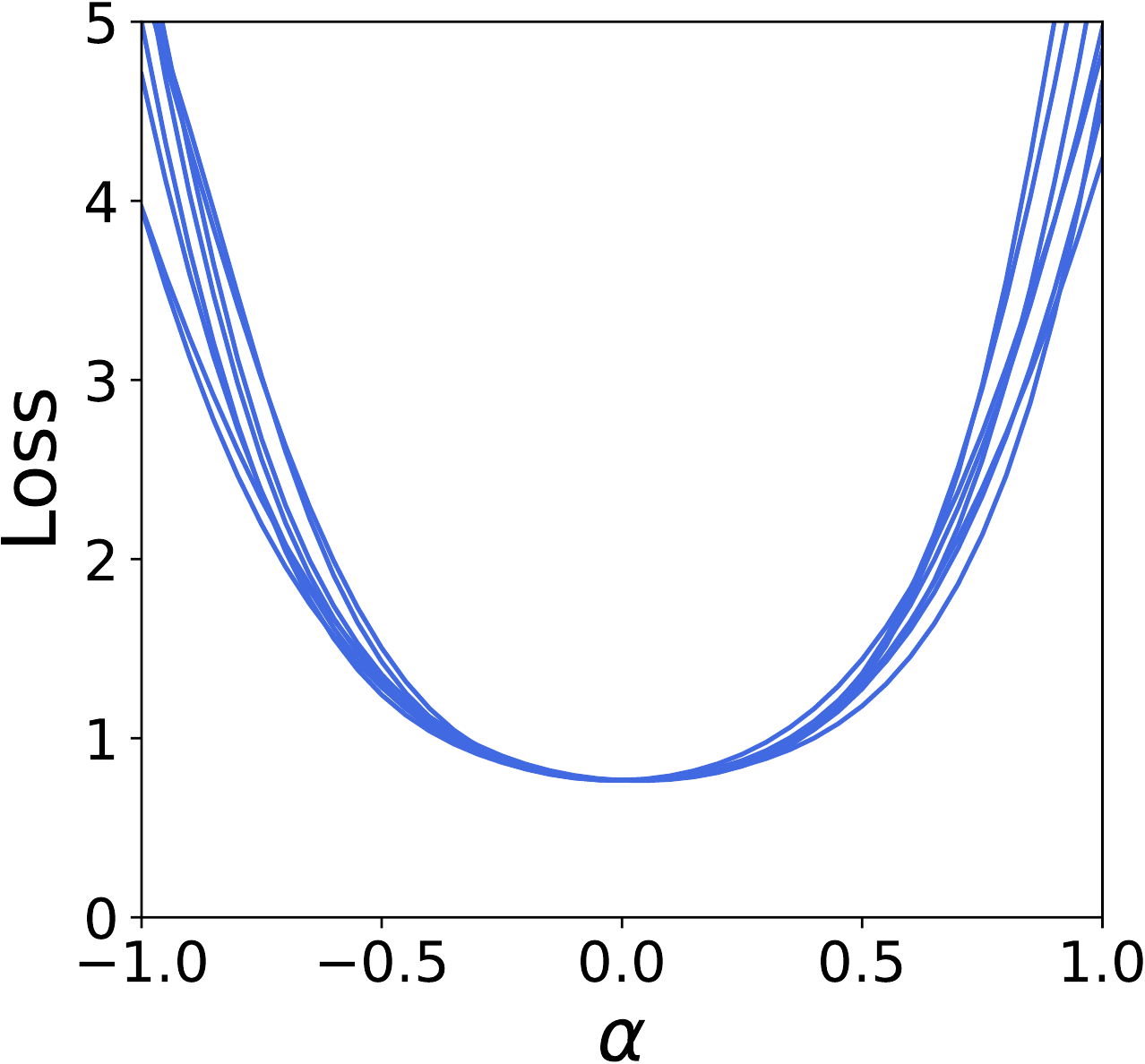}
    }
    \subfigure[Epoch 160]{
        \includegraphics[width=0.18\columnwidth]{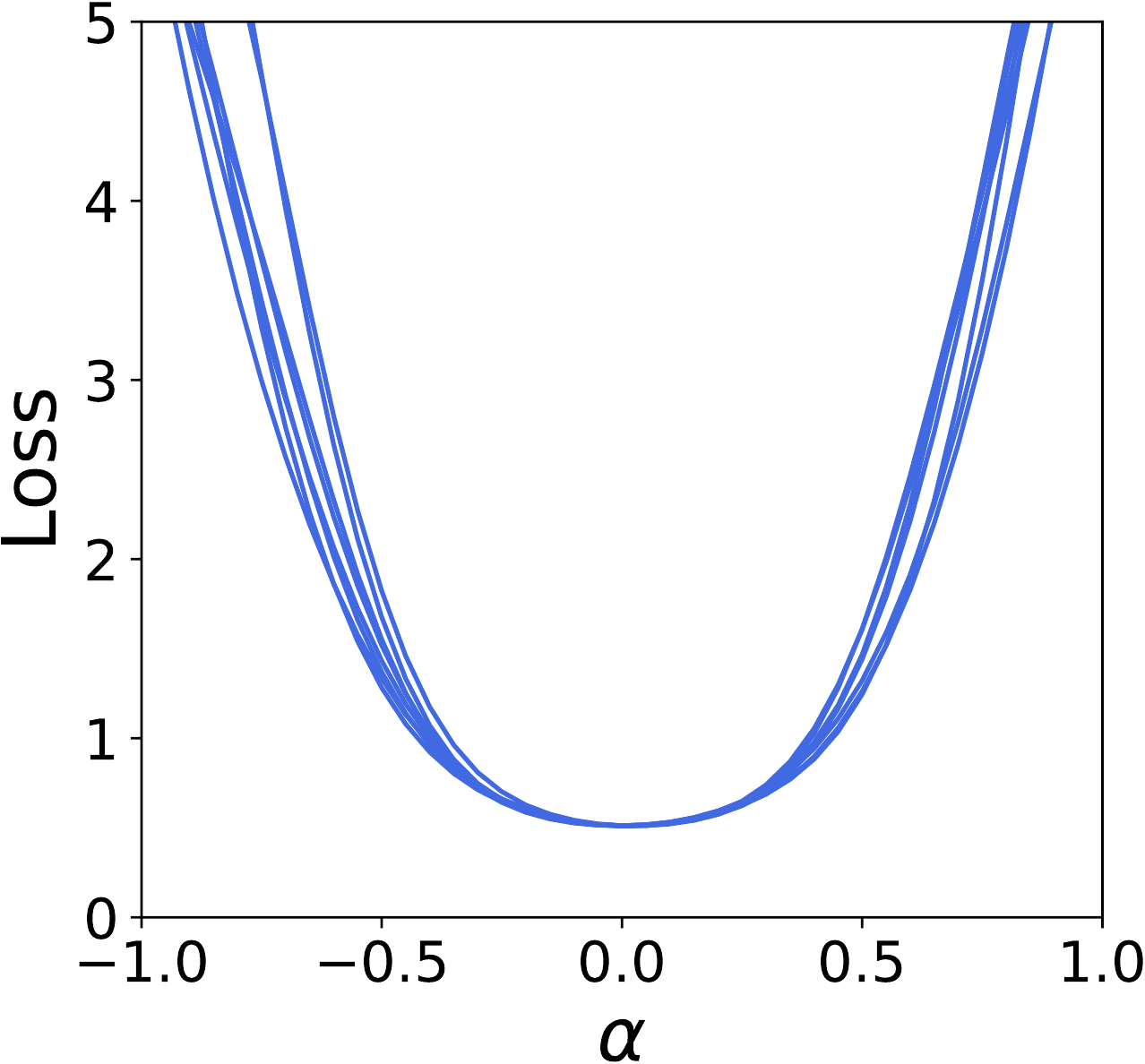}
    }
    \subfigure[Epoch 180]{
	    \includegraphics[width=0.18\columnwidth]{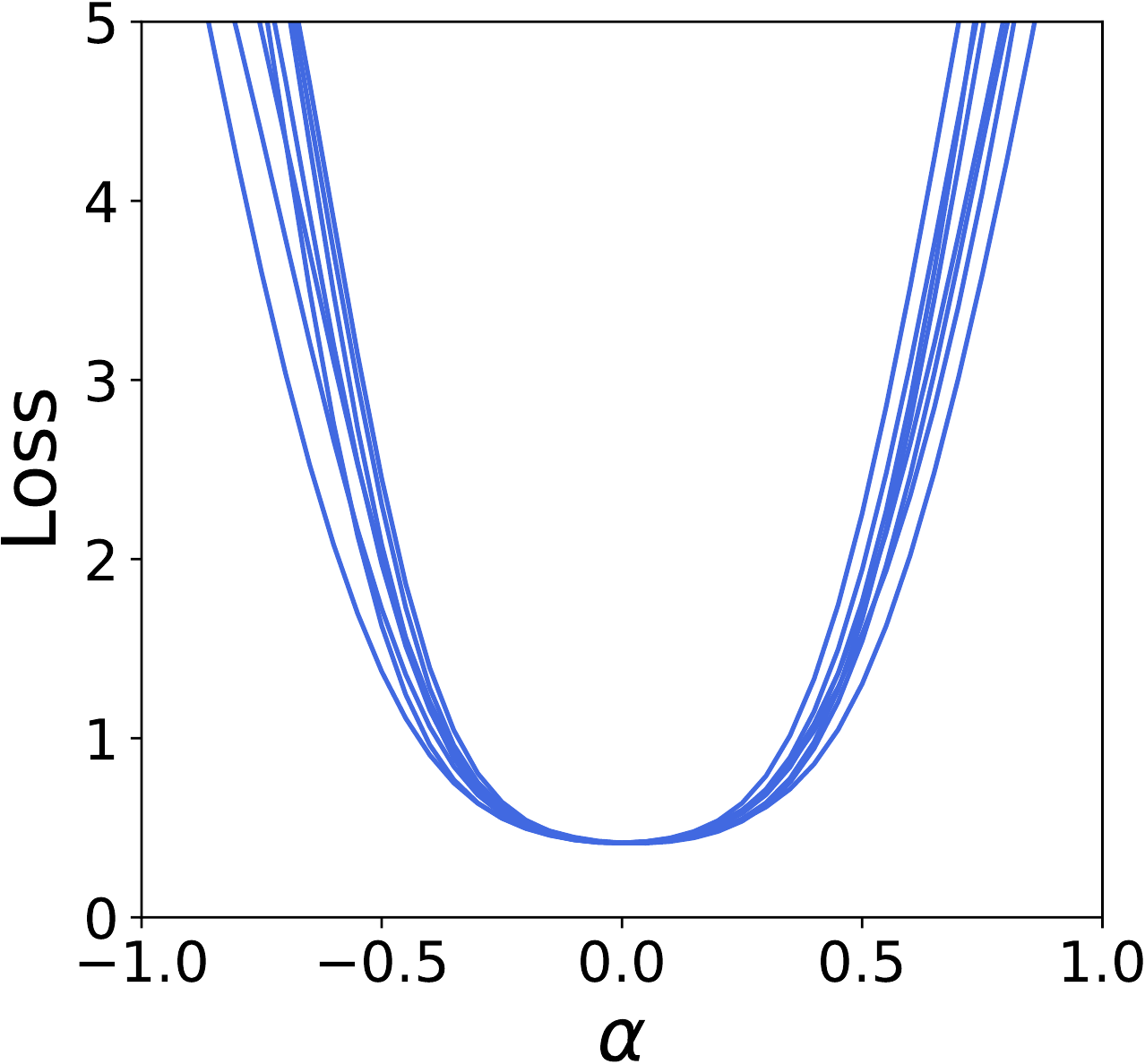}
    }
    \subfigure[Epoch 200]{
	    \includegraphics[width=0.18\columnwidth]{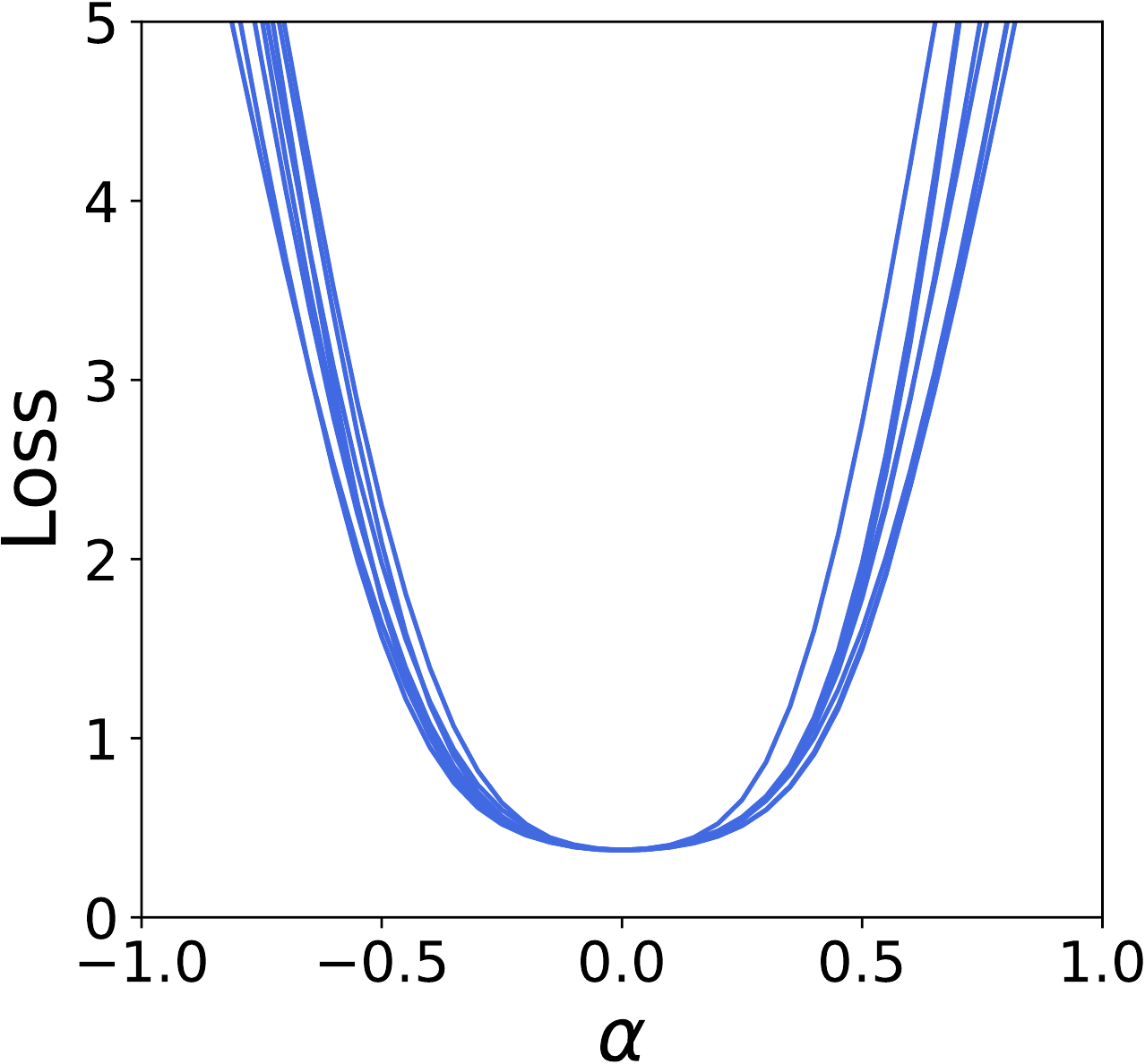}
    }\\
\vspace{-0.1 in}
\caption{Repeatability of the 1-D visualization along 10 different random directions.}
\label{fig:repeat}
\vspace{-0.1 in}
\end{figure*}

\textbf{Repeatability.} We first investigate whether different random directions produce dramatically different plots. We show the weight loss landscape of PreAct ResNet-18 during vanilla adversarial training along 10 randomly selected directions in Figure \ref{fig:repeat}. We find the plots of the same checkpoint are very close in shape, 
which indicates the stability of our visualization method.

\begin{figure}[t]
\centering
    \subfigure[Epoch 100]{
        \includegraphics[width=0.30\columnwidth]{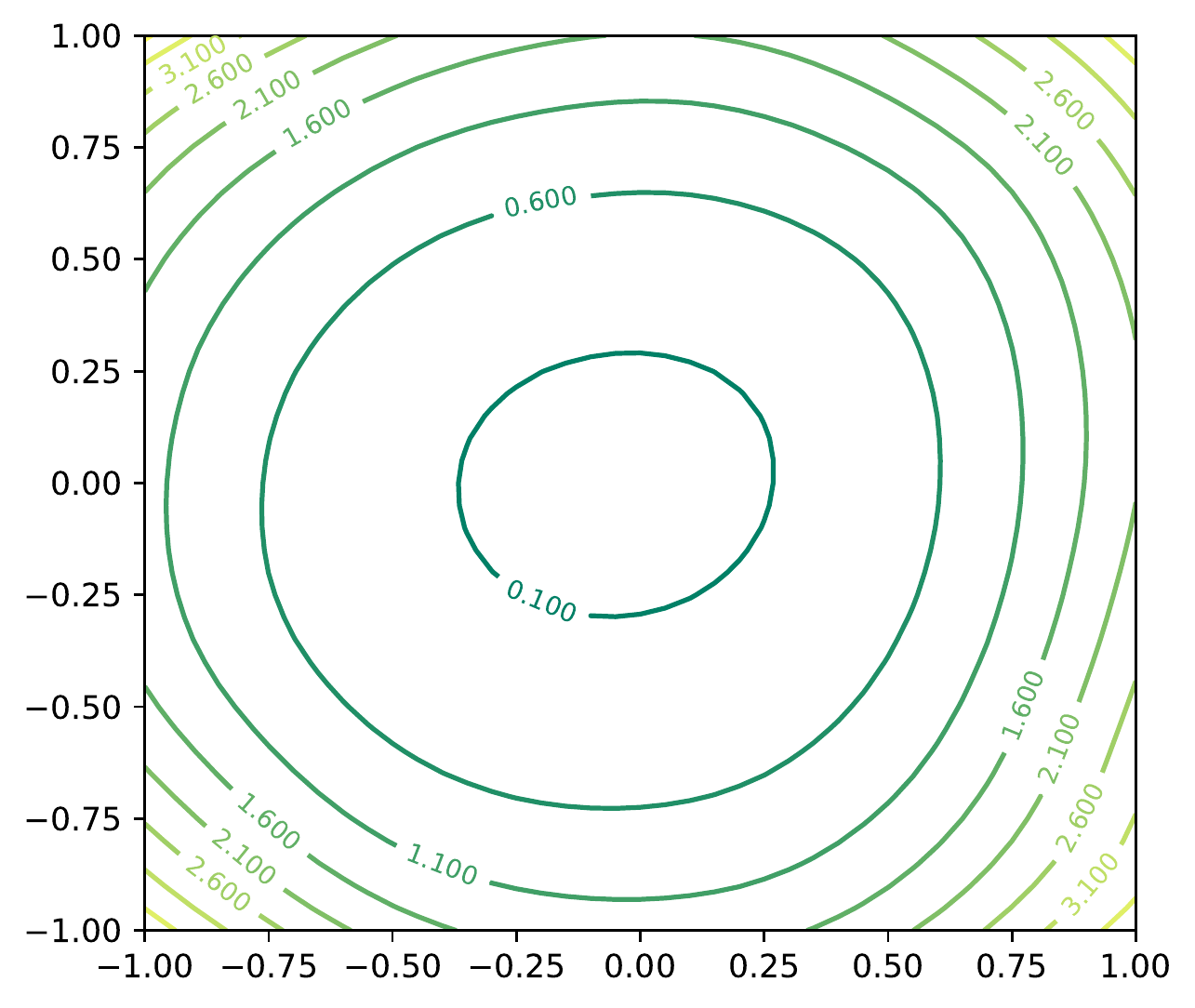}
    }
    \subfigure[Epoch 140]{
        \includegraphics[width=0.30\columnwidth]{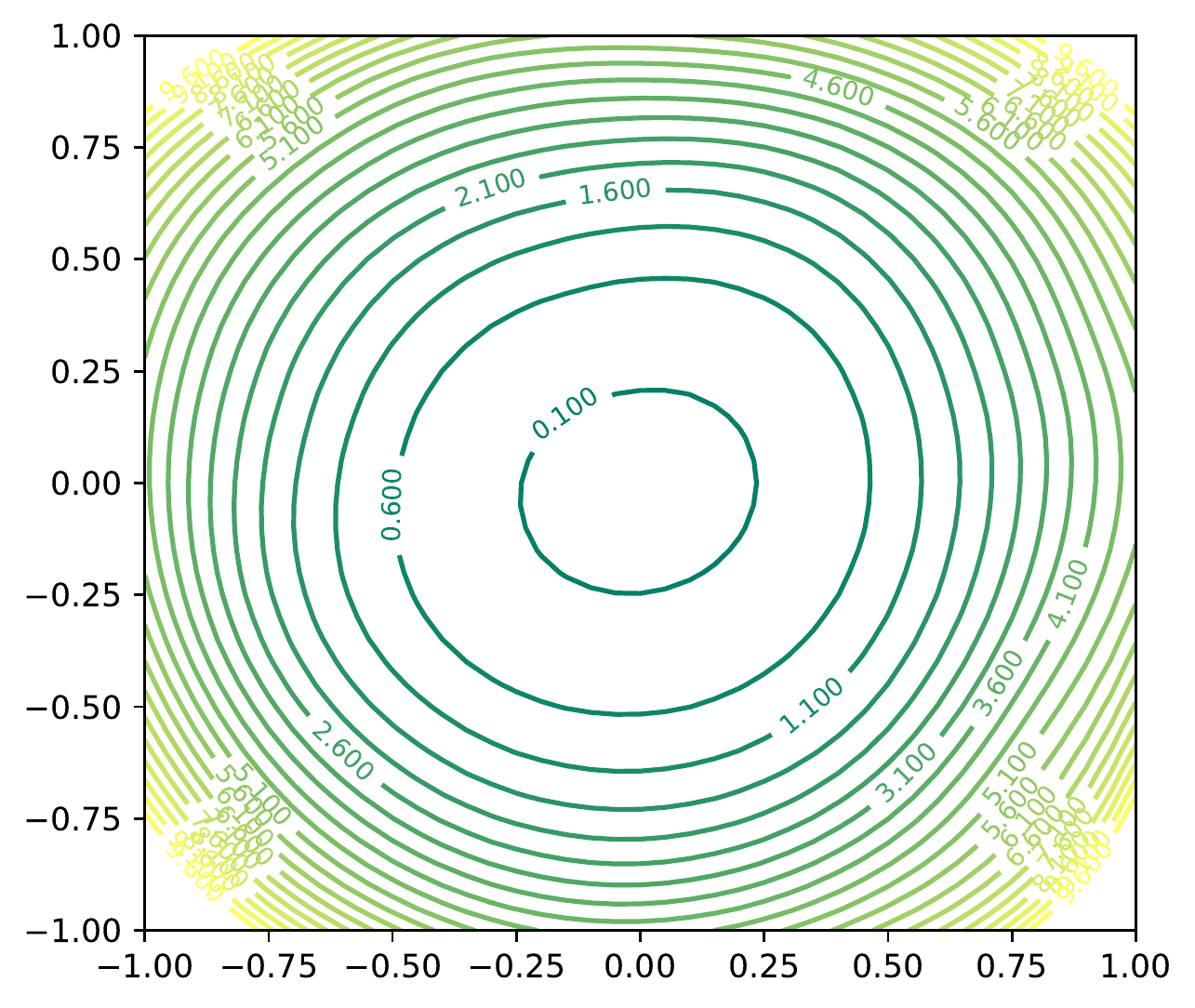}
    }
    \subfigure[Epoch 200]{
	\includegraphics[width=0.30\columnwidth]{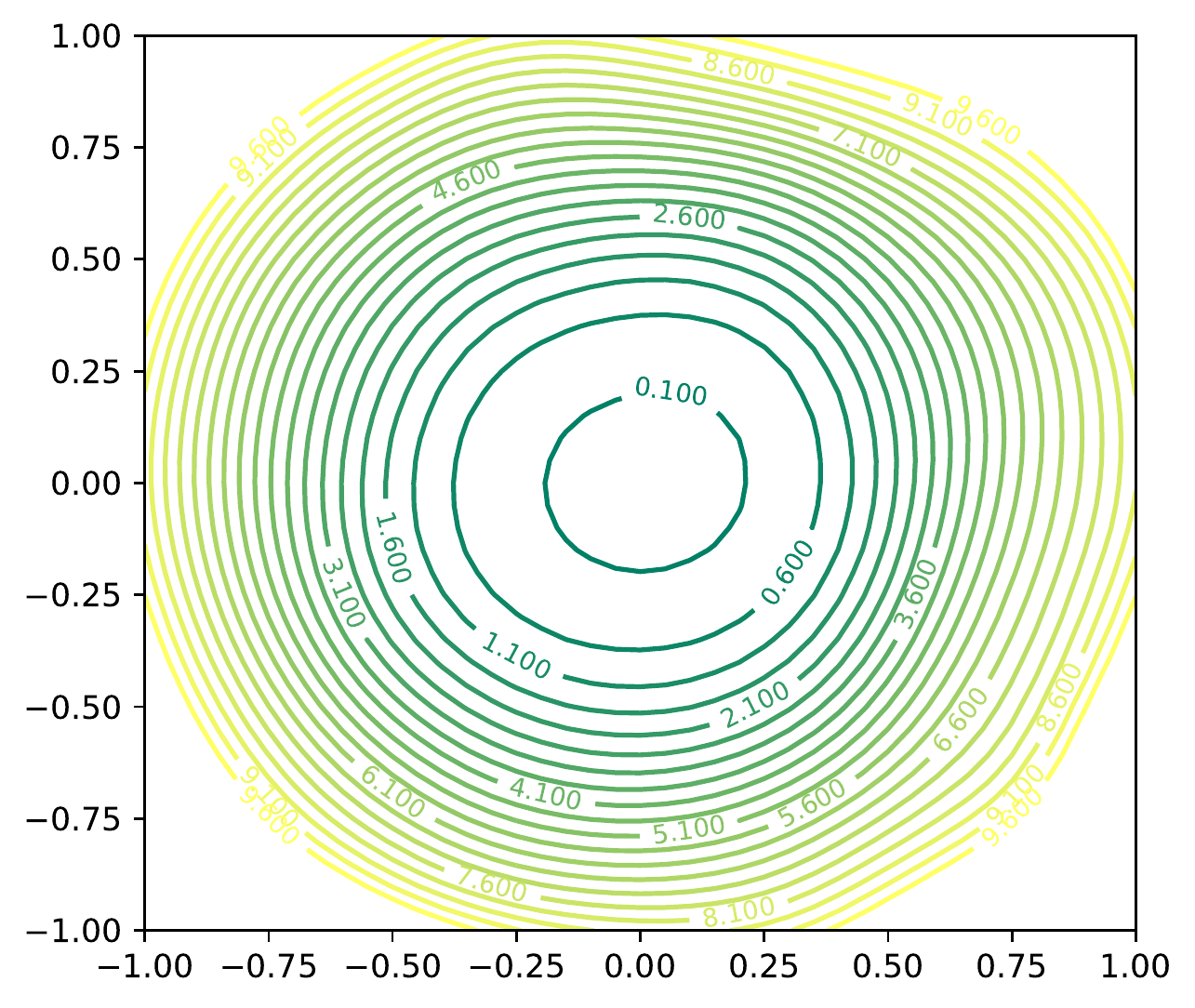}
    }
\vskip -0.1in
\caption{2-D visualization of weight loss landscape at the  different epoch checkpoints.}
\label{fig:repeat_surface}
\end{figure}

\textbf{Comparisons to 2-D Visualization.} Next, we explore whether 1-D visualization obtains similar results to 2-D visualization (a much time-consuming method). The 2-D visualization introduces an extra random filter-normalized direction $\mathbf{e}$, and plots $g(\alpha, \beta) = \rho(\wb + \alpha \db + \beta \mathbf{e})$. Different from the standard training scenario where near-zero loss on the training set can be always achieved, the adversarial loss on the training set is usually larger than zero, \textit{i.e.}, the center point $g(0, 0)$ in the weight loss landscape is often larger than zero, which hampers the comparison on the flatness of weight loss landscape. Therefore, we visualize the relative weight loss landscape $\vert g(\alpha, \beta) - g(0, 0) \vert$ instead. Figure \ref{fig:repeat_surface} presents the 2-D visualization of the same model as Figure \ref{fig:repeat} at the 100-th, 140-th and 200-th epoch checkpoint. We can see that the weight loss landscape is flatter at the 100-th epoch and sharper at the 200-th epoch, which is compatible to the findings from the 1-D visualization. For time costs, it consumes $\sim 4$ days for the 2-D visualization of a single adversarially trained PreAct ResNet-18 using one GeForce RTX 2080Ti, while it is $\sim$ 2 hours for the 1-D visualization of the same model. Thus, we adopt the 1-D visualization in most cases. 

From the above experiments, we can conclude that our 1-D visualization method can characterize the property of the high-dimensional weight loss landscape reliably and efficiently.

\section{More Evidence for the Connection of Weight Loss Landscape and Robust Generalization Gap}
\label{sec:more_visualization}
\vspace{-0.05 in}
In this section, we provide more empricial evidence to identify the connection of  the weight loss landscape and the robust generalization gap across learning rate schedules, model architectures, datasets, and threat models.
\vspace{-0.05 in}

\subsection{The Connection across Learning Rate Schedules}

To investigate whether the learning rate schedule affects the connection of weight loss landscape and robust generalization gap, we test another two commonly used learning rate schedules:
\begin{itemize}
    \item Cosine schedule \cite{loshchilov2016sgdr}: We decrease the learning rate $lr$ using the cosine function from 0.1 to 0 over 200 epochs, \textit{i.e.}, $lr= 0.05 (\cos(\pi t/200) + 1))$ at the $t$-th epoch \citep{carmon2019unlabeled};
    \item Cyclic schedule \cite{smith2017cyclical}: We increase $lr$ linearly from 0 to some maximum (0.2 at the 80-th epoch), and then decrease it linearly to 0 until 200-th epoch \citep{wong2020fast}.
\end{itemize}

\begin{figure}[!htbp]
\vspace{-0.05 in}
\centering
    \subfigure[Cosine schedule]{
        \includegraphics[width=0.27\columnwidth]{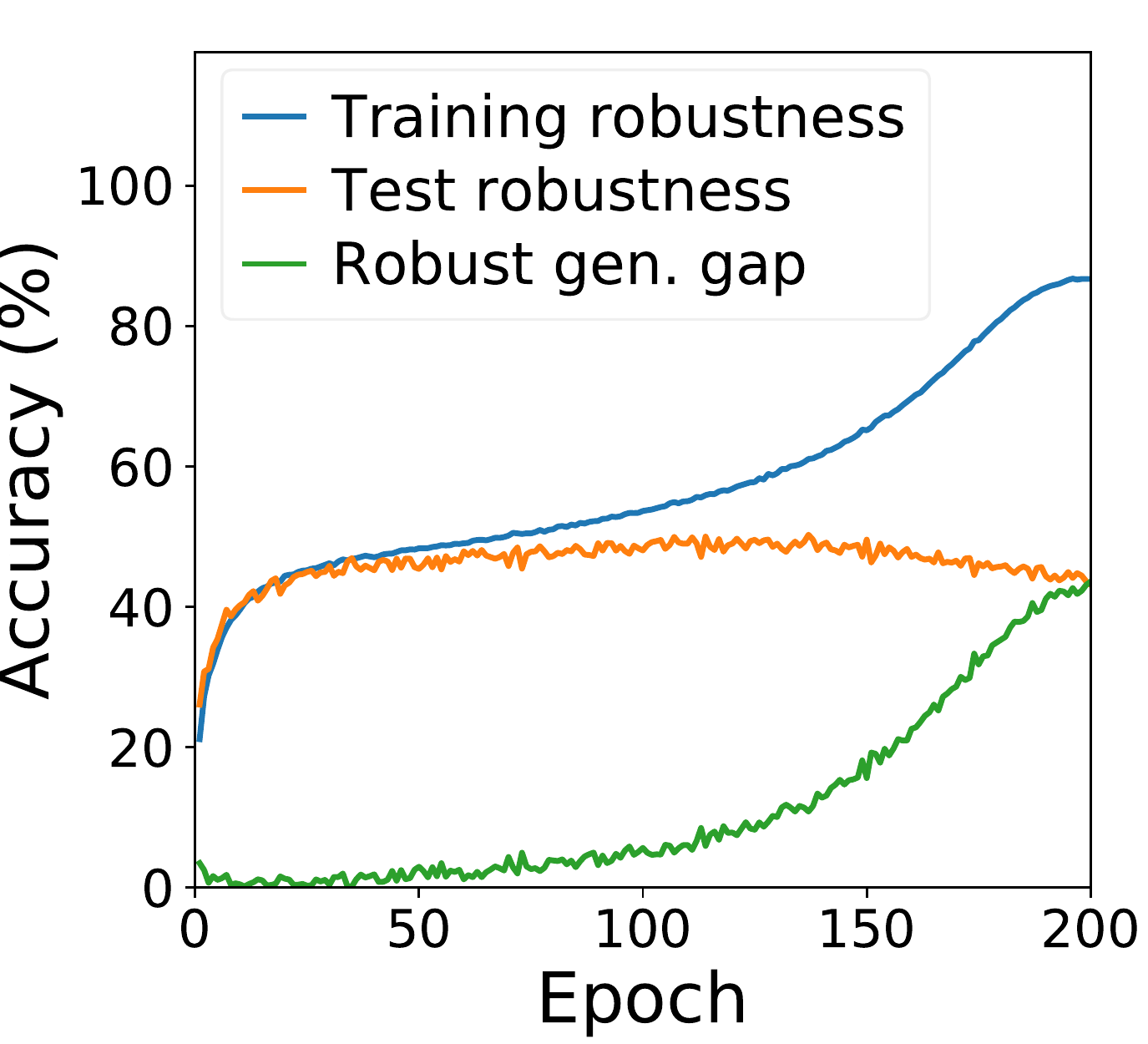}
        \includegraphics[width=0.27\columnwidth]{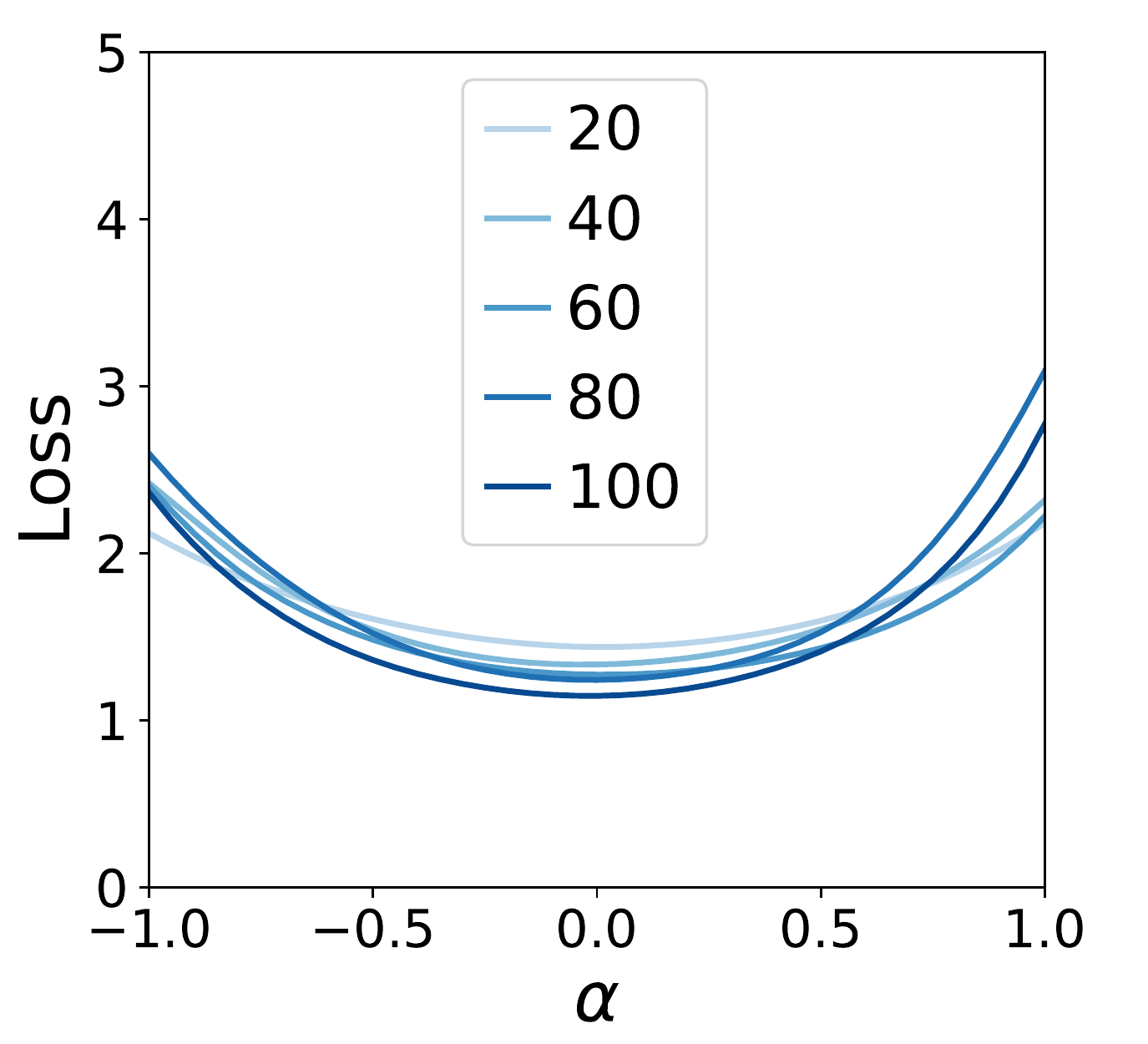}
        \includegraphics[width=0.27\columnwidth]{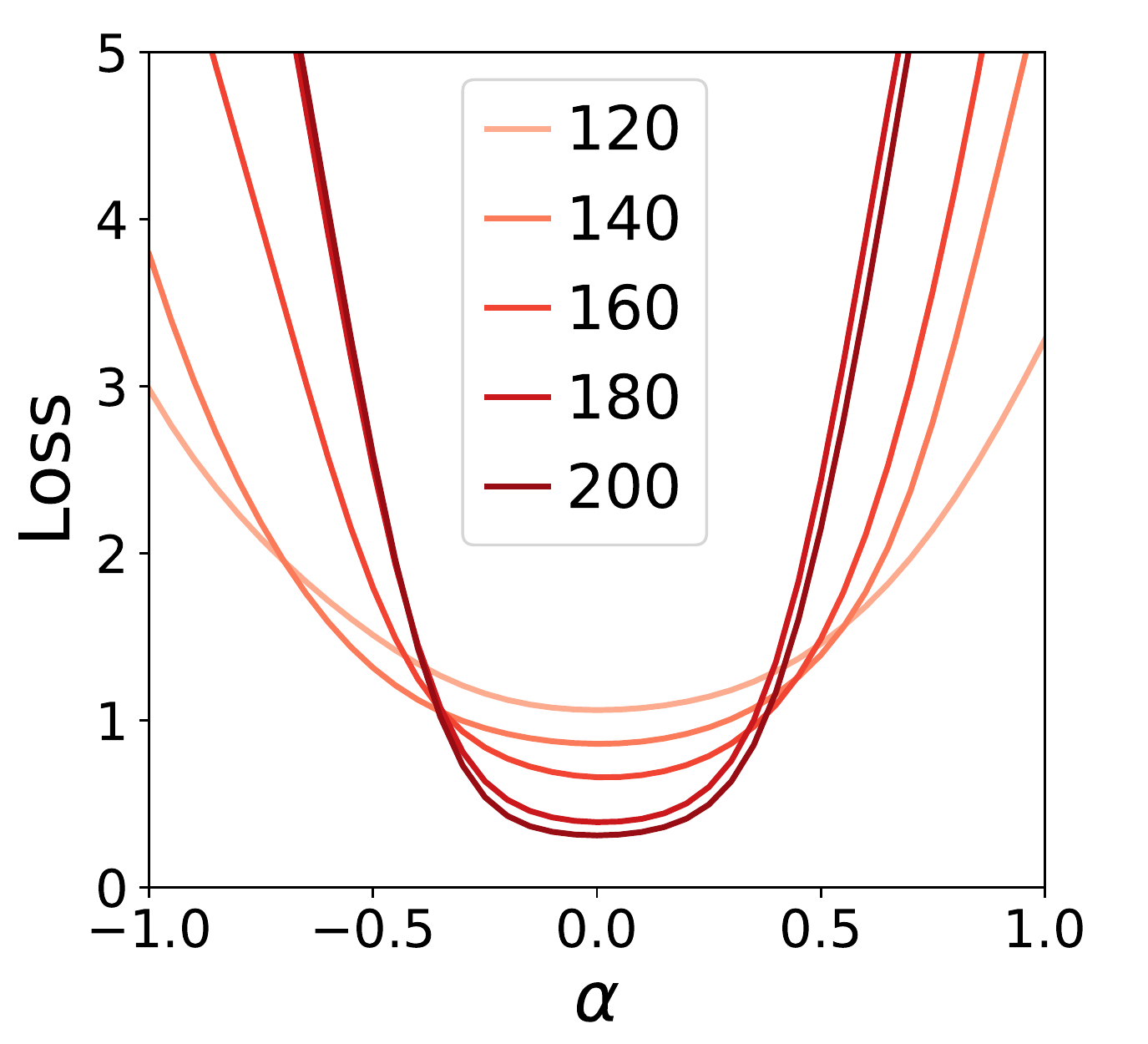}
    }\\
    \subfigure[Cyclic schedule]{
        \includegraphics[width=0.27\columnwidth]{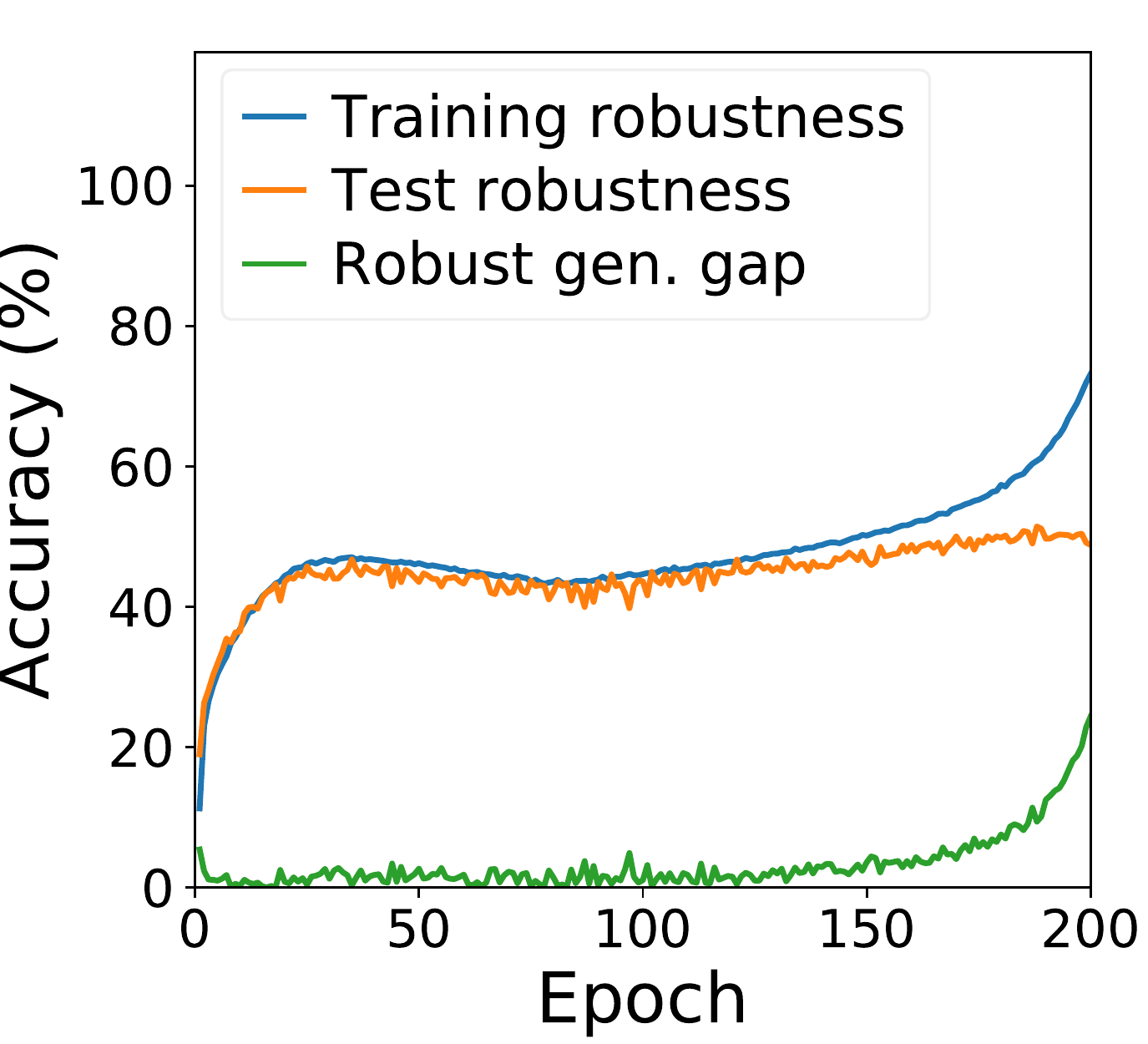}
        \includegraphics[width=0.27\columnwidth]{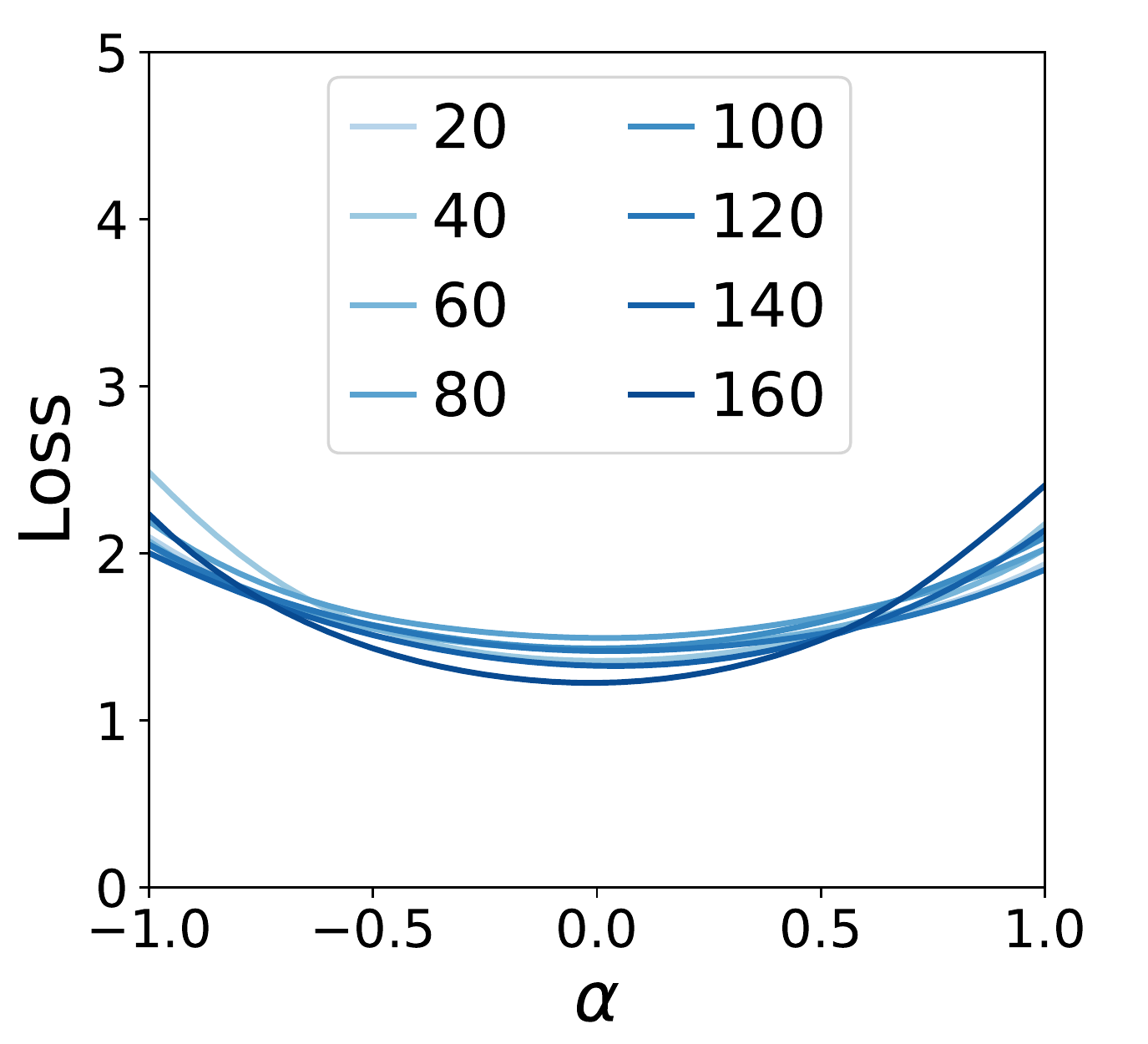}
        \includegraphics[width=0.27\columnwidth]{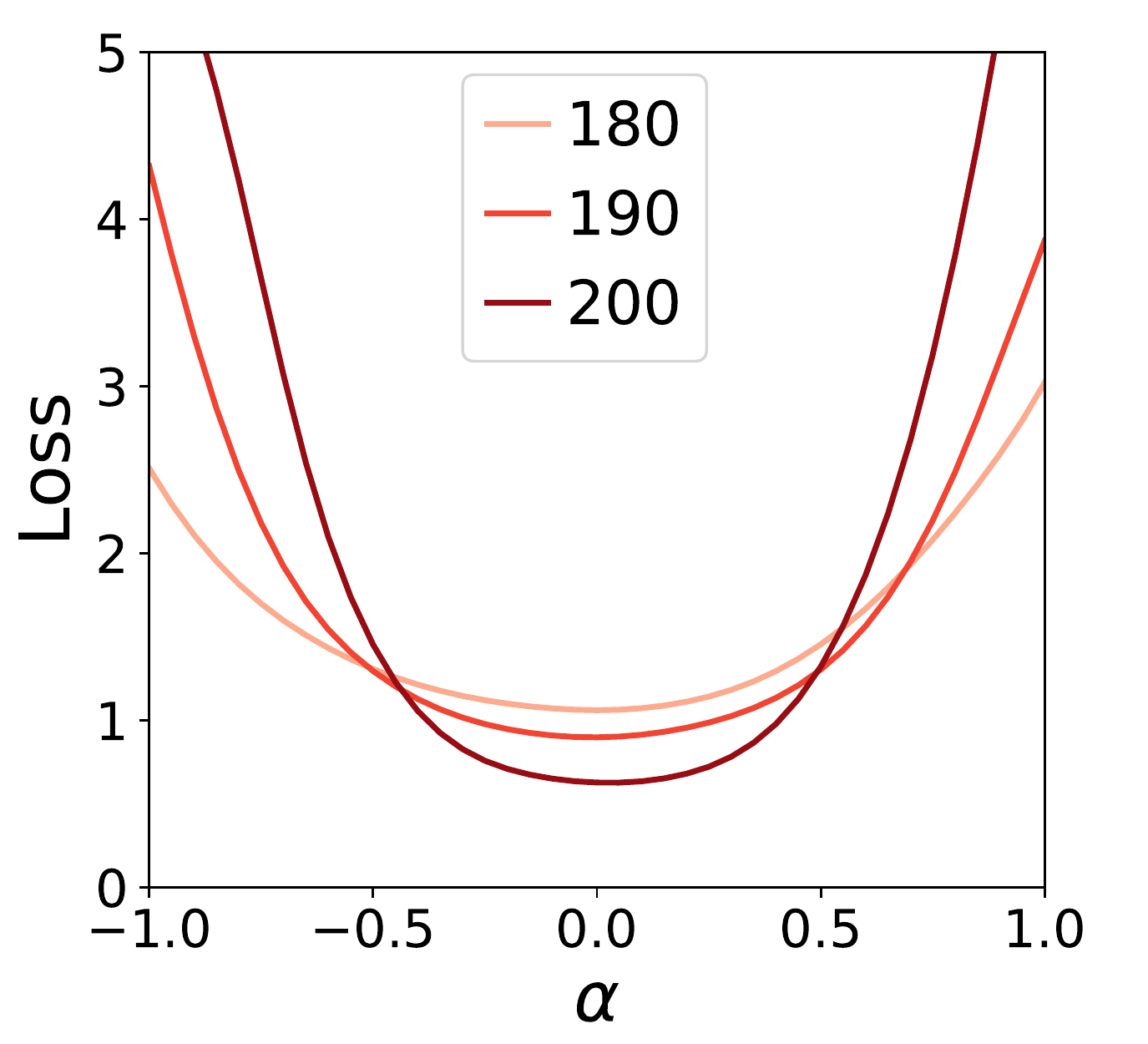}
    }
\vskip -0.1in
\caption{The relationship between weight loss landscape and robust generalization gap across learning rate schedules (cosine and cyclic) with PreAct ResNet-18 on CIFAR-10 under $L_\infty$ attack.}
\label{fig:landscape_schedule}
\end{figure}

We adversarially train PreAct ResNet-18 with different learning rate schedules using the same experimental settings in Section \ref{sec:geometry}. The learning curves are shown on the left column in Figure \ref{fig:landscape_schedule}, where the whole training process can be split into two stages: the early stage with small robust generalization gap ($\le 10\%$) and the late stage with large robust generalization gap ($> 10 \%$). In Figure \ref{fig:landscape_schedule}, the weight loss landscapes of checkpoints at different epochs from the early stage (blue curves) are on the middle column, while the weight loss landscapes from the late stage (red curves) are on the right column.
Due to the different learning rate schedules, the robust generalization gap becomes large at different epochs. The  cosine schedule enlarges the robust generalization gap after the 120-th epoch with $lr < 0.34$. The weight loss landscape becomes sharper correspondingly. The cyclic schedule starts to significantly enlarge the gap much later, almost after the 175-th epoch with $lr < 0.16$. Meanwhile, the weight loss landscape also becomes sharp much later. Generally, no matter which epoch and learning rate it is, we can always find that the weight loss landscape becomes sharper immediately after the robust generalization gap increases, which implies a clear correlation between weight loss landscape and robust generalization gap.

\subsection{The Connection across Model Architectures}
\begin{figure}[!htbp]
\centering
    \subfigure[VGG-19]{
        \includegraphics[width=0.27\columnwidth]{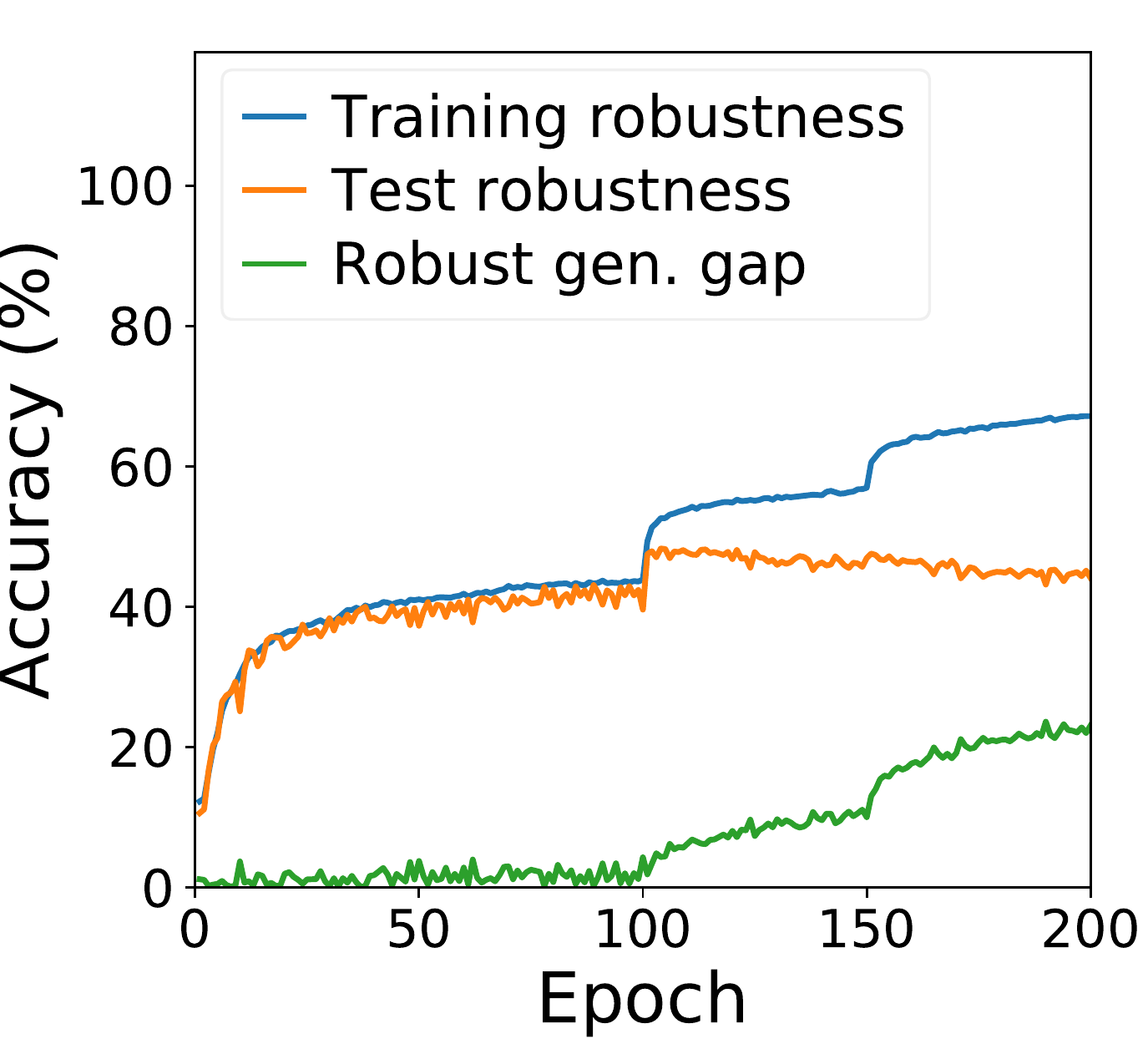}
        \includegraphics[width=0.27\columnwidth]{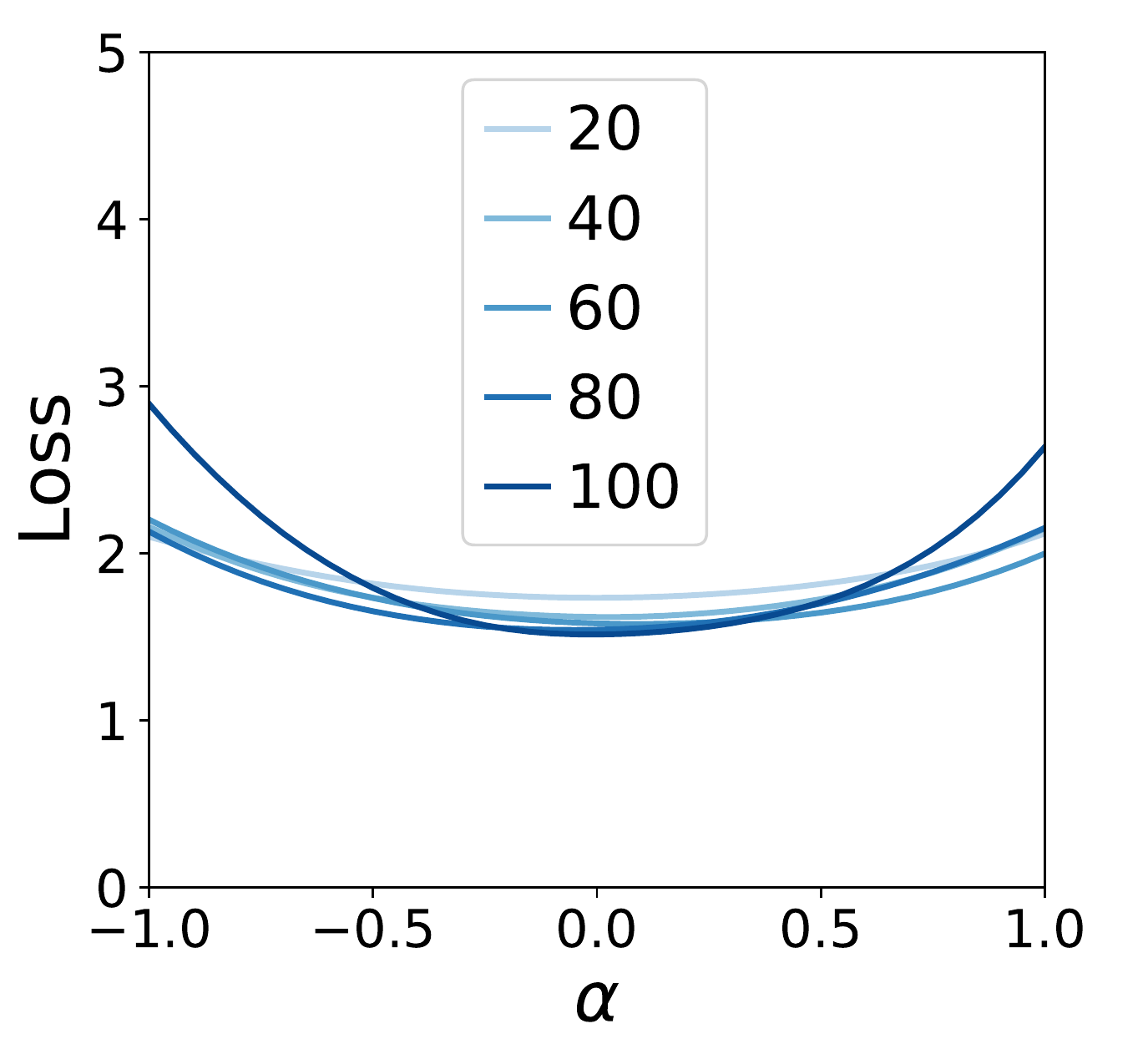}
        \includegraphics[width=0.27\columnwidth]{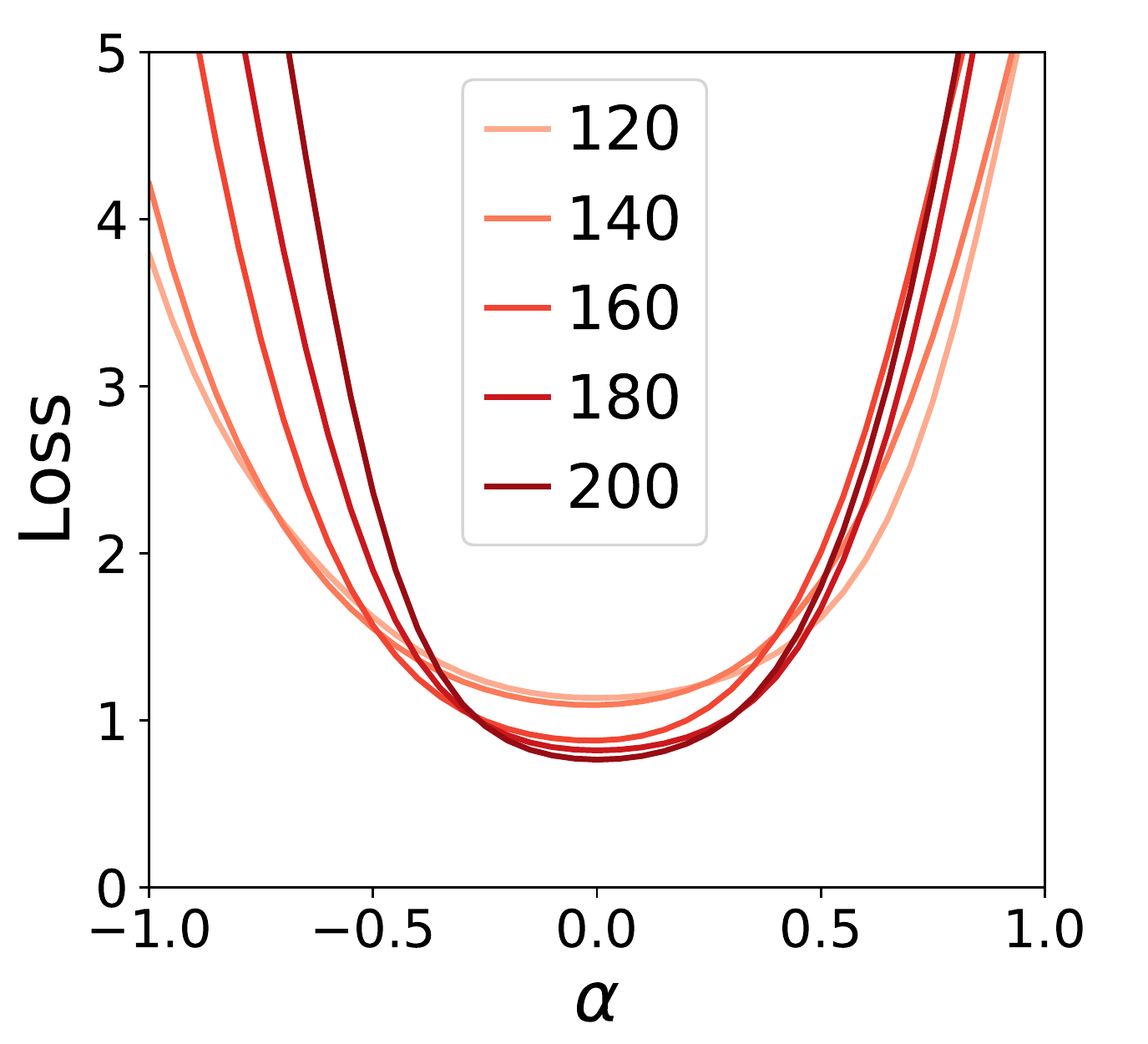}
    }
    \subfigure[WideResNet-34-10]{
        \includegraphics[width=0.27\columnwidth]{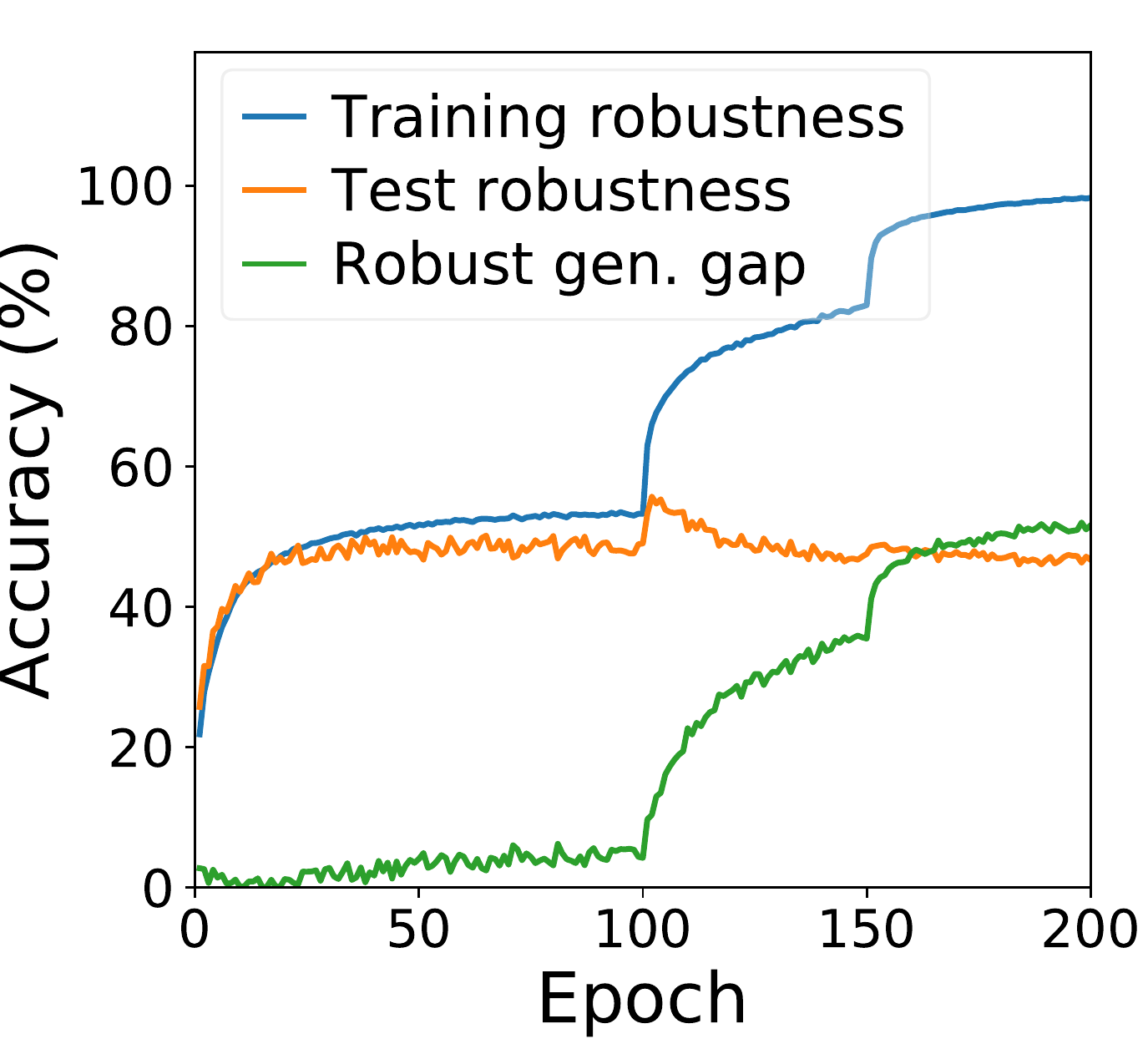}
        \includegraphics[width=0.27\columnwidth]{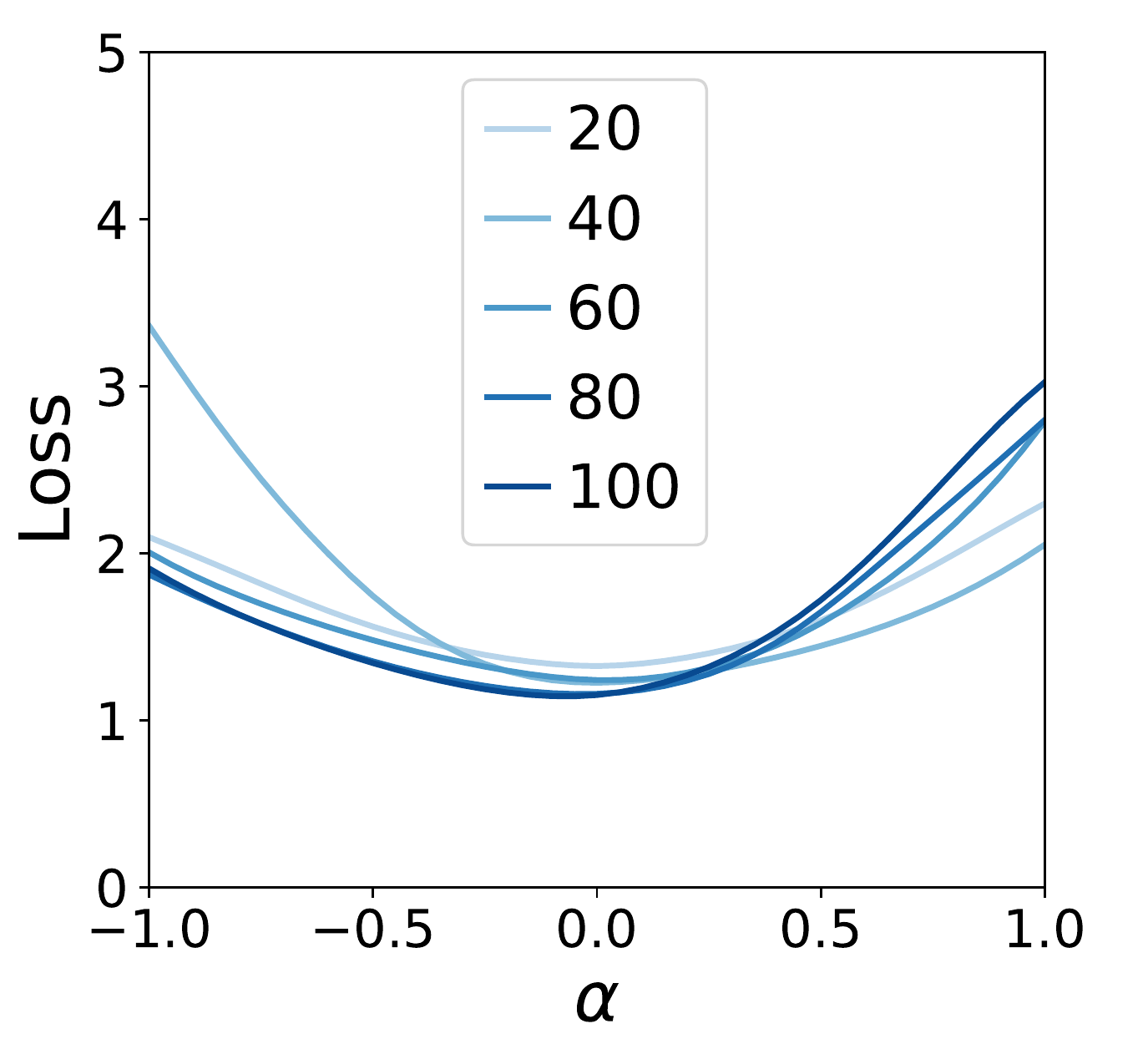}
        \includegraphics[width=0.27\columnwidth]{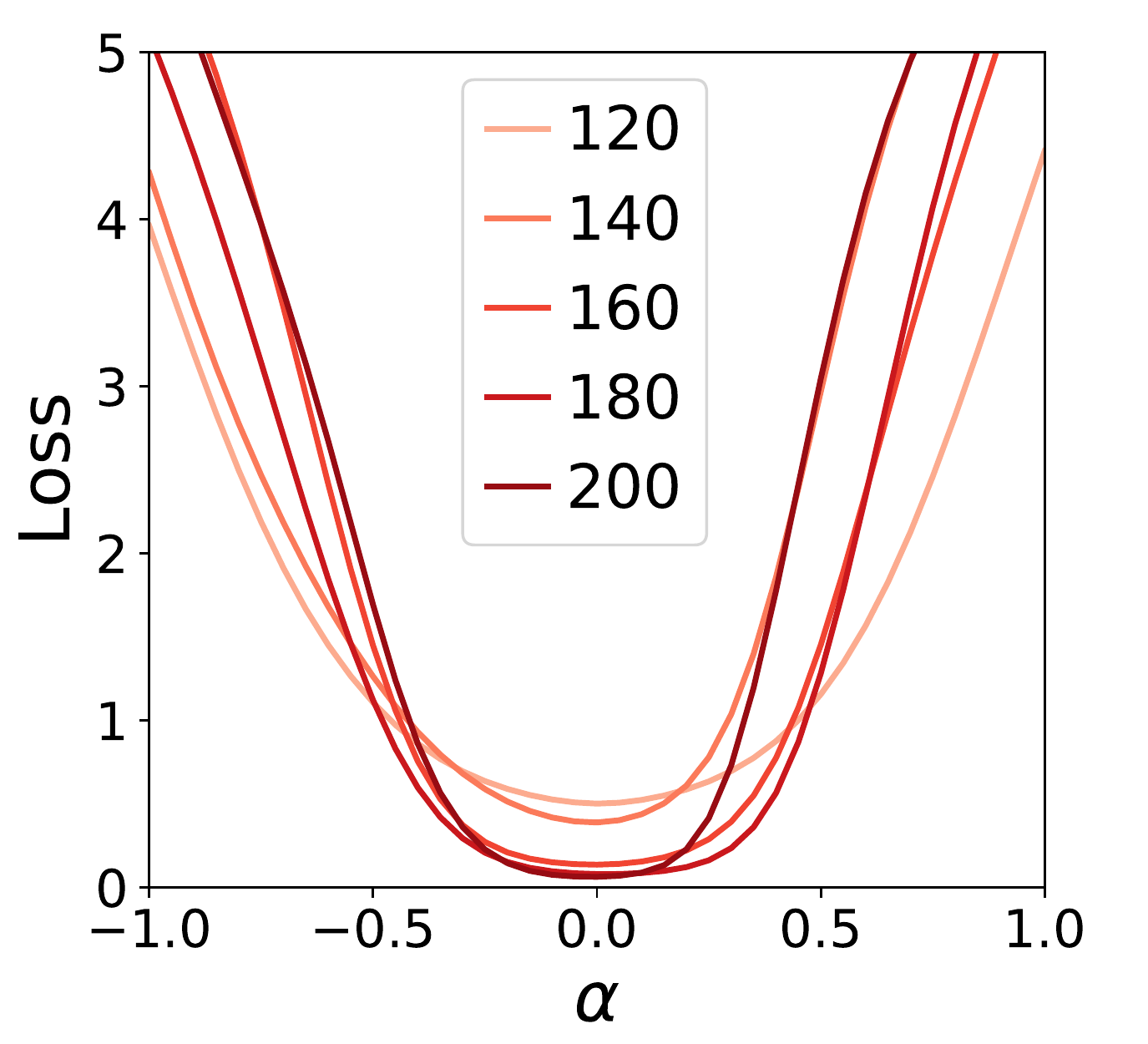}
    }
\vskip -0.1in
\caption{The relationship between weight loss landscape and robust generalization gap across model architectures (VGG-19 and WideResNet-34-10) on CIFAR-10 using piece-wise learning rate schedule and $L_\infty$ attack.}
\label{fig:landscape_arch}
\end{figure}

The previous experiments are all based on PreAct ResNet-18. Here we additionally conduct experiments with VGG-19 \cite{simonyan2015very} and WideResNet-34-10 \cite{zagoruyko2016wide} to verify the connection across network architectures. The same experimental settings as Section \ref{sec:geometry} are adopted and the results are shown in Figure \ref{fig:landscape_arch}. 
They all behave similarly: once the robust generalization gap increases (after the first learning rate decay), the weight loss landscape becomes sharper. This indicates that the connection of weight loss landscape and robust generalization gap still exists across architectures.

\begin{figure}[!htbp]
\centering
    \subfigure[CIFAR-100]{
        \includegraphics[width=0.27\columnwidth]{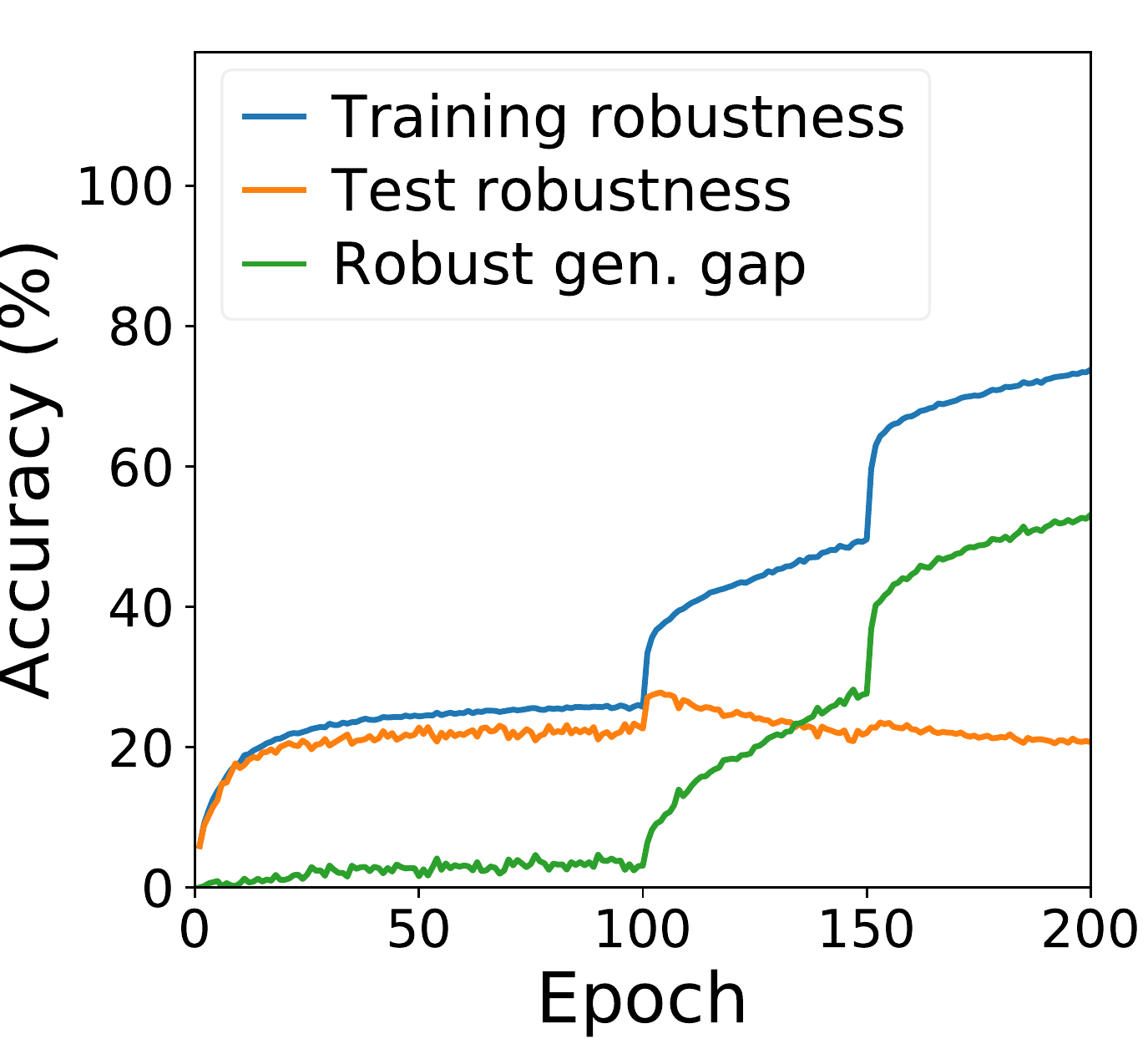}
        \includegraphics[width=0.27\columnwidth]{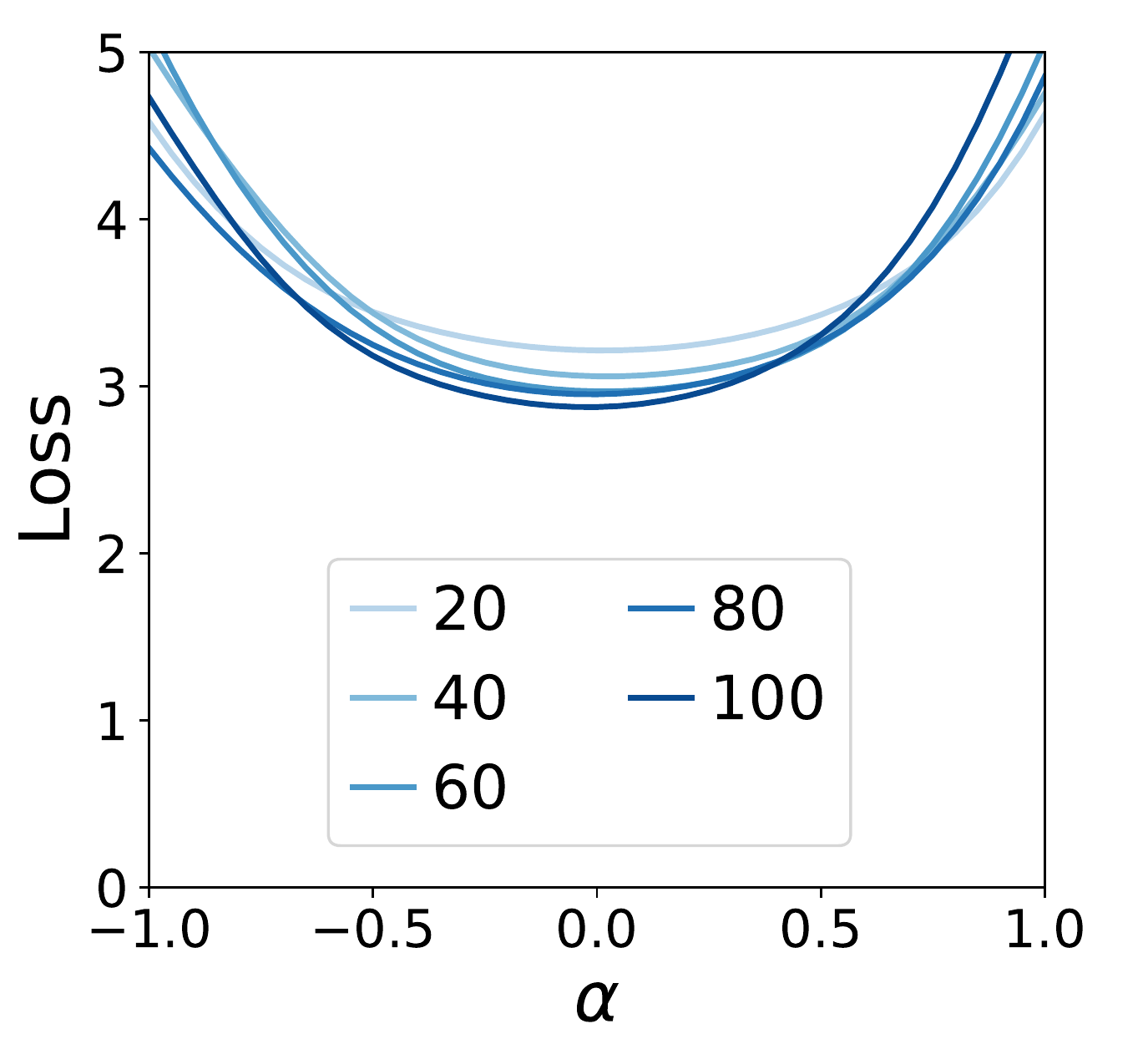}
        \includegraphics[width=0.27\columnwidth]{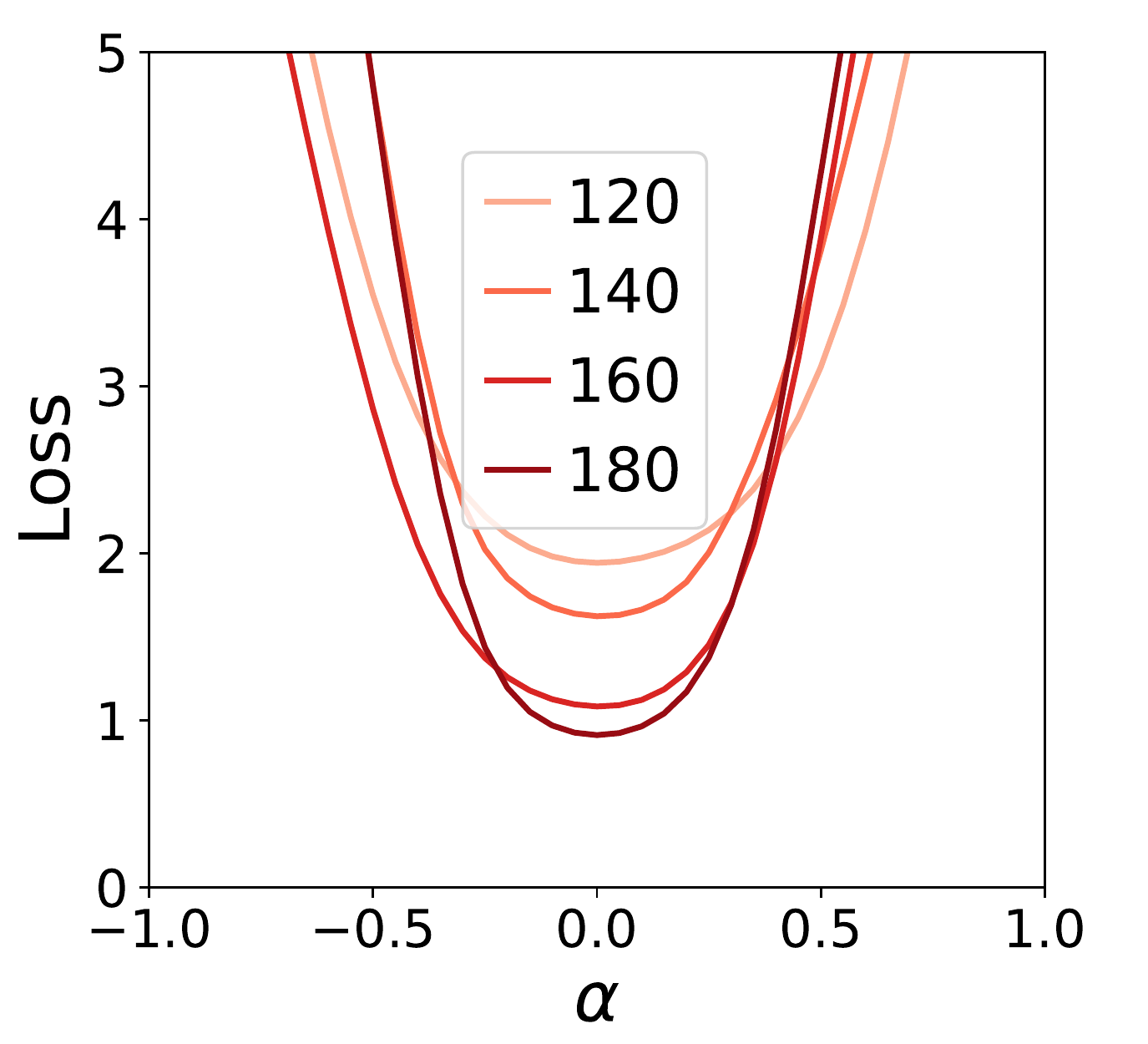}
    }\\
    \subfigure[SVHN]{
        \includegraphics[width=0.27\columnwidth]{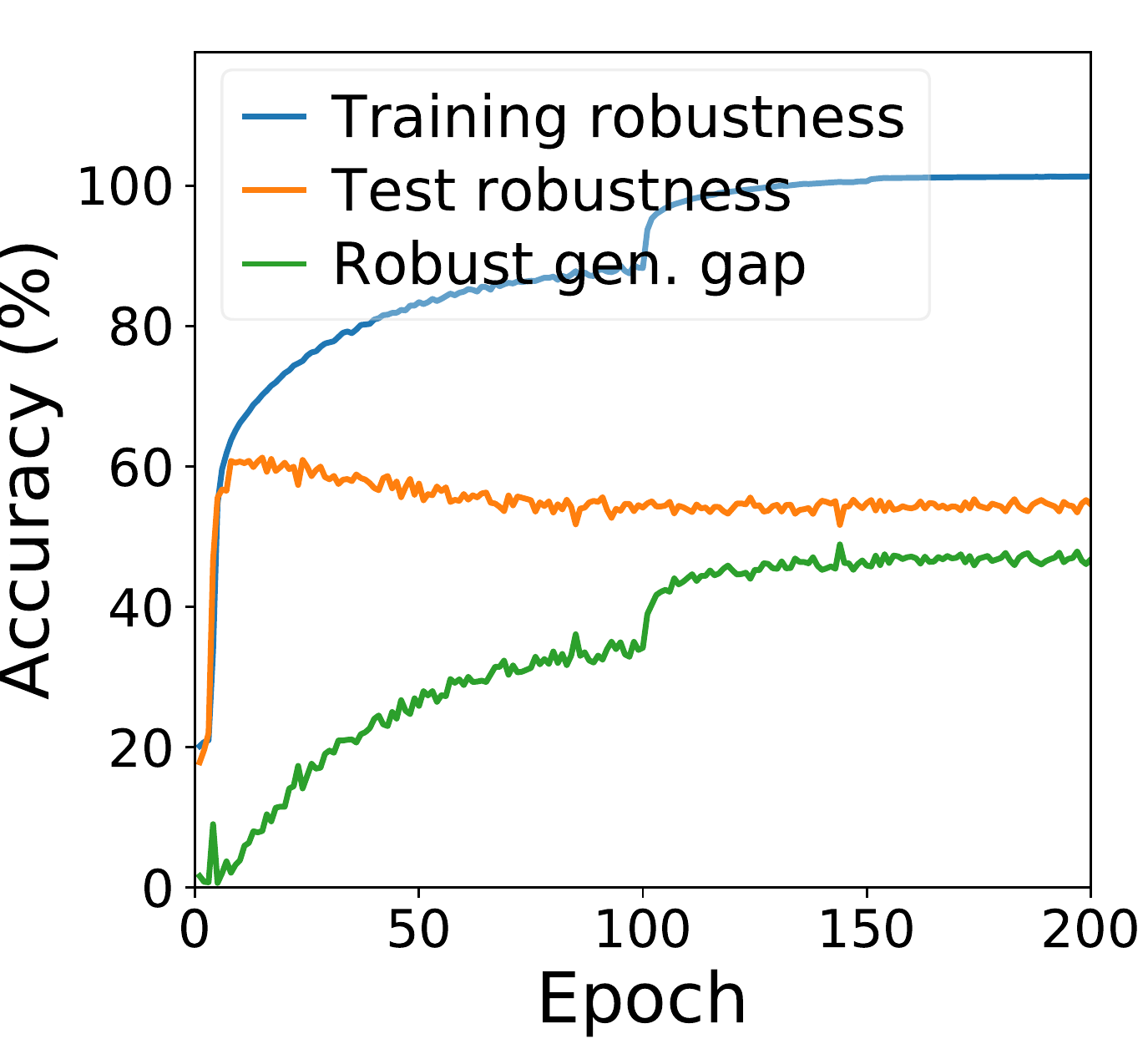}
        \includegraphics[width=0.27\columnwidth]{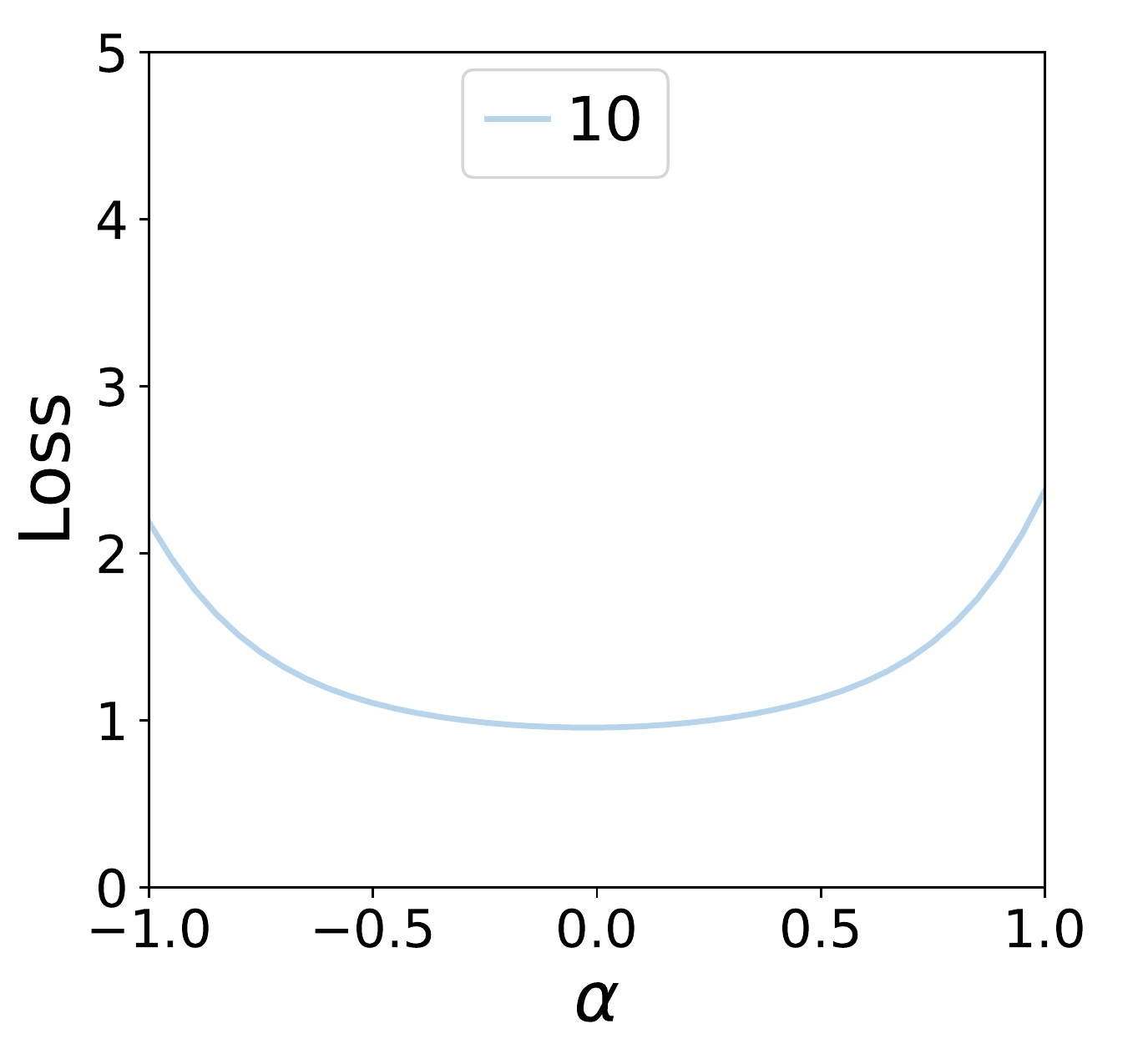}
        \includegraphics[width=0.27\columnwidth]{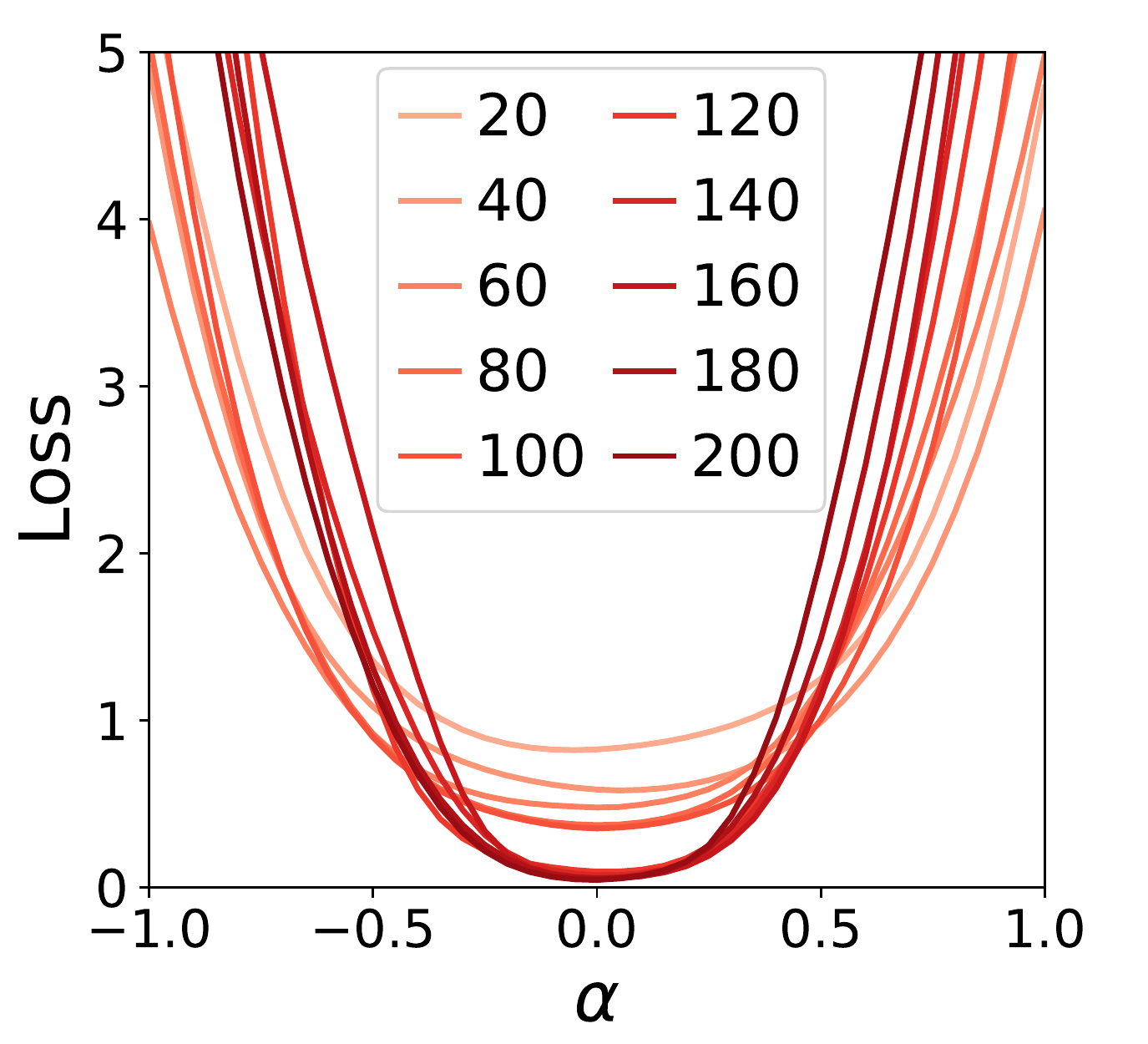}
    }
\vskip -0.1in
\caption{The relationship between weight loss landscape and robust generalization gap across datasets (CIFAR-100 and SVHN) with PreAct ResNet-18 using piece-wise learning rate schedule and $L_\infty$ attack.}
\label{fig:landscape_dataset}
\end{figure}

\subsection{The Connection across Datasets}
\vspace{-0.05 in}
Next, we demonstrate that the connection is an universal phenomenon across datasets on CIFAR-100 \cite{krizhevsky2009learning} and SVHN \cite{netzer2011reading}. We adversarially train PreAct ResNet-18 on different datasets with the same settings as Section \ref{sec:geometry}. The results are shown in Figure \ref{fig:landscape_dataset}. The phenomenon on CIFAR-100 is similar to that on CIFAR-10, \textit{i.e.}, after the first learning rate decay, the robust generalization gap becomes larger and the weight loss landscape becomes sharper as well. For SVHN, the robust generalization gap increases significantly even earlier: it almost achieves its highest robustness around 10-th epoch, and starts overfitting. Meanwhile, the weight loss landscape also keeps flat at 10-th epoch and starts to become sharper. In conclusion, the strong connection of weight loss landscape and robust generalization gap is universal across datasets.

\subsection{The Connection on $L_2$ Threat Model}
\begin{figure}[!htbp]
\centering
    \subfigure[$L_2$ threat model]{
        \includegraphics[width=0.27\columnwidth]{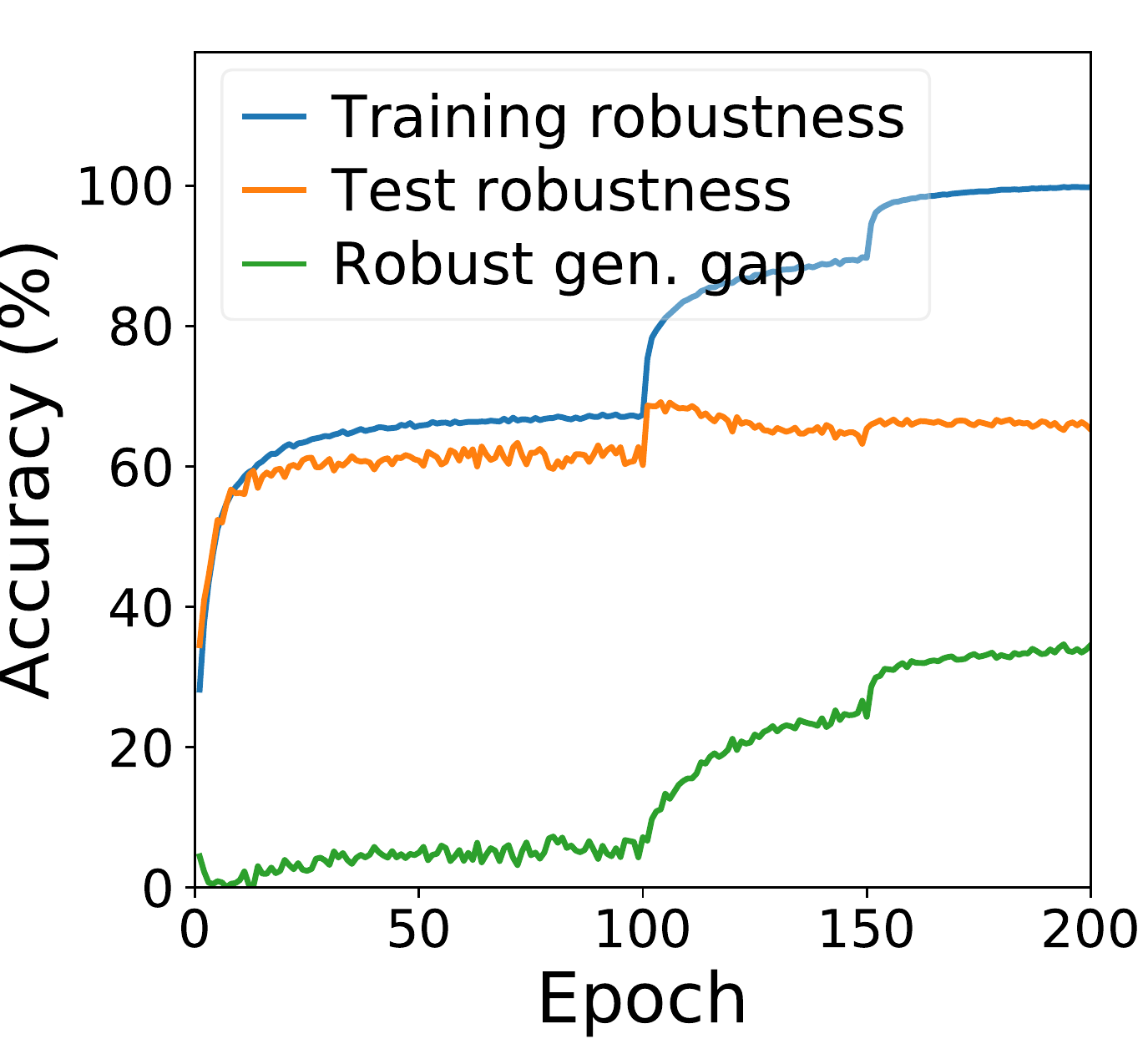}
        \includegraphics[width=0.27\columnwidth]{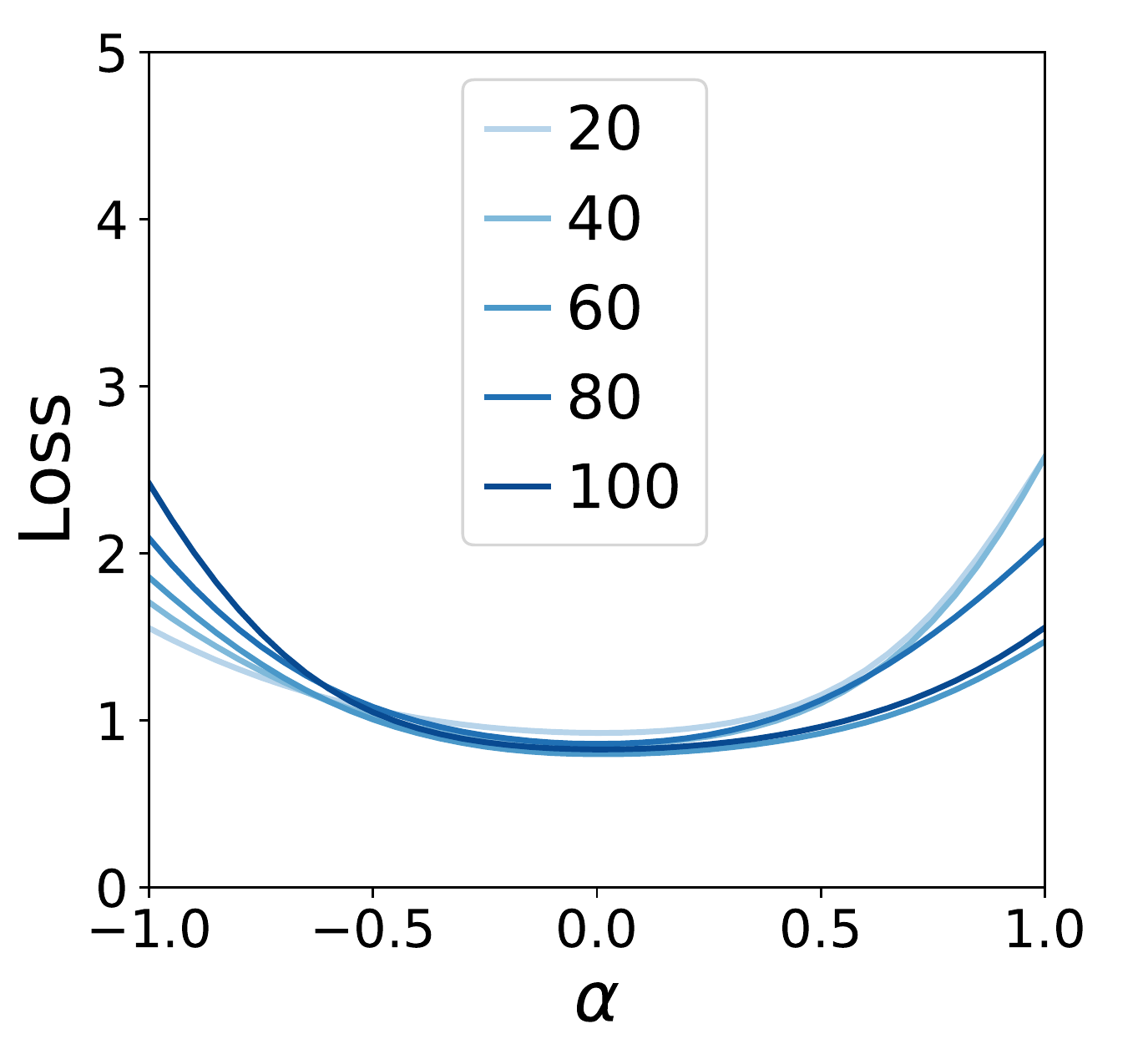}
        \includegraphics[width=0.27\columnwidth]{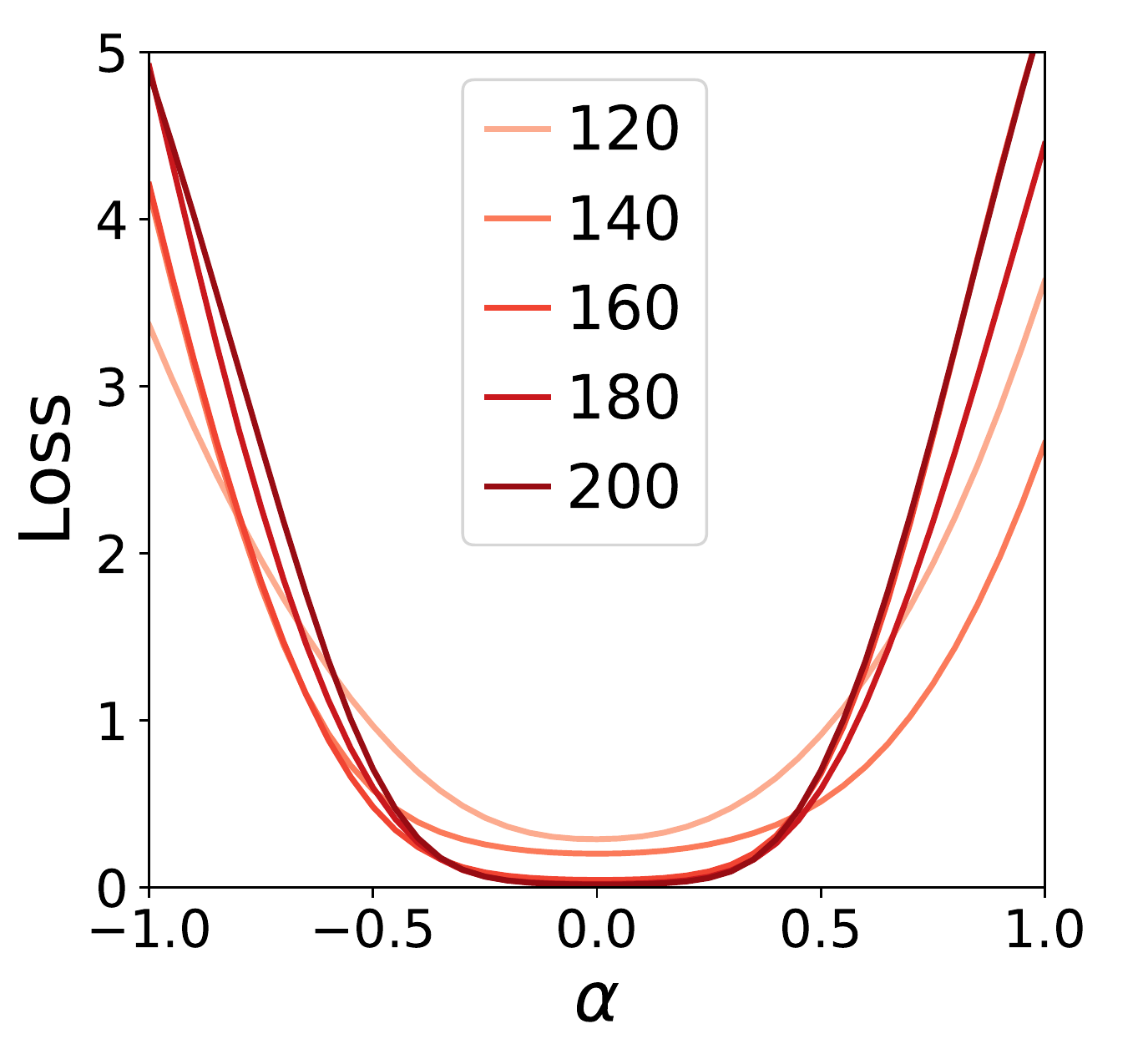}
    }
\vskip -0.1in
\caption{The relationship between weight loss landscape and robust generalization gap with PreAct ResNet-18 on CIFAR-10 using piece-wise learning rate schedule and $L_2$ attack.}
\label{fig:landscape_l2}
\end{figure}

\vspace{-0.05 in}
To further explore the universality of the connection, we additionally conduct experiments on $L_2$ threat model in Figure \ref{fig:landscape_l2}. The other experimental settings are the same as Section \ref{sec:geometry}. Under the $L_2$ threat model, the connection still exists: once the robust generalization gap increases (after the first learning rate decay), the weight loss landscape becomes sharper. 

\vspace{-0.05 in}
\section{Algorithms for the AWP-based Defense}
\label{sec:awp_code}

\vspace{-0.05 in}
In this section, we first provide the pseudo-code of the AWP-based vanilla advesraial trainig (AT-AWP), and then describe how to satisfy the constraint of the perturbation size in Eq. \eqref{eqn:ineq} via the weight update in Eq. \eqref{eqn:nu_direction} and the extensions to other AT variants (TRADES, MART, and RST).

\subsection{Pseudo-code of AT-AWP}
\vspace{-0.05 in}
\begin{algorithm}[!htbp]
   \caption{AT-AWP}
   \label{alg:our_advtrain}
\begin{algorithmic}[1]
   \STATE {\bfseries Input:} Network $\fb_\wb$, training data $\{(\xb_i, y_i)\}_{i=1}^n$, batch size $m$, learning rate $\eta_3$, PGD step size $\eta_1$, PGD steps $K_1$, AWP constraint $\gamma$, AWP step size $\eta_2$, AWP steps $K_2$, alternate iteration $A$.
   \STATE {\bfseries Output:} Robust model $\fb_\wb$.
   \FOR{$t = 0$ to $T - 1$}
   \FOR{$a = 0$ to $A - 1$}
   \FOR{$i = 1, \dots, m$ (in parallel)}
   \STATE $\xb_i^\prime \leftarrow \xb_i + \epsilon \delta$, where $\delta \sim \text{Uniform}(-1, 1)$
   \FOR{$k = 1, \cdots, K_1$}
   \STATE $\xb_i^\prime \leftarrow \Pi_{\epsilon} \big(\xb_i^\prime + \eta_1  \text{sign}( \nabla_{\xb_i^\prime} \ell(\fb_{\wb+\vb}(\xb_i^\prime), y_i)) \big)$
   \ENDFOR
   \ENDFOR
   \FOR{$k = 1, \dots, K_2$}
   \STATE $\vb \leftarrow \Pi_\gamma \big(\vb + \eta_2  \frac{\nabla_\vb \frac{1}{m} \sum_{i} \ell(f_{\wb + \vb}(\xb_i^\prime), y_i)}{\Vert \nabla_\vb \frac{1}{m} \sum_{i} \ell(\fb_{\wb + \vb}(\xb_i^\prime), y_i) \Vert} \Vert \wb \Vert \big)$
   \ENDFOR
   \ENDFOR
   
   \STATE $\wb \leftarrow (\wb+\vb) - \eta_3 \nabla_{\wb+\vb} \frac{1}{m}\sum_{i} \ell(f_{\wb+\vb}(\xb_i^\prime, y_i)) - \vb$
   \ENDFOR
\end{algorithmic}
\end{algorithm}

\newpage
\subsection{Details for the Weight Update under the Constraint}
Recall that we restrict the weight perturbation $\vb_l$ in the $l$-th layer using its relative size to the corresponding weight $\wb_l$, \textit{i.e.}, $\Vert \vb_l \Vert \leq \gamma \Vert \wb_l \Vert$. To implement this restriction on weight perturbation, we apply a layer-wise projection operation,
\textit{i.e.}, once the weight perturbation is out of the ball, we project it back onto the surface of the ball. We have the following layer-wise projection operator,
\begin{equation}
\Pi_\gamma(\vb)=\left\{
\begin{aligned}
\gamma \frac{\Vert \wb_l \Vert}{\Vert \vb_l \Vert} \vb_l, \quad  &\text{if } \Vert \vb_l \Vert > \gamma \Vert \wb_l \Vert, \forall l \in \{1, \cdots, L\} \\
\vb_l, \quad \quad \quad &\text{if } \Vert \vb_l \Vert \leq \gamma \Vert \wb_l \Vert, \forall l \in \{1, \cdots, L\}.
\end{aligned}
\right.
\end{equation}

By default, we set $\eta_2 = \frac{\gamma}{A \cdot K_2} $ (the notations refer to Algorithm \ref{alg:our_advtrain}).

\subsection{Extensions of AWP to Other Adversarial Training Methods}
Our proposed AWP is a general method and can be easily extended to other well-recognized adversarial training variants including TRADES, MART and RST where the only difference is the method-specific adversarial loss in Eq. \eqref{eqn:robust_loss}.

Specifically, for AWP-based TRADES (TRADES-AWP), we also first generate adversarial examples following TRADES method, 
\begin{equation}
\xb_i^\prime \leftarrow \Pi_\epsilon \big(\xb_i^\prime + \eta_1 \text{sign} (\nabla_{\xb_i^\prime} \text{KL}( \fb_{\wb+\vb}(\xb_i) \Vert \fb_{\wb+\vb}(\xb_i^\prime) )) \big),
\label{eqn:craft_adv_example_trades}
\end{equation}
and then the AWP $\vb$ and the DNN parameter $\wb$ of TRADES are updated similarly following Eq. \eqref{eqn:nu_direction} and Eq. \eqref{eqn:w_update} respectively, where $\ell(\fb_{\wb+\vb}(\xb_i^\prime), y_i)$ is TRADES-specific as $\text{CE}( \fb_{\wb+\vb}(\xb_i), y_i ) + \beta \cdot \text{KL}(\fb_{\wb+\vb}(\xb_i) \Vert f_{\wb+\vb}(\xb^\prime_i))$. 

Similarly, for AWP-based MART (MART-AWP), we generate adversarial examples following MART method,
\begin{equation}
\xb_i^\prime \leftarrow \Pi_\epsilon \big(\xb_i^\prime + \eta_1 \text{sign}(\nabla_{\xb_i^\prime} \ell(\fb_{\wb+\vb}(\xb_i^\prime), y_i) ) \big),
\label{eqn:craft_adv_example}
\end{equation}
and then update the AWP $\vb$ follow Eq. \eqref{eqn:nu_direction}. Next, the DNN parameter $\wb$ of MART is updated using Eq. \eqref{eqn:w_update}, where $\ell(\fb_{\wb+\vb}(\xb_i^\prime), y_i)$ is MART-specific loss as  $ \text{BCE} (\fb_\wb(\xb^\prime_i), y_i) + \lambda \cdot \text{KL}(\fb_\wb(\xb_i) \Vert \fb_\wb(\xb^\prime_i)) \cdot (1-[\fb_\wb(\xb_i)]_{y_i})$. 

RST, as an SSL-based method, first generates pseudo labels for unlabeled data, and then adversarially train DNNs using TRADES loss on the new dataset which consists of labeled data and unlabeled data with pseudo labels. Thus, we can incorporate AWP into RST just like TRADES-AWP.

\section{More Results for Section 4.3: A Case Study on Vanilla AT and AT-AWP}
\label{sec:case_study}

Following Section 4.3, we provide the complete results (test robustness and natural accuracy) of AT and AT-AWP in Table \ref{table:overview_preactresnet}. Under $L_2$ threat model, AWP improves both the robustness and natural accuracy on all datasets. While under $L_\infty$ threat model, AWP improves the robustness on condition of sacrificing the natural accuracy on CIFAR-10 and CIFAR-100. This maybe because natural images (CIFAR) are more complicated than color digits (SVHN). 
However, for robustness, AWP demonstrates a general behaviour that consistently improves the best and last robustness by a recognizable gain across datasets and threat models.

\begin{table}[!htbp]
\caption{Performance (\%) of AT and AT-AWP on PreAct ResNet-18 across different datasets and threat models over 5 random runs.}
    \label{table:overview_preactresnet}
    \centering
    \begin{tabular}{lclcccc}
    \toprule
    \multirow{2}[4]{*}{Dataset} &
    \multirow{2}[4]{*}{Norm} & 
    \multirow{2}[4]{*}{Method} & \multicolumn{2}{c}{Robustness} & \multicolumn{2}{c}{Natural accuracy} \\
    \cmidrule(lr){4-5}
    \cmidrule(lr){6-7} & & & Best & Last & Best & Last \\ \midrule
    \multirow{4}{*}{SVHN}      & \multirow{2}{*}{$L_\infty$} & AT & 53.36 $\pm$ 0.09 & 44.49 $\pm$ 0.27 & 92.18  $\pm$ 0.15 & 89.85 $\pm$ 0.33 \\
    &          & AT-AWP & \textbf{59.12 $\pm$ 0.26} & \textbf{55.87 $\pm$ 0.39} & \textbf{93.85 $\pm$ 0.11} & \textbf{92.59 $\pm$ 0.52}   \\ \cmidrule(lr){2-7}
       & \multirow{2}{*}{$L_2$}      & AT   & 66.87 $\pm$ 0.25 & 65.03 $\pm$ 0.24 & 93.69 $\pm$ 0.12 & 93.25 $\pm$ 0.28 \\
       &          & AT-AWP & \textbf{72.57 $\pm$ 0.40} & \textbf{67.73 $\pm$ 0.21} & \textbf{95.95 $\pm$ 0.36} & \textbf{95.22 $\pm$ 0.27}   \\ \midrule
    \multirow{4}{*}{CIFAR-10}  & \multirow{2}{*}{$L_\infty$} & AT & 52.79 $\pm$ 0.21 & 44.44 $\pm$ 0.39 & \textbf{85.57 $\pm$ 0.16} & \textbf{84.56 $\pm$ 0.19} \\
    & &  AT-AWP & \textbf{55.39 $\pm$ 0.39} & \textbf{54.73 $\pm$ 0.16}  & 82.00 $\pm$ 0.19 & 81.11 $\pm$ 0.39 \\ \cmidrule(lr){2-7}
    & \multirow{2}{*}{$L_2$} & AT & 69.15 $\pm$ 0.13 & 65.93 $\pm$ 0.35 & 89.57 $\pm$ 0.09 & 88.96 $\pm$ 0.18  \\
    &          & AT-AWP& \textbf{72.69 $\pm$ 0.19} & \textbf{72.08 $\pm$ 0.39}  & \textbf{90.18 $\pm$ 0.31} & \textbf{89.71 $\pm$ 0.17}  \\ \midrule
    \multirow{4}{*}{CIFAR-100} & \multirow{2}{*}{$L_\infty$} & AT & 27.22 $\pm$ 0.16 & 20.82 $\pm$ 0.20 & \textbf{56.33 $\pm$ 0.23} & \textbf{54.61 $\pm$ 0.33}  \\
    &          & AT-AWP & \textbf{30.71 $\pm$ 0.25} & \textbf{30.28 $\pm$ 0.30}  & 54.19 $\pm$ 0.39 & 54.39 $\pm$ 0.29  \\ \cmidrule(lr){2-7}
    & \multirow{2}{*}{$L_2$} & AT & 41.33 $\pm$ 0.09 & 35.27 $\pm$ 0.29 &  62.65 $\pm$ 0.11 & 60.50 $\pm$ 0.17 \\
    &          & AT-AWP & \textbf{45.60 $\pm$ 0.23} & \textbf{44.66 $\pm$ 0.22}  & \textbf{65.07 $\pm$ 0.31} & \textbf{64.40 $\pm$ 0.41} \\ \bottomrule
    \end{tabular}
\end{table}

\section{More Experiments on Comparison to Other Regularization Techniques}
\label{sec:cifar10}

\begin{figure}[!htbp]
\centering
    \subfigure[Test robustness under $\ell_\infty$ threat model]{
        \includegraphics[width=0.46\columnwidth]{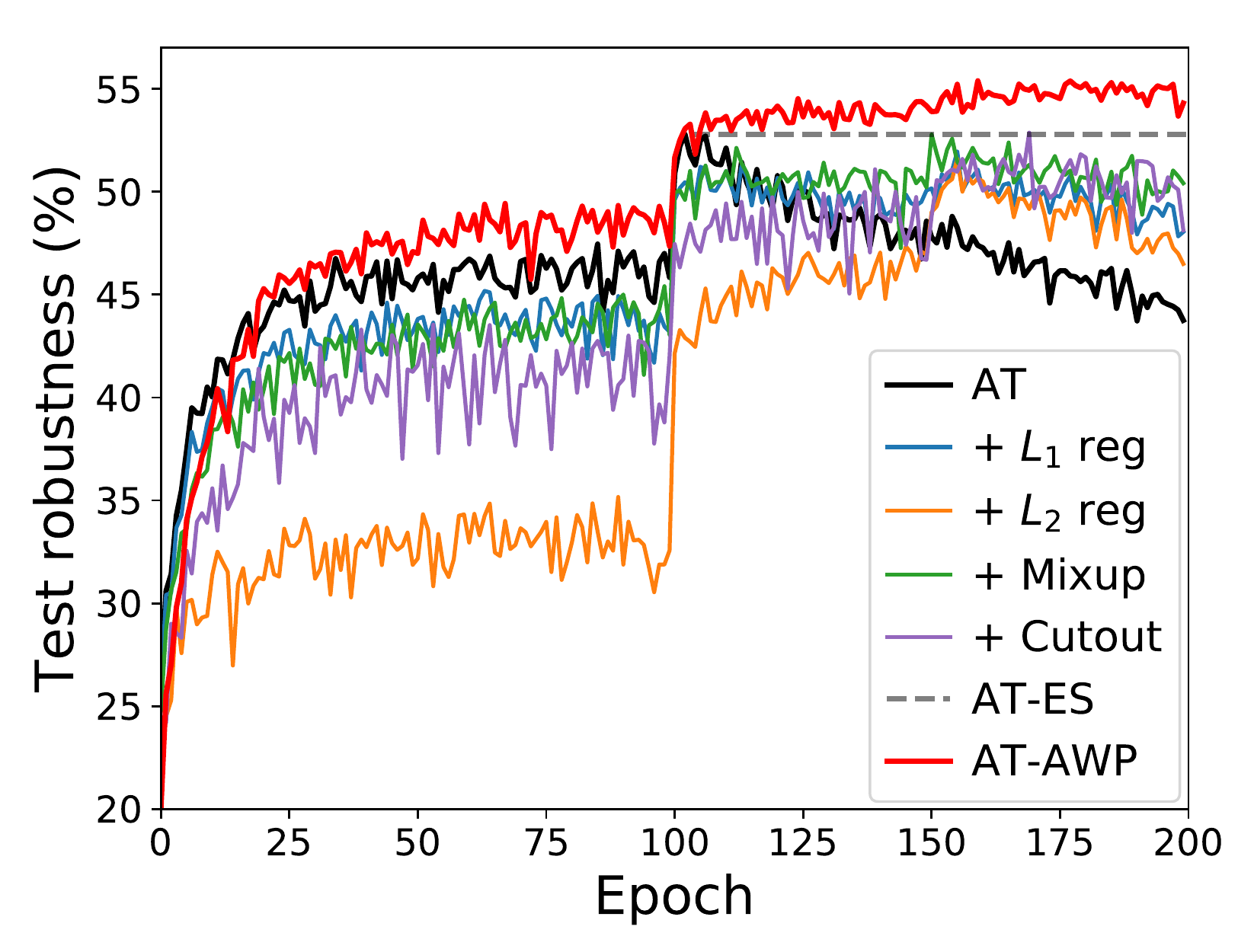}
    }
    \subfigure[Natural accuracy under $\ell_\infty$ threat model]{
        \includegraphics[width=0.46\columnwidth]{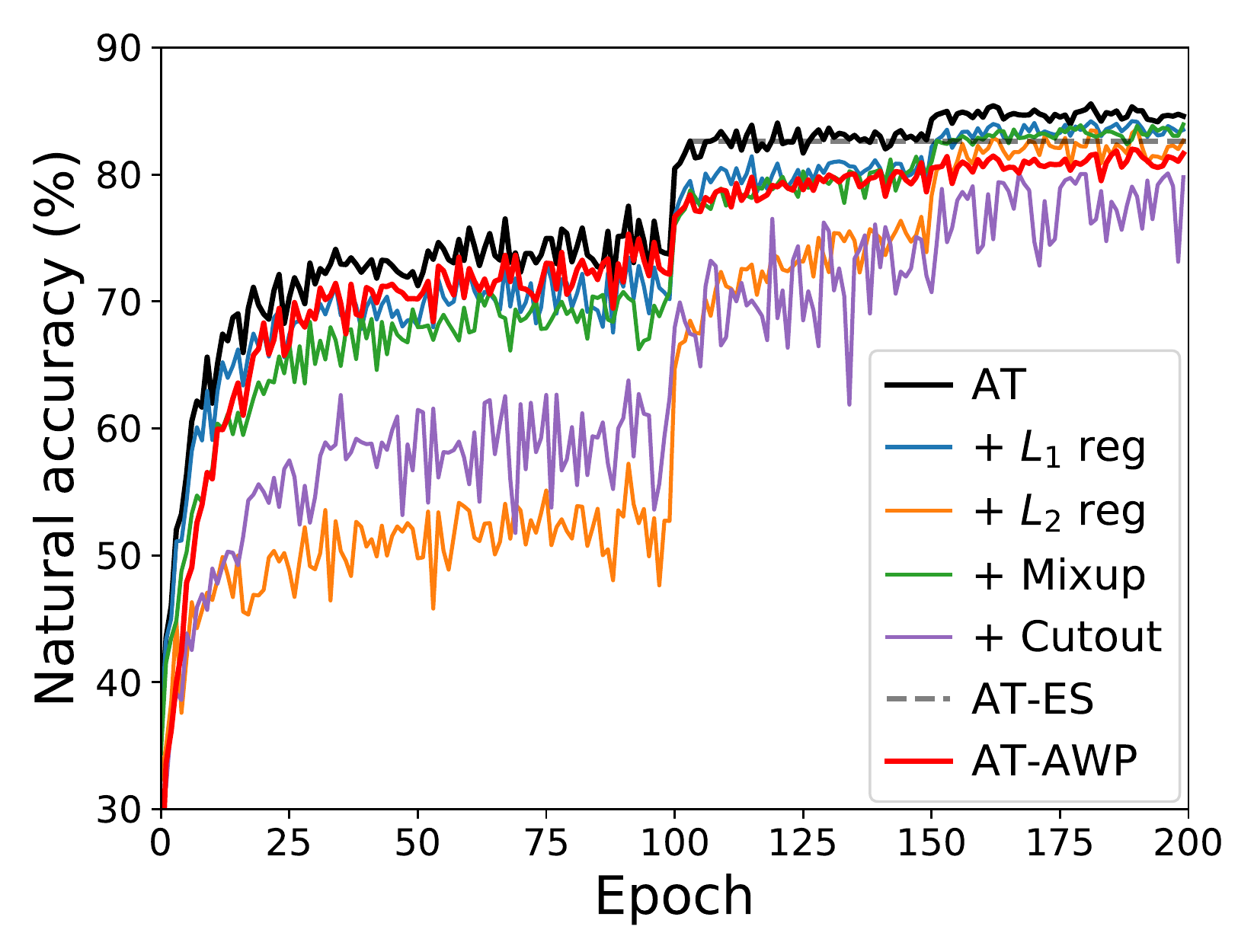}
    } \\
    \subfigure[Test robustness under $\ell_2$ threat model]{
	\includegraphics[width=0.46\columnwidth]{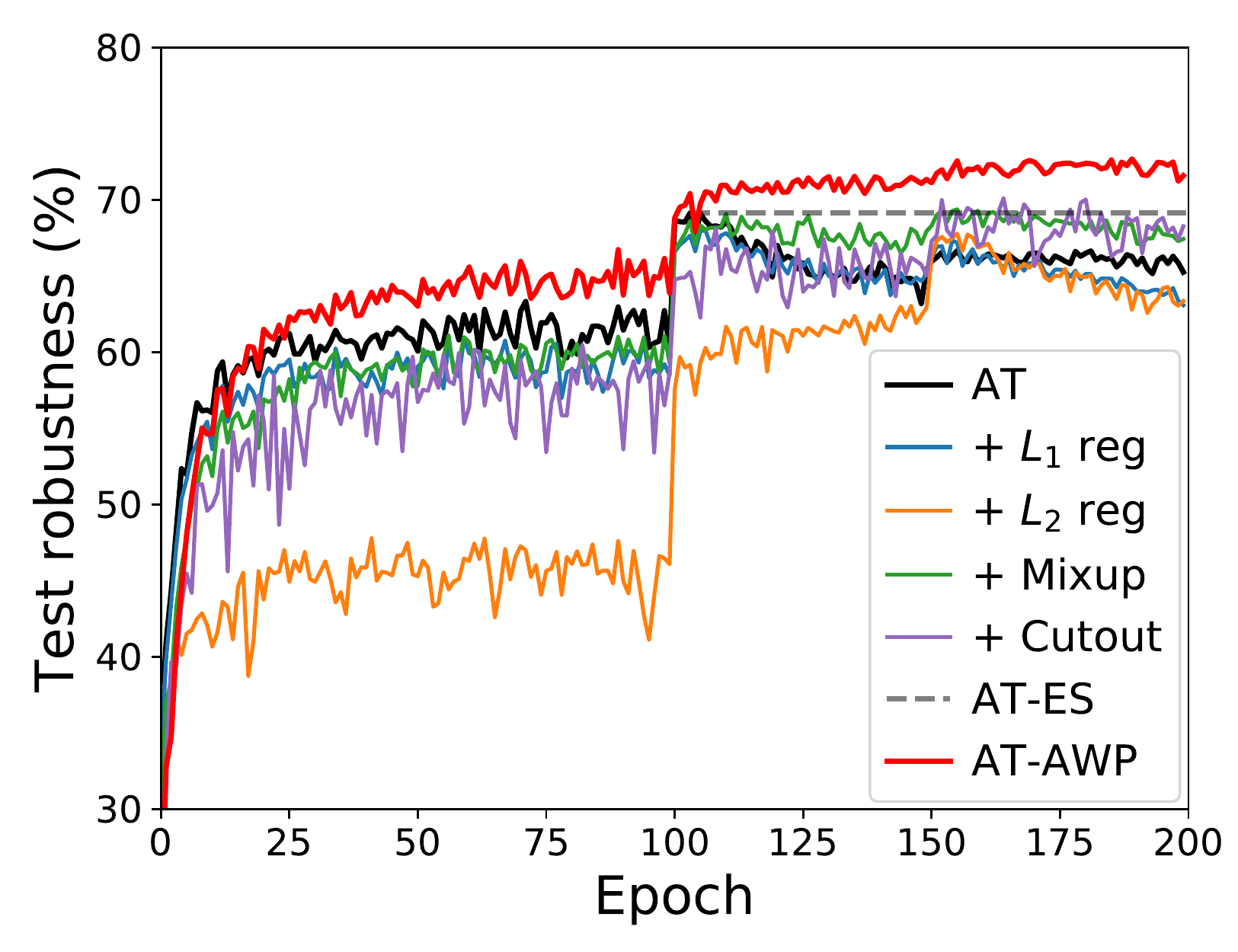}
    }
    \subfigure[Natural accuracy under $\ell_2$ threat model]{
	\includegraphics[width=0.46\columnwidth]{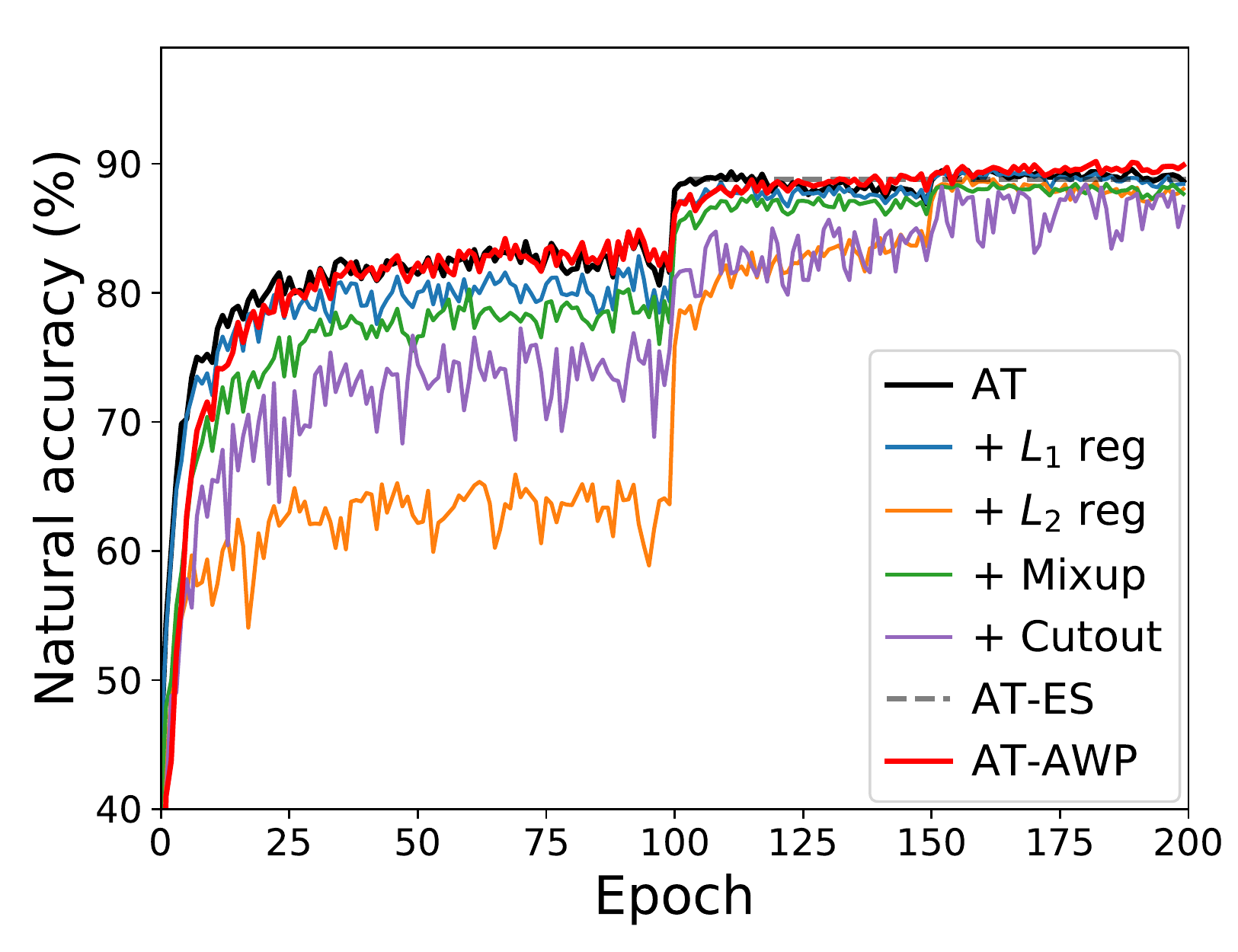}
    }
\vskip -0.1in
\caption{Performance (\%) on the test set of CIFAR-10 during the training for AT, AT-AWP, and AT with other regularization techniques.}
\label{fig:learning_curve}
\end{figure}

Following Section 5.3, here we provide the complete results (test robustness and natural accuracy) of AWP and several regularization techniques under both $L_\infty$ and $L_2$ threat models on CIFAR-10. Under the $L_\infty$ threat model, we follow the best hyper-parameters tuned in \citet{rice2020overfitting}: $\lambda = 5 \times 10^{-6}$/$5 \times 10^{-3}$ for $L_1$/$L_2$ regularization respectively, patch length 14 for cutout, and $\alpha = 1.4$ for mixup. Under the $L_2$ threat model, we use the same hyper-parameters since we find that they almost have the same trends after parameter tuning. For AWP, we all set $\gamma = 5 \times 10^{-3}$. We use the same training settings as Section \ref{sec:exploring} and the test attack is PGD-20. We show test robustness and test natural accuracy in Table \ref{table:reg_cifar10_linf} ($L_\infty$ threat model) and Table \ref{table:reg_cifar10_l2} ($L_2$ threat model). We find AWP indeed improves the test robustness of both the best checkpoint and the last checkpoint by a notable margin. Besides, we visualize the learning curves in Figure \ref{fig:learning_curve}. We find that AWP can constantly improve the robustness under both $L_\infty$ and $L_2$ threat models throughout the entire training process, which demonstrates the superiority of AWP over other regularization methods. 

In addition, we find AWP sacrifices the natural accuracy throughout the whole training on CIFAR-10 under $L_\infty$ threat model, which indicates AWP is still affected by the trade-off between robustness and  natural accuracy \cite{Zhang2019theoretically}. For $L_2$ threat model on CIFAR-10, AWP just has similar natural accuracy to vanilla AT. This is because $L_2$ threat model ($\epsilon=128/255$) is easier than $L_\infty$ threat model and many techniques have the similar natural accuracy.

\begin{table}[!htbp]
\caption{Performance (\%) of AT and AT with other regularization techniques on CIFAR-10 using PreAct ResNet-18 under $L_\infty$ threat model ($\epsilon=8/288$) over 5 random runs.}
  \label{table:reg_cifar10_linf}
  \centering
  \begin{tabular}{lcccc}
  \toprule
  \multirow{2}[4]{*}{Method} & \multicolumn{2}{c}{Robustness} & \multicolumn{2}{c}{Natural accuracy} \\
  \cmidrule(lr){2-3}
  \cmidrule(lr){4-5}
  & Best & Last & Best & Last  \\ \midrule
  AT & 52.79 $\pm$ 0.21 & 44.44 $\pm$ 0.39 & \textbf{85.57 $\pm$ 0.16} & \textbf{84.56 $\pm$ 0.19} \\
  $+$ $L_1$ regularization & 51.95 $\pm$ 0.51 & 48.76 $\pm$ 0.61 & 82.77 $\pm$ 0.37 & 83.42 $\pm$ 0.26 \\
  $+$ $L_2$ regularization & 51.60 $\pm$ 0.41 & 47.37 $\pm$ 0.52  & 81.05 $\pm$ 0.44 & 81.97 $\pm$ 0.50\\
  $+$ Cutout & 52.78 $\pm$ 0.14 & 50.37 $\pm$ 0.41 & 80.99 $\pm$ 0.19 & 83.46 $\pm$ 0.33 \\
  $+$ Mixup & 52.87 $\pm$ 0.47 & 49.76 $\pm$ 0.81 & 78.76 $\pm$ 0.61 & 78.50 $\pm$ 1.21 \\ \midrule
  AT-AWP & \textbf{55.39 $\pm$ 0.39} & \textbf{54.73 $\pm$ 0.16} & 82.00 $\pm$ 0.19 & 81.11 $\pm$ 0.39 \\
  \bottomrule
  \end{tabular}
\end{table}

\begin{table}[!htbp]
\caption{Performance (\%) of AT and AT with other regularization techniques on CIFAR-10 using PreAct ResNet-18 under $L_2$ threat model ($\epsilon=128/255$) over 5 random runs.}
  \label{table:reg_cifar10_l2}
  \centering
  \begin{tabular}{lcccc}
  \toprule
  \multirow{2}[4]{*}{Method} & \multicolumn{2}{c}{Robustness} & \multicolumn{2}{c}{Natural accuracy} \\
  \cmidrule(lr){2-3}
  \cmidrule(lr){4-5}
  & Best & Last & Best & Last \\ \midrule
  AT & 69.15 $\pm$ 0.13 & 65.93 $\pm$ 0.35 & 89.57 $\pm$ 0.09 & 88.96 $\pm$ 0.18 \\
  $+$ $L_1$ regularization & 67.99 $\pm$ 0.27 & 63.75 $\pm$ 0.40 & 88.04 $\pm$ 0.19 & 88.49 $\pm$ 0.27 \\
  $+$ $L_2$ regularization & 67.78 $\pm$ 0.33 & 63.62 $\pm$ 0.46 & 88.57 $\pm$ 0.14 & 87.75 $\pm$ 0.23 \\
  $+$ Cutout & 69.38 $\pm$ 0.27 & 67.68 $\pm$ 0.30 & 88.36 $\pm$ 0.31 & 88.01 $\pm$ 0.26 \\
  $+$ Mixup & 70.11 $\pm$ 0.50 & 68.20 $\pm$ 0.46 & 87.29 $\pm$ 0.71 & 86.91 $\pm$ 0.94 \\ \midrule
  AT-AWP & \textbf{72.69 $\pm$ 0.19} & \textbf{72.08 $\pm$ 0.39} & \textbf{90.18 $\pm$ 0.31} & \textbf{89.71 $\pm$ 0.17} \\
  \bottomrule
  \end{tabular}
\end{table}

\end{document}